\documentclass[final]{cvpr}

\usepackage[dvipdfm]{graphicx}
\usepackage{comment}
\usepackage{amsmath,amssymb} 
\usepackage{color}

\usepackage{times}
\usepackage{epsfig}
\usepackage{dsfont}

\usepackage{subfigure,times,multirow}
\usepackage{array}
\usepackage[dvipsnames]{xcolor}

\usepackage[ruled]{algorithm2e}
\usepackage{multirow}
\usepackage{cite}
\usepackage{booktabs}
\usepackage{kotex}
\usepackage[normalem]{ulem} 
\usepackage[keeplastbox]{flushend}
\usepackage[verbose=silent]{microtype}
\usepackage{balance}
\usepackage{cuted}

\usepackage[pagebackref=true,breaklinks=true,colorlinks,bookmarks=false]{hyperref}

\usepackage{wrapfig}

\usepackage{pifont}

\DeclareMathOperator*{\argmin}{argmin} 

\usepackage{stfloats}

\def\cvprPaperID{2504} 
\def\confYear{CVPR 2021}

\def\etal{\emph{et al}.}

\definecolor{orange}{rgb}{0.75, 0.4, 0.0}

\newcommand{\hy}{\widehat{y}}
\newcommand{\bx}{\mathbf{x}}

\newcommand{\bz}{\mathbf{z}}
\newcommand{\bw}{\mathbf{w}}

\newcommand{\bhy}{\mathbf{\widehat{y}}}
\newcommand{\IE}{\mathds{E}}

\newcommand{\calP}{\mathcal{P}}

\newcommand{\calM}{\mathcal{M}}

\newcommand{\calL}{\mathcal{L}}
\newcommand{\calH}{\mathcal{H}}
\newcommand{\round}{r}
\newcommand{\numsample}{N_\round}
\newcommand{\unlabeledpool}{\calP_U}
\newcommand{\selectedpool}{\calP_S}
\newcommand{\labeledpool}{\calP_L}
\newcommand{\crossentropy}{\calH}
\newcommand{\softmax}{\text{softmax}}

\newcommand{\Fig}[1]{Fig.~\ref{fig:#1}}

\newcommand{\Eq}[1]{Eq.~\eqref{eq:#1}}

\newcommand{\Section}[1]{Section~\ref{sec:#1}}

\newcommand{\Algorithm}[1]{Algorithm~\ref{alg:#1}}

\newcommand{\Table}[1]{Table~\ref{tbl:#1}}

\renewcommand{\paragraph}[1]{\vspace{0.5em}\noindent\textbf{#1}}

\begin{document}
\title{VaB-AL: Incorporating Class Imbalance and Difficulty \\ with Variational Bayes for Active Learning}

\author{Jongwon Choi$^{1*}$, Kwang Moo Yi$^{2*}$, Jihoon Kim$^{3}$, Jinho Choo$^{3}$,\\
Byoungjip Kim$^{3}$, Jinyeop Chang$^{3}$, Youngjune Gwon$^{3}$, Hyung Jin Chang$^{4}$\vspace{3mm}\\ 
$^1$Chung-Ang University, South Korea \hspace{1.5cm} $^2$University of British Columbia, Canada\\
\hspace{1cm} $^3$Samsung SDS, South Korea \hspace{1.6cm} $^4$University of Birmingham, United Kingdom\\
{\tt\footnotesize choijw@cau.ac.kr, kmyi@cs.ubc.ca, h.j.chang@bham.ac.uk}\vspace{-1mm}\\
{\tt\footnotesize \{jihoon00.kim, jinho12.choo, bjip.kim, jinyeop.chang, gyj.gwon\}@samsung.com}
}

\maketitle
\let\thefootnote\relax\footnote{* These authors contributed equally to this work.}

\begin{abstract}
Active Learning for discriminative models has largely been studied with the focus on individual samples, with less emphasis on how classes are distributed or which classes are hard to deal with.
In this work, we show that this is harmful.
We propose a method based on the Bayes' rule, that can naturally incorporate class imbalance into the Active Learning framework.
We derive that three terms should be considered together when estimating the probability of a classifier making a mistake for a given sample; i) probability of mislabelling a class, ii) likelihood of the data given a predicted class, and iii) the prior probability on the abundance of a predicted class.
Implementing these terms requires a generative model and an intractable likelihood estimation. 
Therefore, we train a Variational Auto Encoder (VAE) for this purpose.
To further tie the VAE with the classifier and facilitate VAE training, we use the classifiers' deep feature representations as input to the VAE.
By considering all three probabilities, among them, especially the data imbalance, we can substantially improve the potential of existing methods under limited data budget.
We show that our method can be applied to classification tasks on multiple different datasets -- including one that is a real-world dataset with heavy data imbalance -- significantly outperforming the state of the art.
\end{abstract}

\section{Introduction}
\label{sec:intro}

\begin{figure}
\centering

\includegraphics[width=\linewidth]{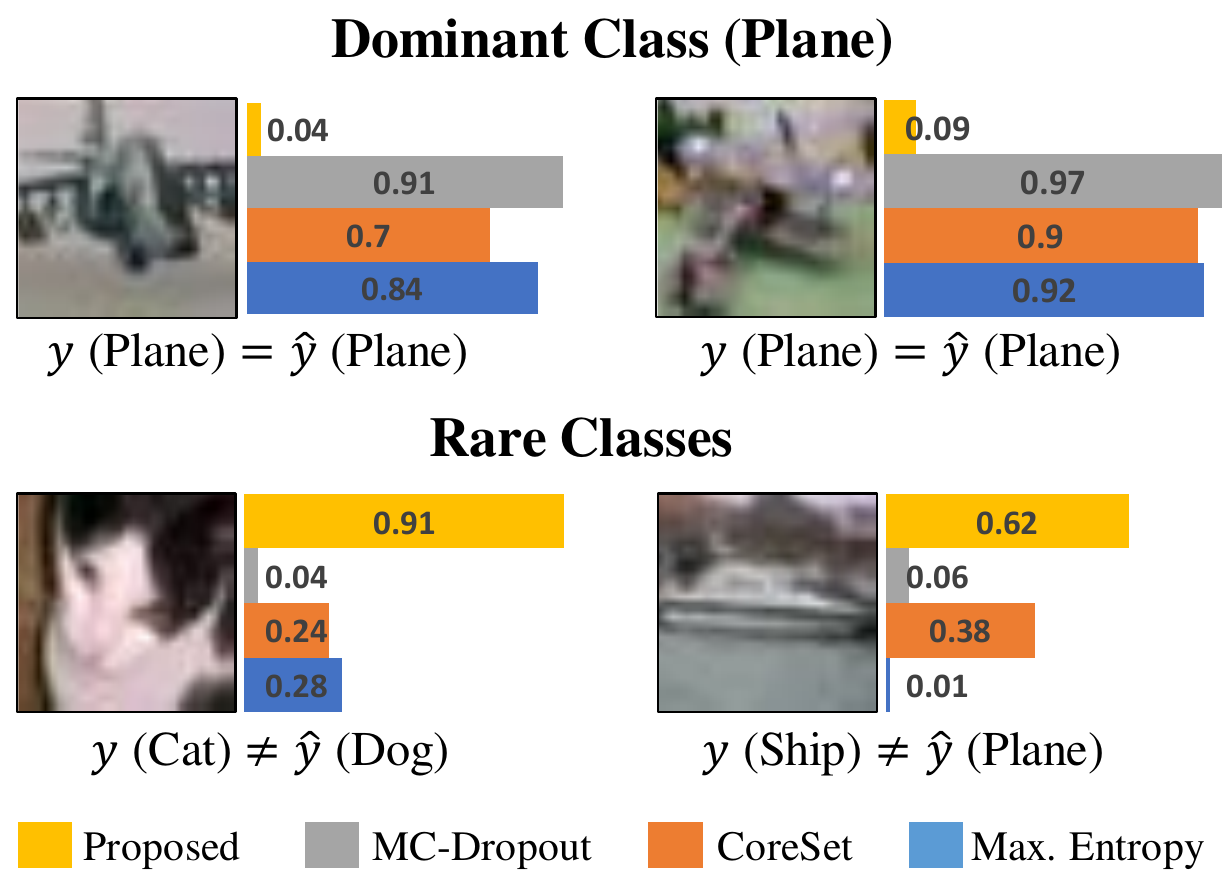}



\caption{
{\bf Class matters -- }
We show an example where existing methods fail to identify important samples to put into the updated training set. 
We show the importance metrics given by various methods -- MC-Dropout~\cite{Gal17_ac}, CoreSet~\cite{Sener18_ac}, Max. Entropy~\cite{dan14_ac}, and our method -- for selected images when the dataset is dominated by plane images.
We further show below each image, the ground-truth class $y$ and the predicted class $\hy$.
Because of the data imbalance, existing methods heavily favour the plane class, even when the results are correct, as seen on the left.
And for rare classes in the right column, they fail to recognise that the samples are important, which leads to the improved performance
On the other hand, our method correctly identifies them as important, by considering also the class difficulty and the class prior.
}
\label{fig:teaser}
\end{figure}

Active learning focuses on efficient labelling of data and has drawn much interest lately~\cite{Yoo19_ac,Sener18_ac,Sinha19_ac}, due to deep learning being attempted at new domains, such as biomedical imaging~\cite{Atzeni18,Yuankai18} and industrial imaging~\cite{zhao2017vision,lin2019automated}, where acquiring data can be costly~\cite{Settles12_ac}.
Even for cases where data is not scarce, the effective usage of data may reduce training time, therefore the computational cost, including carbon foot-prints required to train each model.
There have been various studies based on semi-supervised~\cite{Donahue13b,Kipf16} and unsupervised~\cite{Saito17,Yan17} learning schemes to improve the training efficiency of data.
However, with limited labelling budget, the performance of the studies are significantly worse to the supervised learning with the additionally labelled data~\cite{Rasmus15}.
In other words, their label efficiency could be improved.

Existing methods~\cite{Yoo19_ac,Gal17_ac,Beluch18_ac,Sener18_ac,Sinha19_ac}, regardless of how they are formulated, have a common underlying assumption that all classes are equal -- 
they do not consider that some classes might just be harder to learn compared to others, or some classes might be more prevalent in the dataset than others.
Instead, they focus on, given a data sample, how much error a trained model is expected to make, or the estimated uncertainties~\cite{Yoo19_ac,Gal17_ac}.
These assumptions could be harmful, as in practice, since data is often imbalanced and not all classes are of the same difficulty~\cite{zhu2017synthetic,al2016transfer}.
This can create a bias in the labelled data pool, leading to the trained classifier and active learning methods also being biased in deciding which samples to the label.
As we show in Fig.~\ref{fig:teaser}, this can damage the capabilities of an active learning method significantly, even for typical benchmark datasets~\cite{cifarDataset,stl10dataset}.

In this work, we present a novel formulation for active learning, based on the classical Bayes' rule that allows us to incorporate multiple factors of a classification network together.
Through derivation, we show that the probability of a classifier making mistakes can be decomposed into three terms; i) the probability of misclassification for a given predicted class, ii) the likelihood of a sample given predicted class, and iii) the prior probability of a class being predicted.
In other words, one needs to take into account i) the difficulty of a class, ii) the performance of the classifier and iii) the abundance of a certain class of data holistically when determining the potential of a classification error.
We take all of them into account and choose samples to be labelled by selecting those that have the highest misclassification probability.

While the task is discriminative, our method requires the estimation of likelihood, which could be intractable.
We, therefore, propose to use a Variational Auto Encoder (VAE)~\cite{Kingma14b} to model the lower bound of the likelihood of a sample. 
To make VAE conditioned on a predicted label, a naive way would be applied to train multiple VAEs for each predicted class.
However, this quickly becomes impractical with a large number of classes.
We thus propose to train a single VAE, with regularisation that acts as conditioning based on the predicted label.
To further tie the VAE with the classifier, and for quick training of the VAE, we use the deep feature representations of the classifier as inputs to the VAE.
Being generative, this training of VAE does not involve any labels, and we utilise the vast amount of unlabelled data with their predicted labels while inferring probabilities that are independent of data samples on the labelled ones.

We show empirically in \Section{results} that our method allows a significant leap in performance for the benchmark dataset and the real application dataset, especially when the labelling budget is highly limited.
Furthermore, we posit that considering the prior, that is, the distribution of labels during training, is critical when analysing the uncertainty of estimates.

In summary, our contributions are four-fold:
\begin{itemize}
    \setlength{\itemsep}{0pt}%
    \setlength{\parskip}{0pt}%
    \vspace{-.5em}
    \item we derive a novel formulation for active learning based on the Bayes' rule and posterior probability;
    \item we propose a framework based on VAE that realises this formulation;
    \item we reveal that considering the differences between classes -- abundance and difficulty -- is important;
    \item we outperform the state of the art in the various experiments.
\end{itemize}

\section{Related Works}



\paragraph{Traditional Active Learning.}
Increasing the label efficiency, thus reducing the cost associated with obtaining labels, has been of interest for decades.
Even before deep learning became popular, various methods were suggested towards this goal~\cite{Settles12_ac}.
Methods were proposed to estimate the uncertainty of the unlabelled samples through the probability of prediction~\cite{Lewis94_ac}, the difference between the best prediction and the second one~\cite{Li13_ac,Luo13_ac,Roth06_ac}, or the entropy covering all the possible classes~\cite{Settles08_ac,Joshi09_ac}.
For support vector machine classifiers the methods were suggested to utilise the distance from the decision boundary, for both the classification task~\cite{Li14_ac,Tong01_ac} and the detection task~\cite{Vijayanarasimhan14_ac}.
The algorithms clustering and finding representative samples were also suggested as another way~\cite{Nguyen04_ac,Hasan15_ac,Aodha14_ac}.

Discrete optimisation algorithms have been proposed to consider the relationship between the sampling result and the model performance~\cite{Elhamifar13_ac,Guo10_ac,Yang15_ac}.
In a voting scheme-based algorithm~\cite{McCallumzy14_ac}, multiple different models are trained by the labelled pool, which determines the next queries according to their disagreement.
Despite these efforts, the classical approaches are geared towards simple features and may hold limitations when applying to a large deep network with many nonlinear estimations.

\paragraph{Active Learning for deep networks.}
Active learning algorithms for deep networks can be categorised into uncertainty-based methods and representation-based methods.
The uncertainty-based methods aim to select the uncertain samples from the unlabelled data pool and annotate them to increase the labelled pool~\cite{Yoo19_ac,Gal17_ac,Beluch18_ac}.
Yoo and Kwon~\cite{Yoo19_ac} proposed to use a small auxiliary ``module'' network predicting the training loss of the baseline network that is being trained with the active learning scheme.
They then select the samples that are expected to give high losses.
While their method is similar to ours in that an additional network is trained, as they require a ground-truth loss value while training, the auxiliary network can only be trained with labelled data, creating yet another network that the performance depends on how data is sampled.
In contrast, we train our VAE with unlabelled data since we only rely on the predicted labels from the baseline network during training, and result in stable performance even with few labelled data.
Gal~\etal~\cite{Gal17_ac} proposed a method based on the estimation of sample-wise posterior probability through a Bayesian deep learning~\cite{Gal16} framework.
The method can be implemented simply by locating several dropout layers in a deep network, but this increases training time significantly until the convergence.
Beluch~\etal~\cite{Beluch18_ac} suggest an active sampling method that estimates the disagreement of the prediction by using multiple deep networks.
The downside of their method is that, as they use multiple networks, the memory and the computational requirement increases proportionally.

The representation-based methods target on finding representative samples within the high-dimensional space that deep networks learn~\cite{Sener18_ac,Sinha19_ac}.
Sener and Savarese~\cite{Sener18_ac} proposed the Coreset algorithm that determines representative samples by using the feature maps of the intermediate layers of a deep network, rather than the last layer.
However, the optimisation method in the Coreset algorithm does not scale well as the number of classes, and the number of unlabelled samples grows.
To improve the scalability, Sinha~\etal~\cite{Sinha19_ac} proposed to map the high dimensional feature maps into a lower dimension through adversarial training.
Unfortunately, being based on adversarial training, the method requires a large amount of training data for the mapping to be reliable.

Beyond them, hybrid approaches combine the best of both uncertainty-based and representation-based methods~\cite{Zhou17_ac,Paul17_ac}.
Some works focused on a specific task: for example
person re-identification~\cite{Liu19_ac} and a human pose estimation~\cite{Liu17_ac}.

While our work is most similar to uncertainty-based methods, it falls into neither uncertainty-based nor representation-based methods.
Contrary to the previous uncertainty-based works, we take into account characteristics that are not restricted to a single sample -- we consider the class difficulty and class imbalance.
Also, unlike the representation-based methods, we are not aiming to find representative samples, but a global trend of samples that are predicted to belong to a certain class.


\paragraph{Semi-supervised learning with VAEs.}
As we utilise VAEs~\cite{Kingma14b}, we also briefly review works related to VAEs that associate them with labelled data.
Since VAEs model the likelihood of data, Lee~\etal~\cite{Lee2018adv} used them to identify out-of-distribution samples for each class.
We are loosely inspired by them, as we also use conditioned VAEs.
However, unlike them, we estimate one portion of our conditional probabilities in estimating the label correctness.
M2 VAE models~\cite{kingma2014semi} and Conditional VAEs~\cite{sohn2015learning} have been proposed to model conditional distributions.
They directly add the condition as an additional latent dimension that is trained independently with the other latent dimensions for the reconstruction.
In contrast, we apply conditioning implicitly during training to represent the class information and the feature distribution in the same latent dimensions.
In our early attempts, we were not able to obtain successful modelling for our application with the former.
\section{Methodology}
\label{sec:method}
We first formally describe the active learning problem and detail how we select new samples to be labelled with the estimated correctness of predictions.
We then derive our method via Bayes' rule and describe how our derivation can be implemented in practice through a VAE.
Afterwards, we provide a summary of how we tie every component together as an active learning algorithm and implementation details.

\subsection{Problem formulation}
Active learning can be formulated as a problem of increasing the pool of labelled data at every round by labelling subsets of the unlabelled pool.
Formally, given a pool of data $\calP$, we keep a pool of labelled data $\labeledpool$ and unlabelled data $\unlabeledpool$, such that $\calP = \labeledpool \cup \unlabeledpool$ and $\varnothing = \labeledpool \cap \unlabeledpool$.
Then, for each active learning round $\round$, we select $\selectedpool^{(\round)}$ that is a subset of $\unlabeledpool$ with $\numsample$ samples according to a criterion that defines the active learning method, which is moved from $\unlabeledpool$ to $\labeledpool$.
Thus, $\unlabeledpool^{(\round+1)} = \unlabeledpool^{(\round)} - \selectedpool^{(\round)}$ and $\labeledpool^{(\round+1)} = \labeledpool^{(\round)} + \selectedpool^{(\round)}$.

The core of active learning is how $\selectedpool^{(\round)}$ is selected, which in our case is based on the probability of a model providing a wrong answer for the given data.

\paragraph{Active learning based on the hardest examples.} 
The underlying idea of our method is that when acquiring data with a limited labelling budget, one should acquire those that have the highest probability of making wrong predictions~\cite{Yoo19_ac,Li13_ac,Gal17_ac,Liu19_ac}.
Formally, if we let $y$ denote the real label and $\hy$ the label predicted by a model, we find samples $\bx$ with large
\begin{equation}
    p( y \neq \hy | \bx ).
\end{equation}
Unfortunately, this is not a probability distribution that can be modelled directly, and we, therefore, estimate this probability via Bayes' rule and approximations.

\subsection{Bayesian active learning}
We now derive our Bayesian formulation.
To estimate $p( y \neq \hy | \bx )$, we take an alternative route through Bayes' rule, instead of the direct estimation that could be given by a discriminative model.
We first represent this probability with its complement, which can then be written as the sum of joint probabilities.
We write
\begin{equation}\label{eq:posteriorUncertainty}
    \begin{aligned}
        p(y \neq \hy | \bx)
        = 1 - p(y = \hy | \bx)
        = 1 - \sum_{n=1}^{N_c} p(y_n, \hy_n | \bx)
        ,
    \end{aligned}
\end{equation}
where $N_c$ is the number of classes and $p(y_n)$ and $p(\hy_n)$ is a shorthand for $p(y=n)$ and $p(\hy=n)$, respectively.
Then, each $p(y_n, \hy_n | \bx)$ within the summation can be written through Bayes' rule as following:
\begin{equation}
    \begin{aligned}
        p(y_n, \hy_n | \bx) 
        = \frac{p(y_n | \hy_n, \bx) p(\bx | \hy_n) p(\hy_n)}{\sum_{n=1}^{N_c}{p(\bx | \hy_n)p(\hy_n)}}.
    \end{aligned}
\end{equation}
Here, $p(y_n | \hy_n, \bx)$ corresponds to the probability of the real label being $n$, given that the predicted label is $n$ and the data being $\bx$.
However, this is a probability that cannot be evaluated unless we have the paired true label.
Thus, we instead do an informed guess, by ignoring $\bx$ and approximating with $p(y_n | \hy_n)$.
In other words, we assume that the probability of a model making a mistake is highly related to the label. 
Thus we approximate by writing
\begin{equation}
        p(y_n, \hy_n | \bx) 
        \approx
        \frac{p(y_n | \hy_n) p(\bx | \hy_n) p(\hy_n)}{\sum_{n=1}^{N_c}{p(\bx | \hy_n)p(\hy_n)}}.
\end{equation}
Finally, with \Eq{posteriorUncertainty}, we have
\begin{equation}\label{eq:uncertainty}
    p(y\neq\hy | \bx)
    \approx
    1 - \sum_{n=1}^{N_c}{
        \frac{p(y_n | \hy_n) p(\bx | \hy_n) p(\hy_n)}{\sum_{n=1}^{N_c}{p(\bx | \hy_n)p(\hy_n)}}.
    }
\end{equation}
Note that here, i) $p(y_n | \hy_n)$ is the probability of a model making mistake based on label, ii) $p(\bx | \hy_n)$ is the \emph{likelihood} of a sample given predicted label, iii) $p(\hy_n)$ is the \emph{prior} on the distribution of predicted labels, which represents how imbalanced the predictions of a model are.
As mentioned earlier in \Section{intro}, the likelihood estimation is non-trivial and requires a generative model.
We now detail how we model these three probabilities.

\subsection{Estimating probabilities with regularized VAE} \label{sec:prob}
\begin{figure}[t]
\centering
\includegraphics[width=\linewidth]{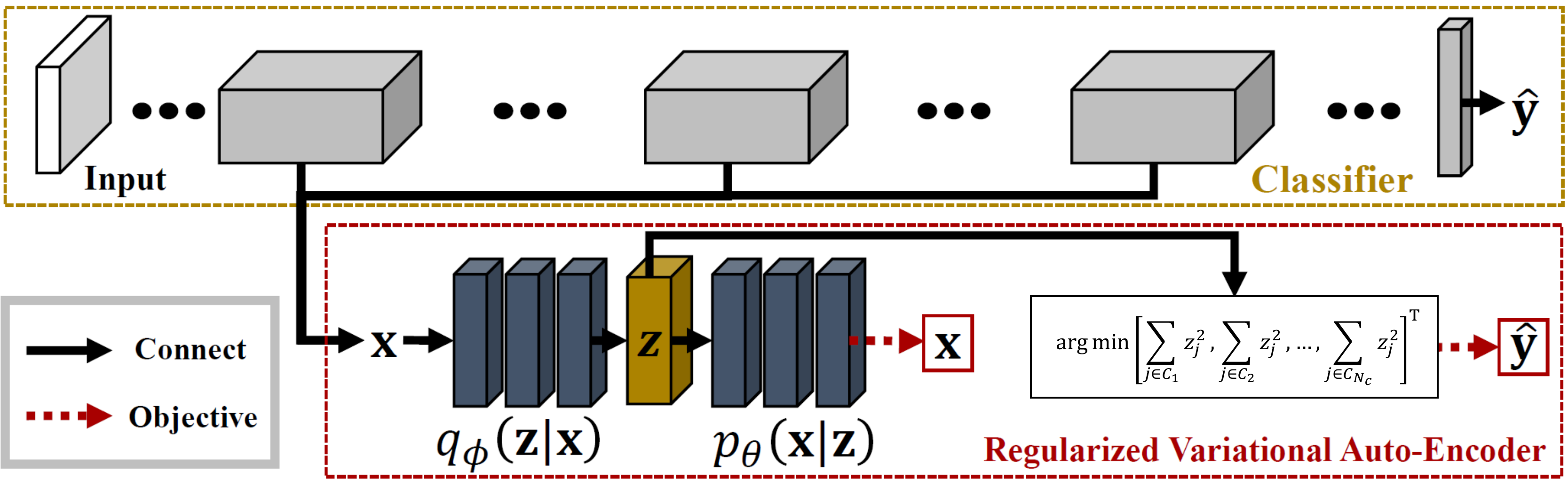}
\caption{
{\bf Framework --}
An illustration of our framework.
We train a Variational Auto Encoder (VAE) that models the deep feature representation of the classifier with unlabelled data.
We then use the VAE to estimate $p(\bx|\hy_n)$, $p(y_n|\hy_n)$, and $p(\hy_n)$, which in turn is used to estimate the probability of labelling error, $p(y\neq\hy|\bx)$.
See \Section{prob} for details.
}
\label{fig:framework}
\end{figure}

To estimate the probabilities, we use a VAE~\cite{Kingma14b}. 
Before we discuss the details, let us first clarify that we train this VAE exclusively in the unlabelled data pool $\unlabeledpool$, and associate it with the behaviour of the discriminative model on unseen data.
We do not use the labelled pool $\labeledpool$, as it is likely that the classifier has overfitted to the data.
Note also that while we explain in terms of $\bx$, in fact, we use the deep feature representations given by the classifier
to tie the VAE more with the classifier and to facilitate training by removing the need of learning the deep features. 
See \Fig{framework} for an illustration of our framework.

We first detail why and how we estimate the likelihood with a VAE and then discuss the other two probabilities.

\paragraph{Likelihood of a sample -- $p(\bx | \hy_n)$.}
Estimating $p(\bx | \hy_n)$ is not straightforward.
A naive idea would be to implement multiple generative models that model $p(\bx)$, each trained with labels predicted to be of a certain class.
However, this becomes quickly impractical as the number of classes grows.
Moreover, estimating $p(\bx)$ can be intractable in practice~\cite{Kingma14b}.

In our work, we use a \emph{single} VAE to estimate the lower bound of $p(\bx | \hy_n)$ for all $\hy_n$.
We use a VAE, as it learns to reconstruct the data sample using the tractable lower bound for $p(\bx)$.
Distinct from existing work, to condition the $p(\bx)$ based on predicted labels, we propose to learn a latent space where the \emph{absence} of parts of the latent embeddings are related to the predicted label.
In other words, this is as if we are training multiple VAEs with shared weights and overlapping latent spaces.
Once such latent embeddings are learned, we compute the lower bound of $p(\bx | \hy_n)$, by simply enforcing the absence manually via masking -- thus selecting a VAE dedicated to a certain predicted class among the multiple virtual VAEs -- and computing $p(\bx)$.
This strategy allows us to have a manageable latent space while still being able to deal with many classes.

In more detail, if we denote 
the $j$-th embedding dimension of VAE as $z_j$ and write $j \in C_n$ to denote dimension $j$ is related to class $n$, we write this absence condition as
\begin{equation}
    \begin{aligned}
    \hy = \argmin_n \left[
        \sum_{j\in C_1} z_j^2, \sum_{j\in C_2} z_j^2, \dots, \sum_{j\in C_{N_c}} z_j^2
    \right]^\top .
    \end{aligned}
    \label{eq:condition}
\end{equation}
Notice how this condition is conceptually similar to disentangled representations~\cite{sohn2015learning,burgess2018understanding}.
In our earlier attempts, we have also tried forming this condition as disentangled representations or enforcing $\sum_{j\in C_n} z_j^2$ to be zero if $\hy_n$, which neither was successful.
We suspect that enforcing such constraints limit the capacity of the latent space and interferes with the training of the VAE too much.
We have also tried other ways of enforcing absence -- using the $\ell-1$ norm or the sigmoid -- but using the square worked best. We provide empirical results in \Section{ablation}.

We enforce this constraint as a form of regularisation. 
Let $\bw=\left[w_1, w_2, \dots, w_n\right]^\top$, where $w_n = \sum_{j\in C_n} z_j^2$, then we form an additional regularisation loss $\calL_\text{Class}$ to be used during training of VAE as
\begin{equation}
    \calL_\text{Class} = \crossentropy\left(
        \softmax\left(-\bw\right), \bhy
    \right),
    \label{eq:l_class}
\end{equation}
where $\crossentropy$ denotes the cross entropy, $\softmax$ is the softmax, and $\bhy$ is the one-hot encoded vector representation of $\hy$.
With this regularisation term, recall from \cite{Kingma14b}, that the training loss for VAEs $\calL_\text{VAE}$ is the inverse of the empirical lower bound (ELBO) of the likelihood, which is defined as 
\begin{equation}\label{eq:likelihoodProb}
    \begin{aligned}
    \calL_\text{VAE}
    = -
    \IE_{\bz\sim q_{\phi}\left(\bz | \bx\right)}\left[
        \log p_{\theta}(\bx |\bz)
    \right]
     + 
    D_{KL}\left(
        q_{\phi}\left(\bz |\bx\right) 
        \parallel 
        p(\bz)
    \right)
    \end{aligned}
    \;,
\end{equation}
where $p_{\theta}(\bx |\bz)$ is the decoder with parameters $\theta$ and $q_{\phi}\left(\bz |\bx\right)$ is the encoder with parameters $\phi$.
Therefore, the total loss to train our VAE is
\begin{equation}\label{eq:vaeLoss}
    \calL = \calL_\text{VAE} + \lambda \calL_\text{Class}
    ,
\end{equation}
where $\lambda$ is a hyperparameter that controls the regularisation strength.
Note that with this loss, the VAE now also tries to mimic the behaviour of the classifier.


Once the VAE is trained, this VAE -- without any conditioning -- is now able to estimate the lower bound of the likelihood of a given data $p(\bx)$~\cite{Kingma14b} by simply computing the inversed value of $\calL_{VAE}$.
Furthermore, by masking the embedding space associated with $\hy$ with zero, we can compute the lower bound of $p(\bx|\hy)$.
To avoid the decoder going too far from the latent embeddings, it was trained for; we use \Eq{condition} to obtain $\hy$, instead of the one from the classifier.

\paragraph{Probability of labelling error -- $p(y_n | \hy_n)$.}
While the labelled pool $\labeledpool$ is the only set of data that we have labels for, as the trained classifier is likely to have overfitted to $\labeledpool$, we cannot use it for modelling this probability.
We, therefore, use the labels given by the VAE for the data samples in the labelled pool $\labeledpool$.
Note that the VAE has never seen these data points during training.
Mathematically, if we denote the $i$-th sample as $\bx^{(i)}$, its label as $y^{(i)}$, the predicted label from \Eq{condition} as $\hy^{(i)}$, and introduce an indicator function $\delta$ that is $1$ if all inputs are equal and $0$ otherwise, we write
\begin{equation}\label{eq:labelUncertaintyProb}
    p(y_n | \hy_n) \approx \frac{ 
        \IE_{\bx\in\labeledpool, \bz\sim q_\phi\left(\bz|\bx\right)}
        \left[
            \delta\left(y^{(i)}, \hy^{(i)}, n\right)
        \right]
    }{
        \IE_{\bx\in\labeledpool, \bz\sim q_\phi\left(\bz|\bx\right)}
        \left[
            \delta\left(\hy^{(i)}, n\right)
        \right]
    }
    .
\end{equation}
Here, we approximate the expectations with Monte Carlo estimates.
Note that we also take the expectation over $\bz$, to take into account the stochastic nature of VAEs.


\paragraph{Prior -- $p(\hy_n)$.}
The prior is also acquired by using the labelled samples included in $\labeledpool$ as this probability should be related to how the classifier is trained.
Same as in the case of $p(y_n | \hy_n)$, we cannot use the classifier predictions for the labelled pool, and we use the predictions from the VAE.
Thus, sharing the notations as in \Eq{labelUncertaintyProb}, we write
\begin{equation}\label{eq:prior}
    p(\hy_n) \approx
        \IE_{\bx\in\labeledpool, \bz\sim q_\phi\left(\bz|\bx\right)}
        \left[
            \delta\left(\hy^{(i)}, n\right)
        \right]
        .
\end{equation}

\subsection{Summary and implementation details}
\label{sec:impl}

\begin{algorithm}
\small
    \label{alg:method}
    \caption{Proposed Method
    }
    Input: $\mathcal{P}_U^{(0)}$, $\mathcal{P}_L^{(0)}$, $N_r$, $\mathcal{M}$\\
    
    $r$ = 0
    \While{not at the maximum round}{
        Train $\mathcal{M}$ using $\mathcal{P}_L^{(r)}$\\
        Freeze $\mathcal{M}$\\
        Train the VAE module using $\mathcal{P}_U^{(0)}$ by \Eq{vaeLoss}\\
        
        Estimate $p(\bx | \hy_n)$ by \Eq{likelihoodProb}\\
        Estimate $p(y_n | \hy_n)$ by \Eq{labelUncertaintyProb}\\
        Estimate $p(\hy_n)$ by \Eq{prior}\\
        Estimate $p(y\neq\hy | \bx)$ by \Eq{posteriorUncertainty}\\
        $\mathcal{X}$ $\leftarrow$ $N_r$ samples with the highest uncertainty\\
        $\mathcal{P}_U^{(r+1)}$ $\leftarrow$ $\mathcal{P}_U^{(r)}$ $-$ $\mathcal{X}$\\
        $\mathcal{P}_L^{(r+1)}$ $\leftarrow$ $\mathcal{P}_L^{(r)}$ $\bigcup$ $\mathcal{X}$\\
        $r$ $\leftarrow$ $r$ $+$ $1$
    }
    \normalsize
\end{algorithm}



We summarise our method in \Algorithm{method}. 
For each Active Learning round $\round$, we train a classifier $\calM^{(\round)}$ with the labelled pool $\labeledpool^{(\round)}$.
We then freeze the weights of the classifier, and train our VAE -- $p^{(\round)}_\theta\left(\bx|\bz\right)$ and $q^{(\round)}_\phi\left(\bz|\bx\right)$ -- with  the unlabelled pool $\unlabeledpool^{(\round)}$.
We then estimate the three probabilities, which we use to construct the final estimate of $p(y\neq\hy|\bx)$, and sample those from $\unlabeledpool^{(\round)}$ that have the highest value.

As noted earlier, we use the deep features of the baseline network (the classifier) as input to the VAE.
Specifically, we extract deep features from the specific four layers of baseline.
If the feature representations are from fully-connected layers, we use it as is.
If it is from a convolutional layer, thus a feature map, we first apply global average pooling.
We then apply batch normalisation to each feature, so that they are roughly in the same range, and feed it to a fully connected layer with 128 neurons and Sigmoid activation.
Finally, the outputs from these fully-connected layers are concatenated to form a $512\times1$ vector and given as input to the VAE.

For the architecture of VAEs, we opt for a simple one as the deep features already have abstracted context. 
We apply four layers of fully connected layers with again 128 neurons and ReLU activations, with the exception of the layer that outputs the latent embeddings.
For this layer, we use $10\times N_c$ number of neurons and dedicate 10 dimensions per each class.


We implement our method with PyTorch~\cite{paszke2017automatic}.
To train our VAE, we set $\lambda{=}0.005$ in Eq.~\eqref{eq:vaeLoss}, for all tasks.
We use Adam optimiser~\cite{Kingma15} with a learning rate of $10^{-4}$ and default parameters.
We train for 20 epochs.
To remove randomness from experiments, we perform our training multiple times with different initial conditions.
We report both the average and the standard deviation of our experiments.
For inference, to perform the Monte Carlo estimation in Eqns.~\eqref{eq:likelihoodProb}, \eqref{eq:labelUncertaintyProb}, and \eqref{eq:prior}, we iterate the probabilistic inference 100 times for every sample.


\section{Experimental Results}
\label{sec:results}

We first introduce the dataset, then the experimental setup, including the classification network and the compared baselines. 
We then report our results and discuss the effect of the individual probability terms through an ablation study.



\subsection{Dataset}

To validate our experiments we apply our method to the standard \textit{CIFAR-10} and \textit{CIFAR-100}~\cite{cifarDataset} datasets as well as the Northeastern University surface defect classification dataset (NEU)~\cite{neudataset}.
We use the CIFAR datasets to be comparable with recent Active Learning studies~\cite{Sinha19_ac}, and NEU to demonstrate that class imbalance is important in real-world applications.
In more detail, the \textit{CIFAR-10} dataset contains 60,000 images sized by $32\times 32\times 3$ with 10 classes.
The \textit{CIFAR-100} dataset also consists of 60,000 images of the $32\times 32\times 3$ size with 100 classes.
Both the \textit{CIFAR-10} and the \textit{CIFAR-100} datasets are well balanced -- all included images are assigned exactly one class among the 10 or the 100, and the number of instances per each class is equal.
The datasets come with a pre-determined training and evaluation split of 50,000 and 10,000 images, respectively.
The goal of the NEU dataset is to classify the 9 defect types on steels.
The main feature of this dataset is that it is heavily imbalanced -- the number of instances per class varies from 200 to 1589.
In total, the NEU dataset contains 7226 defect images of size $64\times 64$ pixel.
This dataset does not provide pre-determined splits, and we thus randomly reserve 20\% of the dataset for evaluation.
This results in 5778 training samples and 1448 validation samples.

\paragraph{Introducing synthetic class imbalance.}
To demonstrate the importance of considering class-wise probabilities, we build additional variants of \textit{CIFAR} datasets by removing a part of the samples in \textit{CIFAR-10} and \textit{CIFAR-100}.
First, we build four datasets that have dominant classes within them -- we hereafter refer to them as \textit{dominant} datasets.
A real-world example would be when scraping data from the internet -- there will be many images of people, cats, and dogs, but not so many of platypus.
The first three dominant datasets are built from \textit{CIFAR-10} by randomly removing 90\% samples of every category except for the $\{1, 5, 10\}^{th}$ classes, respectively.
For \textit{CIFAR-100}, as there are many classes, there are only a few instances each -- 500 for each class in the training set.
We, therefore, build the last dominant dataset by removing 40\% of the original samples for all categories other than the middle ones from $45^{th}$ class to $55^{th}$ class, from the \textit{CIFAR-100} dataset.
We denote these dominant datasets as \textit{CIFAR-10$^{+[1]}$}, \textit{CIFAR-10$^{+[5]}$}, \textit{CIFAR-10$^{+[10]}$}, and \textit{CIFAR-100$^{+[45:55]}$}, respectively.

We further consider the case when some samples are rare.
This would be, for example, cases where there are rare events, such as accidents or defects that need to be discovered.
Similar to the \textit{dominant} datasets, we build \textit{rare} datasets by taking certain classes away from \textit{CIFAR} datasets.
Specifically, we use three variants, where we remove 90\% of the samples that correspond to the $\{1, 5, 10\}^{th}$ classes.
We denote these as \textit{CIFAR-10$^{-[1]}$}, \textit{CIFAR-10$^{-[5]}$}, and \textit{CIFAR-10$^{-[10]}$}, respectively.

With these datasets, we run different active learning setups, based on the size of the dataset.
For NEU, we run five active learning rounds, where each round samples 250 samples.
For CIFAR-10 based ones, we run six active learning round, which 500 samples each.
For CIFAR-100, we run five rounds with 1000 samples each.





\subsection{Experimental setup}

\paragraph{The baseline classification network. }
As the baseline classifier, we utilise ResNet-18~\cite{He16}.
This network is composed of an initial simple convolution layer, followed by four basic residual blocks.
As discussed previously in \Section{impl}, the output feature maps from these four residual blocks are used to form the input to our VAE.
At every Active Learning round, we train the network for 200 epochs, with a batch size of 128.
We train with a typical setup for classification on CIFAR-10: SGD with the momentum of 0.9, weight decay of 0.0005, the learning rate is also initialised as 0.1 and decreased to 0.01 after 160 epochs.
We repeat the same experiments with the different random seeds three times to remove the randomness from our experiments.

\paragraph{The competitors.}
To demonstrate the effectiveness of the proposed algorithm, we compare our method with the state-of-the-art for active learning.
We consider
\textit{VAAL}~\cite{Sinha19_ac}, \textit{MC-DROPOUT}~\cite{Gal17_ac}, and \textit{Core-set}~\cite{Sener18_ac}.
We utilise the author's implementation for \textit{VAAL} and \textit{Core-set}.
For \textit{MC-DROPOUT} we integrate the author's implementation to our framework to utilise our baseline classification network -- this method is architecture dependant.
In more detail, we utilise the authors' code for uncertainty estimation and embed it into ours.
Among the multiple methods to estimate the uncertainty in \textit{MC-DROPOUT}, we use the Bayesian Active Learning by Disagreement (BALD)~\cite{houlsby2011bayesian} as the estimation method -- it showed the best generality in our experiments.
For the dropout layers of \textit{MC-DROPOUT}, we place them before each the residual blocks of the ResNet and fix the dropout ratio to 0.25. 
As suggested by the authors~\cite{Gal17_ac}, we further use 100 samples to estimate the uncertainty.
In addition, 
we also consider uniform random sampling as a baseline.
Finally, for a fair comparison, after new samples are extracted by these methods, we train the networks with the same random seed to further exclude any random surprises.


\subsection{Comparison results}
\begin{figure*}[t]
\centering
\includegraphics[width=\linewidth]{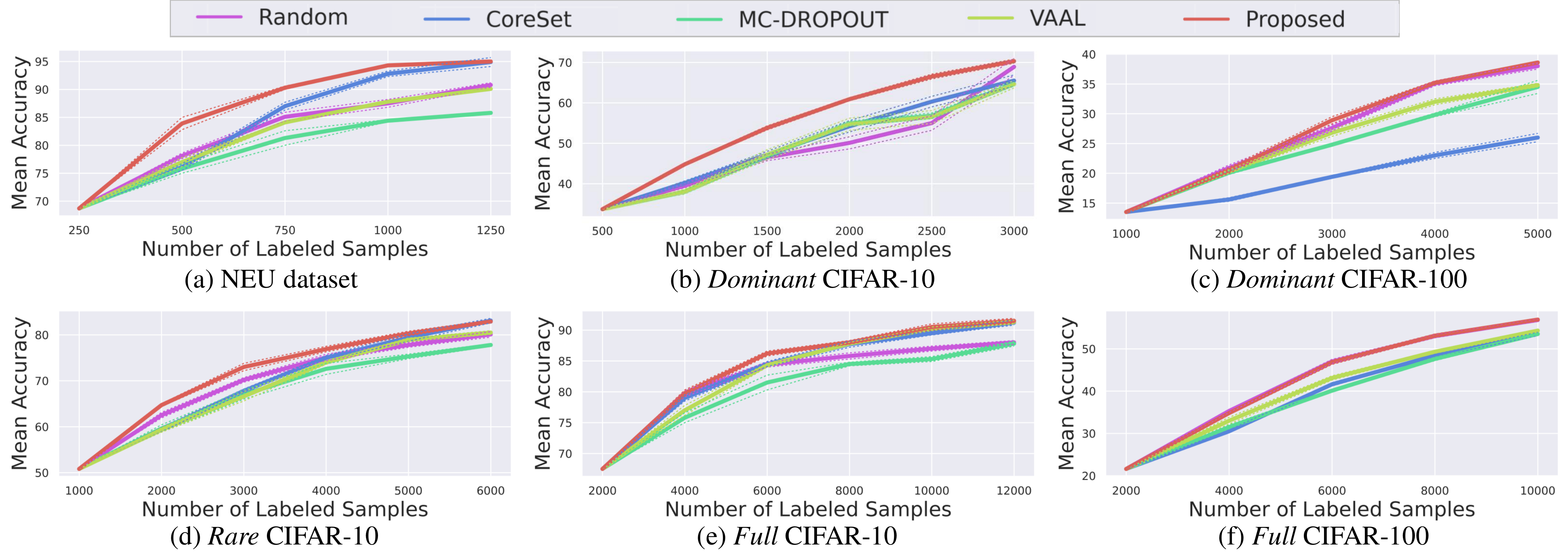}
\caption{Results for the dominant, rare, and full datasets.}
\vspace{-4mm}
\label{fig:allDataset}
\end{figure*}

\paragraph{NEU dataset -- a real-world imbalanced dataset.}
We first compare different methods on the NEU dataset, which is a typical case of an imbalanced real-world dataset.
We report the results in \Fig{allDataset}(a).
As shown, the proposed method delivers improved performance, especially for the early iterations.
Methods here mostly converge after 5 rounds, as the dataset is relatively small.
However, judging by the gap in early iterations, it could be expected that a similar gap would exist in a larger dataset.
Interestingly, for this small dataset, VAAL performs worse than simple random selection. 
This is because the method requests a lot of labels to \emph{bake in}.



\paragraph{\textit{Dominant} datasets.}
We now consider the dominant variants.
In the dominant situation, a large part of the entire training samples is dedicated to a few specific categories -- in the case of \textit{CIFAR-10}, just one.
In \Fig{allDataset}(b), we average the results from \textit{CIFAR-10}$^{+[1]}$, \textit{CIFAR-10}$^{+[5]}$, and \textit{CIFAR-10}$^{+[10]}$ that we run three times each -- a total of nine experiments.
As shown, the proposed method outperforms the compared methods by a significant margin.
This gap comes from the fact that the dataset is biased, and this causes the active learning methods also to become biased.
However, our method successfully mitigates this class imbalance and provides improved performance.

A similar phenomenon happens when there are more classes. 
In \Fig{allDataset}(c), we report results for \textit{CIFAR-100}$^{+[45:55]}$.
Here, it is worth noting that all compared methods perform worse than \textit{Random}.
This is because the dataset has 100 classes, and it is easy to start ignoring a certain class.
Nonetheless, our method is the only method that provides improved label efficiency.
To mitigate this, existing works have only studied when there is a sufficient number of labels from the beginning -- for example in \textit{VAAL}~\cite{Sinha19_ac}, 10\% of the entire set.
However, this greatly limits the applicability of active learning methods, as 10\% is sometimes already too much to afford.


\paragraph{\textit{Rare} datasets.}
We further test the case when some classes are rare.
In \Fig{allDataset}(d), we show the average performance of each method on the three rare datasets -- \textit{CIFAR-10}$^{-[1]}$, \textit{CIFAR-10}$^{-[5]}$, and \textit{CIFAR-10}$^{-[10]}$.
Similar to the dominant case, our method shows the best performance among the compared methods.
Interestingly, in this case, similar to the results with \textit{CIFAR-100}$^{+[45:55]}$, existing methods perform worse than the random baseline.
Again, this is because existing methods can break down easily without enough labelled samples from the beginning. 
This is a limitation our method does not have.



\paragraph{On the full dataset.}
For completeness, we further study how methods perform with the full \textit{CIFAR-10} and \textit{CIFAR-100} datasets.
However note that, as we demonstrated previously, these datasets are with perfect data balance, which is not the case for real-world datasets.
We report a summary of these results in Figures~\Fig{allDataset}(e)~and~(f).
Our method performs comparable to existing methods for \textit{CIFAR-10} and outperforms existing methods for \textit{CIFAR-100}.
However, experiment for \textit{CIFAR-100} shows a limitation of active learning methods including ours, that when there are too many classes and very few examples, their performances are close to random, and sometimes even worse. 
Nonetheless, compared to existing methods, our method performs best, even in this extreme situation.


\paragraph{Additional results.}
In the supplementary document, we further provide the original plots for all experiments before averaging.
From them, we can confirm that the proposed method outperforms existing methods consistently for various situations.


\begin{table}[t]
    \begin{center}
    \resizebox{0.80\linewidth}{!}{%
        \begin{tabular} {cc||cc|cc}
        \toprule
             \multirow{2}{1cm}{$p(\hy)$} & \multirow{2}{1cm}{$p(y|\hy)$} & \multicolumn{2}{c}{\textit{CIFAR-10}} & 
             \multicolumn{2}{c}{\textit{CIFAR-10}$^{+[1]}$}\\
             &  & avg. & final & avg. & final\\
            \midrule
            - & - & 82.56\% & 88.66\% & 48.02\% & 60.13\%\\
            \checkmark & - & 82.91\% & 90.68\% & \textbf{54.23}\% & 70.25\%\\
            - & \checkmark & 82.71\% & 90.51\% & 51.86\% & 65.98\%\\
            \checkmark & \checkmark & \textbf{83.36}\% & \textbf{91.12}\% & 54.16\% & \textbf{70.35}\%\\
            \bottomrule
        \end{tabular}
        }
    \end{center}
    \vspace{-1mm}
    \caption{Validity of the prior $p(\hy)$ and label difficulty $p(y|\hy)$.}
    \label{tbl:ablation_prior}
\end{table}
\begin{table}[t]
    \begin{center}
    \resizebox{1.00\linewidth}{!}{%
        \begin{tabular} {c||cc|cc|cc}
        \toprule
            \multirow{2}{1cm}{$w_n$Type} & \multicolumn{2}{c}{\textit{CIFAR-10}$^{+[1]}$} & \multicolumn{2}{c}{\textit{CIFAR-10}$^{+[5]}$} & \multicolumn{2}{c}{\textit{CIFAR-10}$^{+[10]}$}\\
            & avg. & final & avg. & final & avg. & final\\
            \midrule
            $Sigmoid(z_j)$ & 39.88\% & 50.66\% & 41.28\% & 49.84\% & 40.88\% & 50.10\%\\
            $|z_j|$ & 43.03\% & 59.03\% & 44.98\% & 57.92\% & 46.80\% & 59.80\%\\
            $z_j^2$ & \textbf{54.16}\% & \textbf{70.35}\% & \textbf{55.18}\% & \textbf{71.85}\% & \textbf{55.82}\% & \textbf{71.13}\%\\
            \bottomrule
        \end{tabular}
        }
    \end{center}
    \vspace{-1mm}
    \caption{Different ways of enforcing constraint on the latent space.}
    \label{tbl:ablation_cond}
\end{table}

\subsection{Ablation study}

\label{sec:ablation}

\subsubsection{Effectiveness of $p(\hy)$ and $p(y_n|\hy)$.}
To validate their effectiveness, we perform an ablation study by excluding one or both of the prior ($p(\hy)$) and the label difficulty ($p(y|\hy)$) -- we artificially set either to 1.
We use \textit{CIFAR-10} and \textit{CIFAR-10}$^{+[1]}$ for these experiments to remove randomness.
To avoid tuning on the test set, we use the validation splits for these experiments.
We take the average performance over all active learning rounds and also the average final performance.
We summarise the results in \Table{ablation_prior}.
We report both the average performance over all six active learning rounds (avg.) and the performance of the final round (final).
We run each experiment three times and report the average.
As shown, all terms contribute to performance improvement.
We observe that all terms contribute to providing the best performance.
Among the two terms, the prior ($p(\hy)$) provides more gain compared to the label difficulty ($p(y|\hy)$), demonstrating that it is critical to take data imbalance into account.
We show an example of the two terms in action in \Fig{qual_dominant}.
Our method balances the training data even when the dataset is imbalanced.

\begin{figure}[t]
    \input{figs/qual_dominant/item.tex}
\end{figure}



\subsubsection{Other choices for $w_n$}
To motivate our design decision, we further build two variants of our method, where we replace $w_n$ in \Eq{condition}, either by a Sigmoid function ($Sigmoid(z_j)$) or a $\ell1$-norm ($|z_j|$).
We summarise the results in \Table{ablation_cond}.
We observe that $\ell2$-norm outperforms the other two regularisation types for all compared cases.
This is unsurprising, considering that in the case of $Sigmoid(z_j)$, it forces the latent embedding $z_j$ to have extreme values, thus conflicting too much with the Gaussian distribution assumption that VAE aims to satisfy. 
This leads to the worst performance among the three types that we tried.
In case of $|z_j|$, it would not suffer from this problem but would create constant gradients that are irrelevant to the magnitude of $z_j$, thus making it hard to enforce absence.
Our choice, $\ell2$-norm, on the other hand, does not suffer from these shortcomings, becoming a natural choice for enforcing absence.


\begin{figure}[t]
\centering
\subfigure[Varying budget sizes.]{\includegraphics[width=0.49\linewidth]{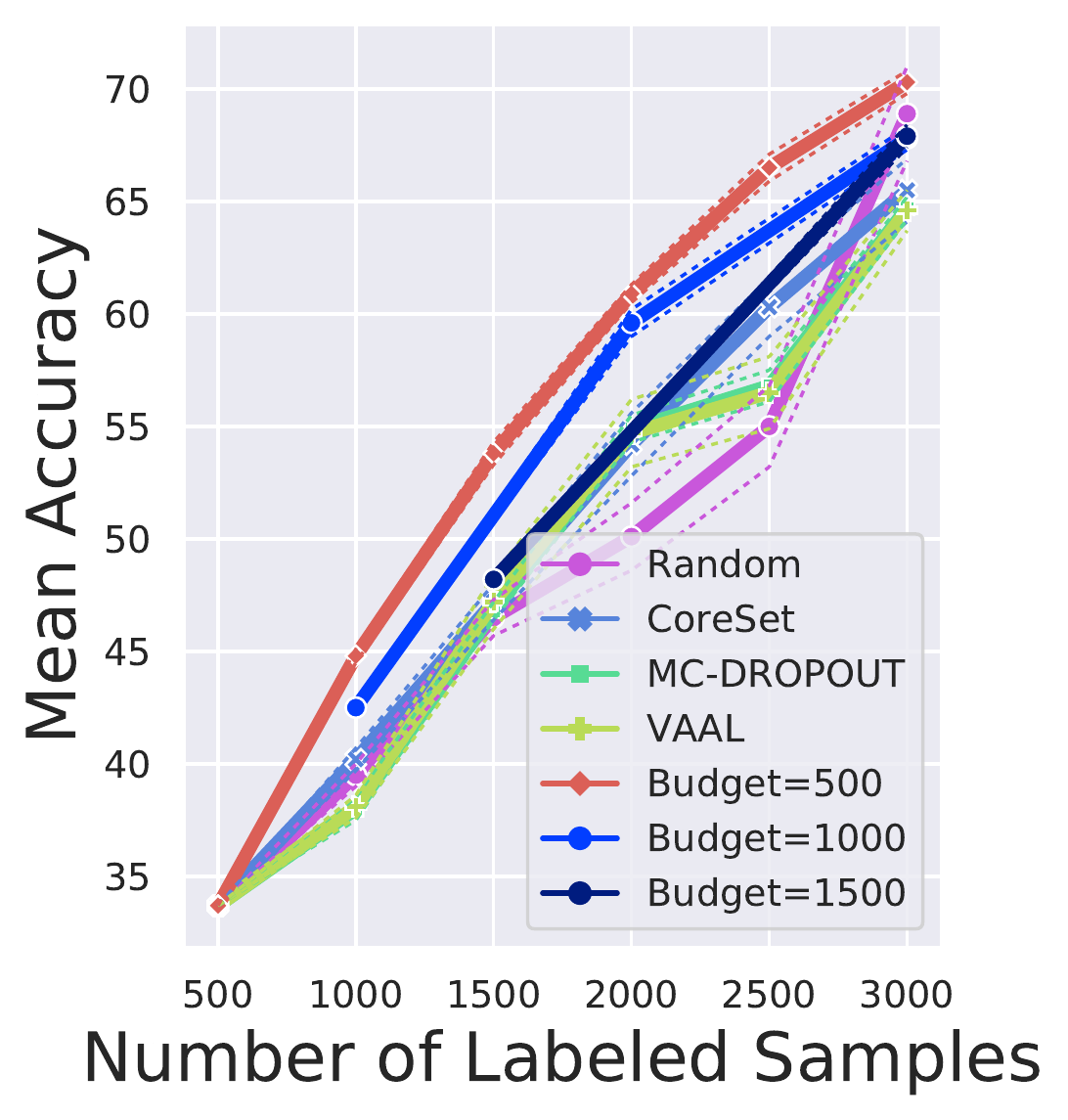}}
\subfigure[Varying $\lambda$.]{\includegraphics[width=0.49\linewidth]{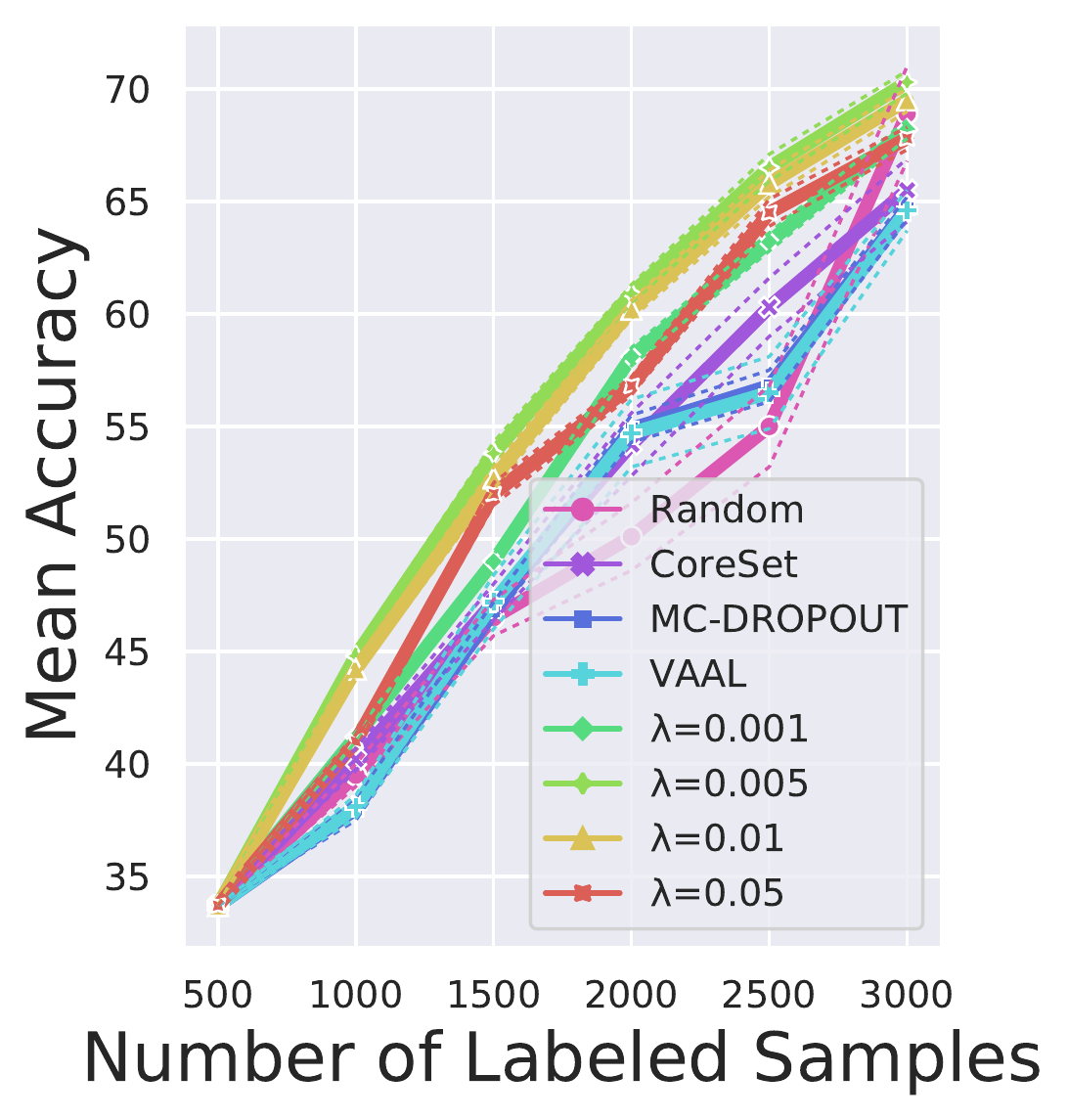}}
\caption{Results of ablation tests on \textit{dominant} CIFAR-10.}
\label{fig:ablation_graph}
\end{figure}

\subsubsection{Robustness for various budgets and $\lambda$ values}
To analyse the sensitivity of the proposed framework against the user-defined parameters, we report the result of our method with various sampling budgets and $\lambda$ values in \Eq{vaeLoss}: see~\Fig{ablation_graph}(a).
Regardless of the choice of the budget size, our method outperforms all compared methods.
Smaller budget size tends to allow methods to react faster to the training outcomes and increase the effectiveness of active learning.

In \Fig{ablation_graph}(b) we report the performance of our method with varying $\lambda$.
We test with various $\lambda$, which do affect the performance of the proposed method to some degree.
However, as before in the case of the budget, our method outperforms all compared methods regardless of the choice of $\lambda$.
This further hints that the two-loss terms that we train for in \Eq{vaeLoss} do not compete.
Note that $\lambda=0$ completely disables $p(\hy)$ and $p(y|\hy)$, as $\hy$ estimation becomes meaningless, which causes the method to perform significantly worse as shown earlier in \Table{ablation_prior}, and we have therefore left it out in this figure to reduce cluster; see supplementary appendix for full results.
 


\section{Conclusion}

We have proposed a novel active learning method that incorporates class imbalance and label difficulty.
Our method was derived through the Bayes' rule, which results in three types of probabilities -- data likelihood, label prior, label difficulty -- being considered together.
We implement our method via a VAE, that is regularised to behave as a conditional VAE.
We have shown that this creates a significant difference for a real-world dataset that exhibits data imbalance, as well as in cases when data imbalance is introduced to CIFAR-10 and CIFAR-100 datasets.

While we limit our experiments to classification in this work, our method is application agnostic. In the future, we plan to extend our work to other discriminative tasks, for example, object detection and segmentation.

\section*{Acknowledgements}
This work was supported by the Natural Sciences and Engineering Research Council of Canada (NSERC) Discovery Grant and Compute Canada.

{\small
    \balance
    \bibliographystyle{ieee_fullname}
    \bibliography{string,vision,optim,learning,kwang_paper,egbib,active_learning,etc,biomed}
}

\clearpage
\nobalance
\twocolumn[
\centering
\Large
\textbf{VaB-AL: Incorporating Class Imbalance and Difficulty \\ with Variational Bayes for Active Learning} \\
\vspace{0.5em}Supplementary Material \\
\vspace{1.0em}
]
\appendix

\section{Detailed results}
Here, we show results that were omitted from the main paper due to spatial constraints, including the raw experimental results.

\subsection{With a different classifier}
\begin{figure}[h]
\centering
\includegraphics[width=\linewidth]{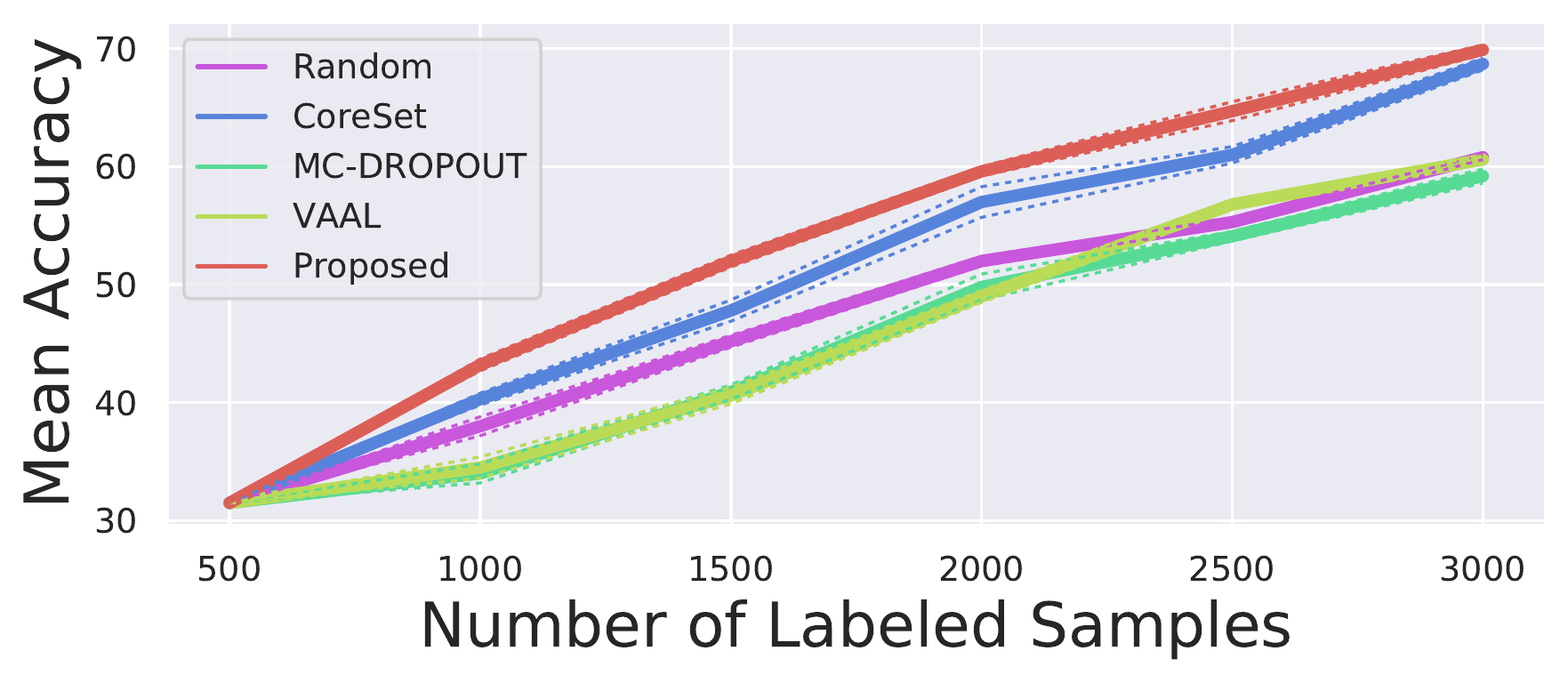}
\caption{Results for CIFAR-10$^{+[1]}$ with ResNet-34}
\label{fig:dominantCIFAR10_resnet34}
\end{figure}
Our method also works well with more complex networks -- ResNet-34.
In \Fig{dominantCIFAR10_resnet34}, we replace the classifier network with ResNet-34 and repeat CIFAR-10$^{+[1]}$ experiment in \Section{results}.
As shown,
our method outperforms the state of the art even with a different classifier network.

\subsection{How much are we focusing on rare classes?}
\begin{figure}[!h]
\centering
\subfigure[CIFAR-10$^{+[1]}$]{\includegraphics[width=\linewidth]{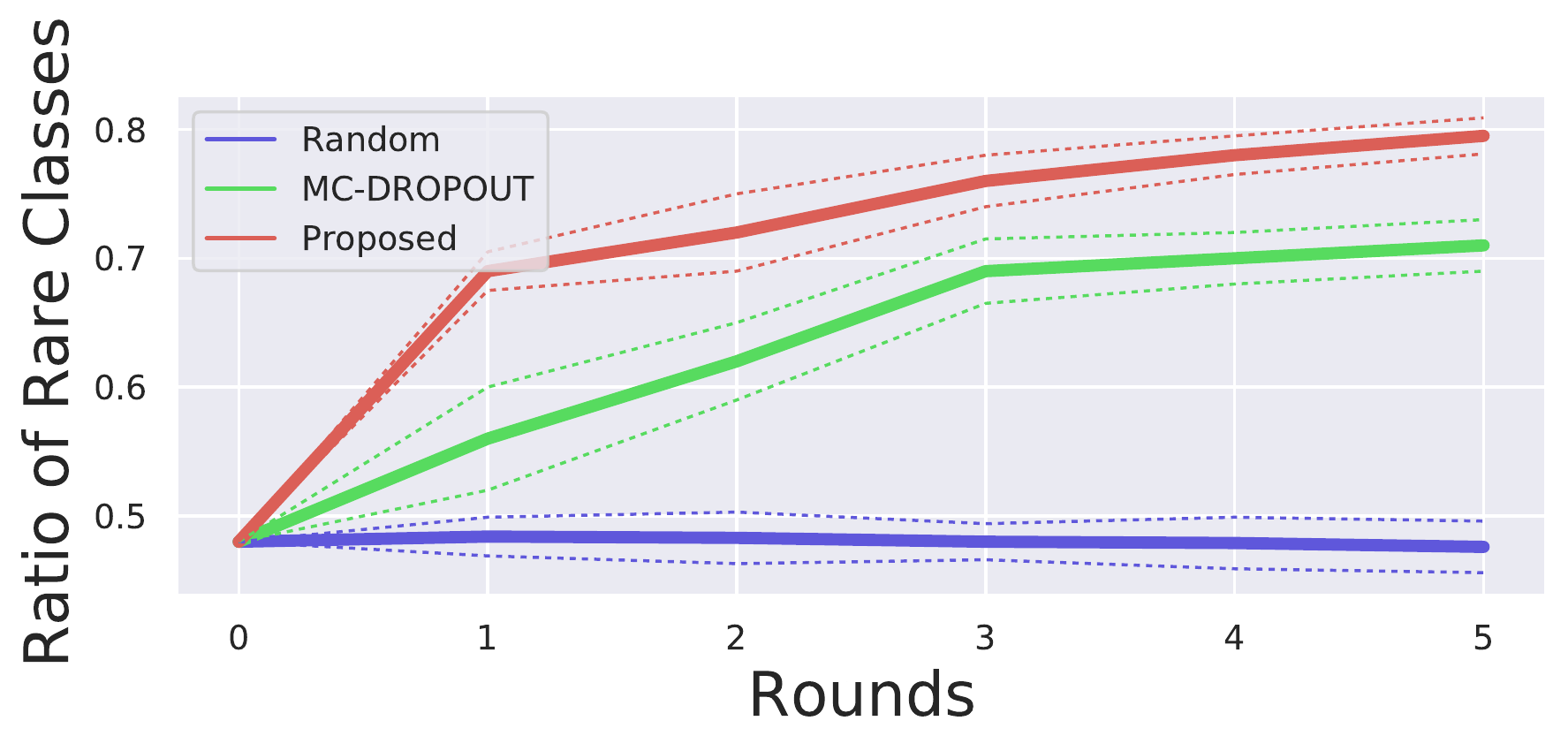}}
\subfigure[CIFAR-10$^{+[5]}$]{\includegraphics[width=\linewidth]{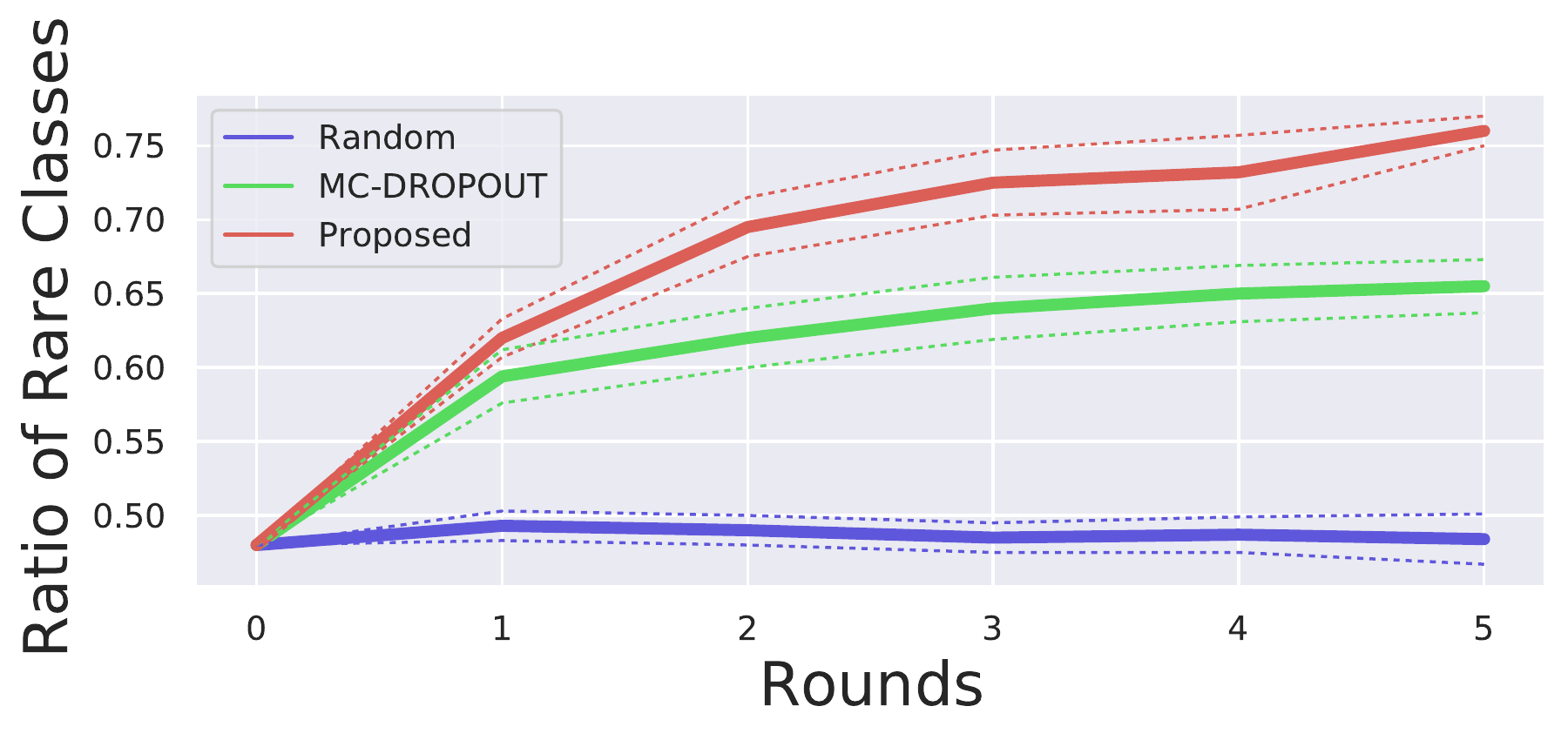}}
\subfigure[CIFAR-10$^{+[10]}$]{\includegraphics[width=\linewidth]{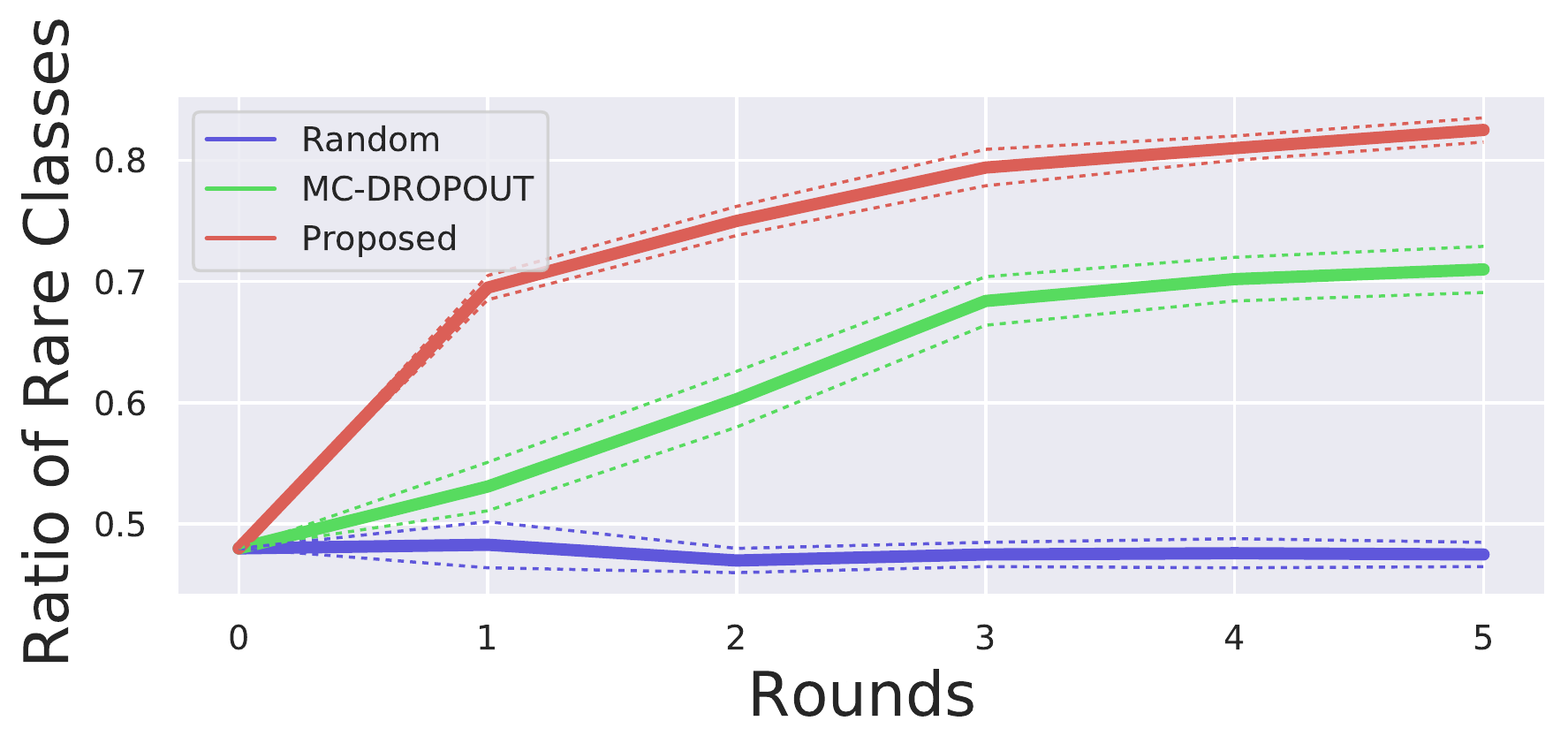}}
\caption{Sample ratio of the rare classes for dominant datasets}
\label{fig:dominantCIFAR10_sampleratio}
\end{figure}

When rare classes are present, our method focuses on those, as the prior probability drives towards a balanced training set.
In Fig.~\ref{fig:dominantCIFAR10_sampleratio}, we report the ratio of rare classes in the labelled pool for CIFAR-10$^{+[1]}$, CIFAR-10$^{+[5]}$, and CIFAR-10$^{+[10]}$.
As shown, the ratio of rare classes increases, showing that our method indeed creates a balanced training set, which eventually leads to a better classifier as reported in \Section{results}.
Note that MC-DROPOUT also tends to deliver similar results, but not as fast as our method. As a result, this contributes to performance improvement in terms of classification accuracy.



\begin{figure*}[h!]
\centering
\includegraphics[width=0.90\linewidth]{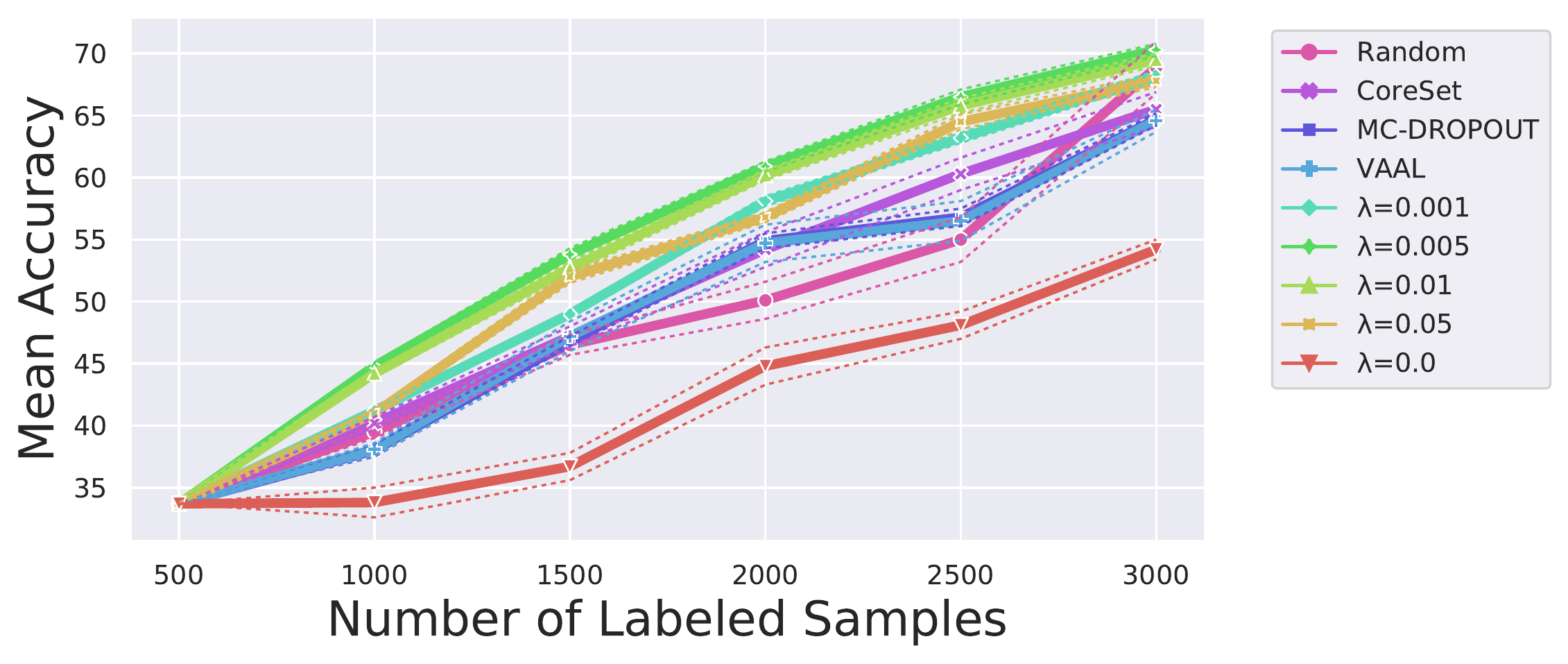}
\caption{Results with varying $\lambda$ on \textit{dominant} CIFAR-10.}
\label{fig:ablation_graph_lambda_full}
\end{figure*}

\subsection{Detailed results with various budget sizes}
\begin{figure*}[h!]
\centering
\includegraphics[width=\linewidth]{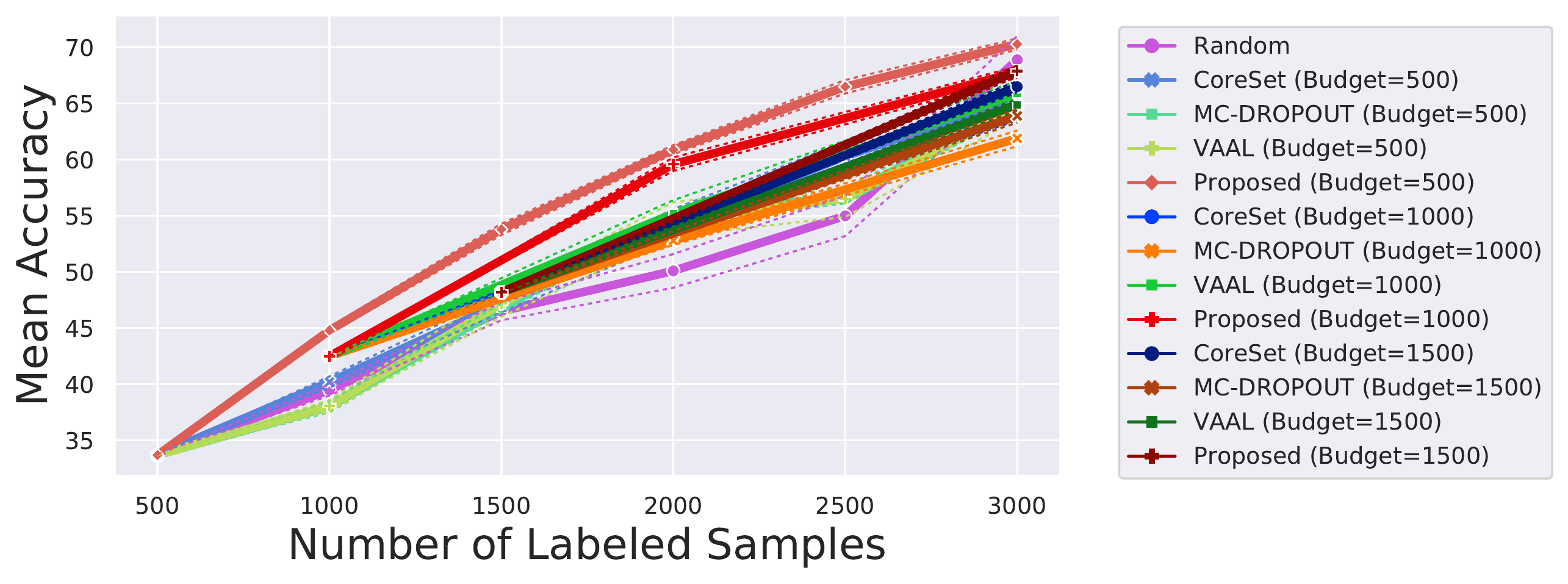}
\caption{Results of ablation tests with various budget sizes on \textit{dominant} CIFAR-10.}
\label{fig:ablation_graph_budget_full}
\end{figure*}

In Fig.~\ref{fig:ablation_graph_budget_full}, we present the entire results of comparisons with various budget sizes.
Since smaller budget size tends to allow methods to react faster to the newly labelled training samples, the final performance of each method conventionally reduces as its budget size increases.
According to the results, our method outperforms all the compared methods, even with various budget sizes.

\subsection{Detailed results with various $\lambda$ value}
In Fig.~\ref{fig:ablation_graph_lambda_full}, we report the performance of our method with various $\lambda$ including $\lambda=0$ that was omitted in Fig.~\ref{fig:ablation_graph}(b).
In contrast to the outperforming performance with other $\lambda$ values, when $\lambda=0$ the performance dramatically drops.
Since $\lambda=0$ means that the class-wise condition in the latent features of VAE is entirely ignored, it becomes infeasible to correctly estimate the three probability terms in Eq.~\ref{eq:uncertainty} by using the VAE.
Therefore, the samples to be labelled at every stage are extracted without considering the label prediction, which results in worse performance even than the \textit{Random} scheme.

\section{Individual results}
\begin{figure*}[!b]
\centering
\subfigure[Results for CIFAR-10$^{+[1]}$]{\includegraphics[width=0.32\linewidth]{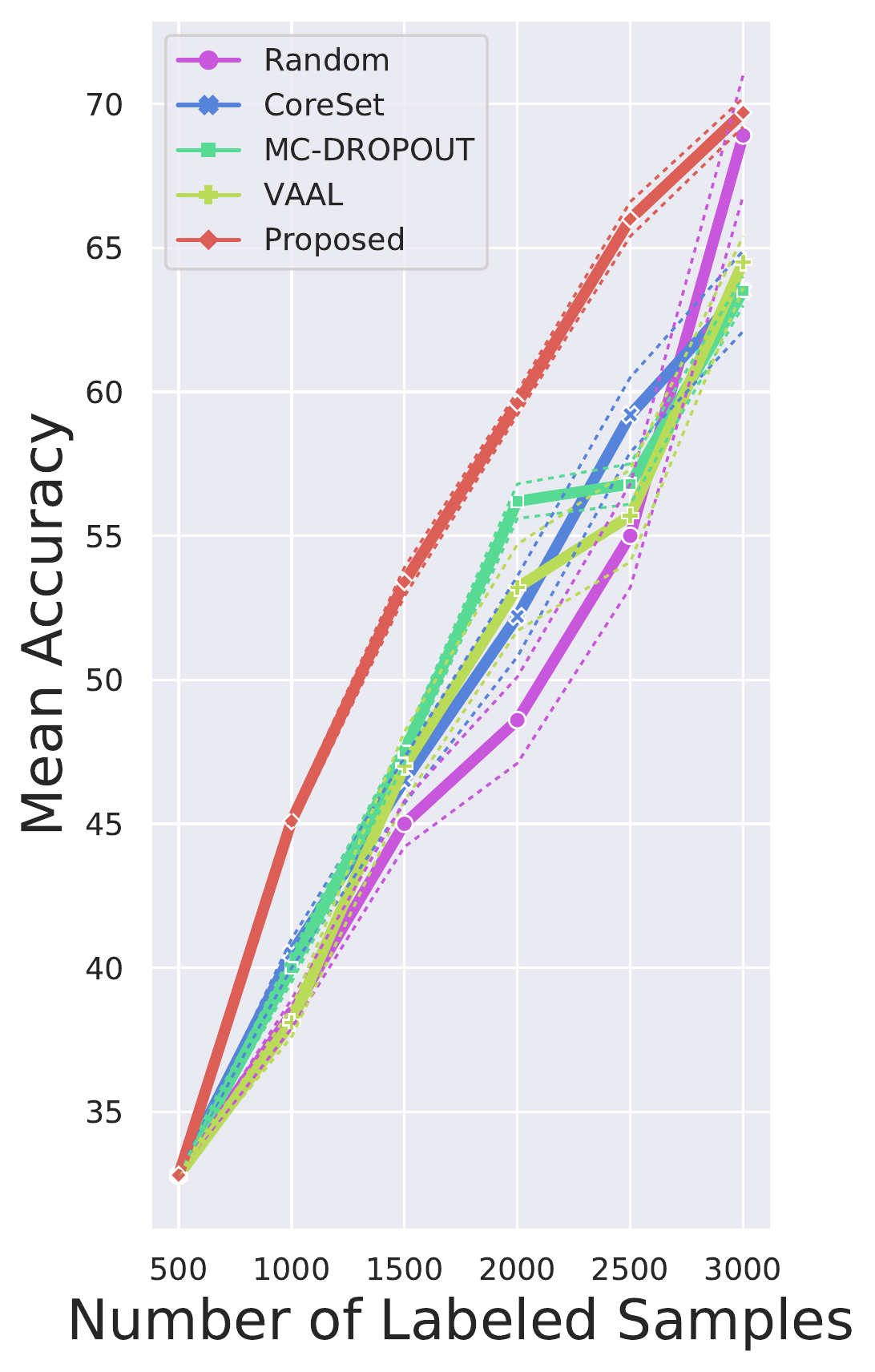}}
\subfigure[Results for CIFAR-10$^{+[5]}$]{\includegraphics[width=0.32\linewidth]{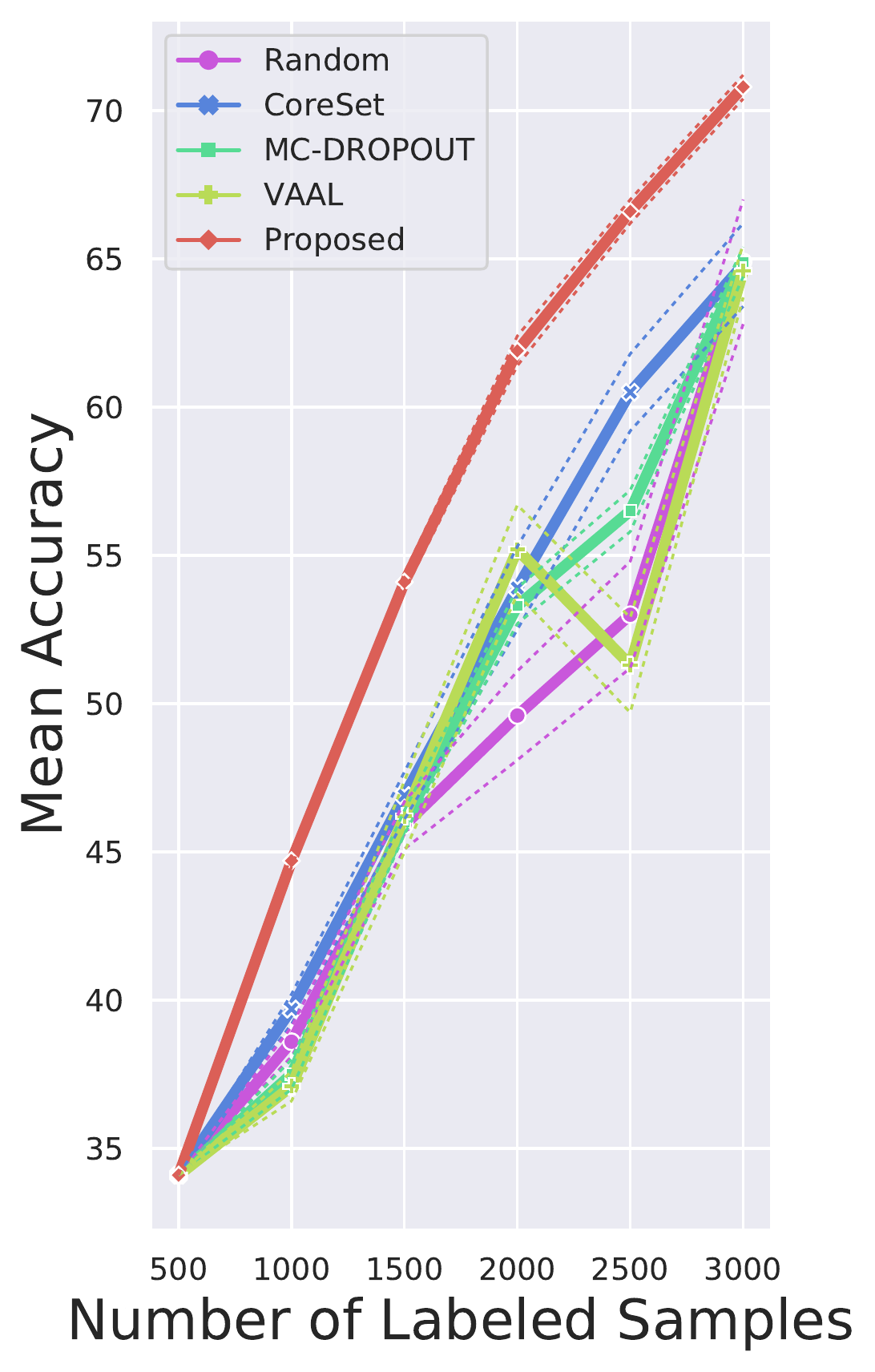}}
\subfigure[Results for CIFAR-10$^{+[10]}$]{\includegraphics[width=0.32\linewidth]{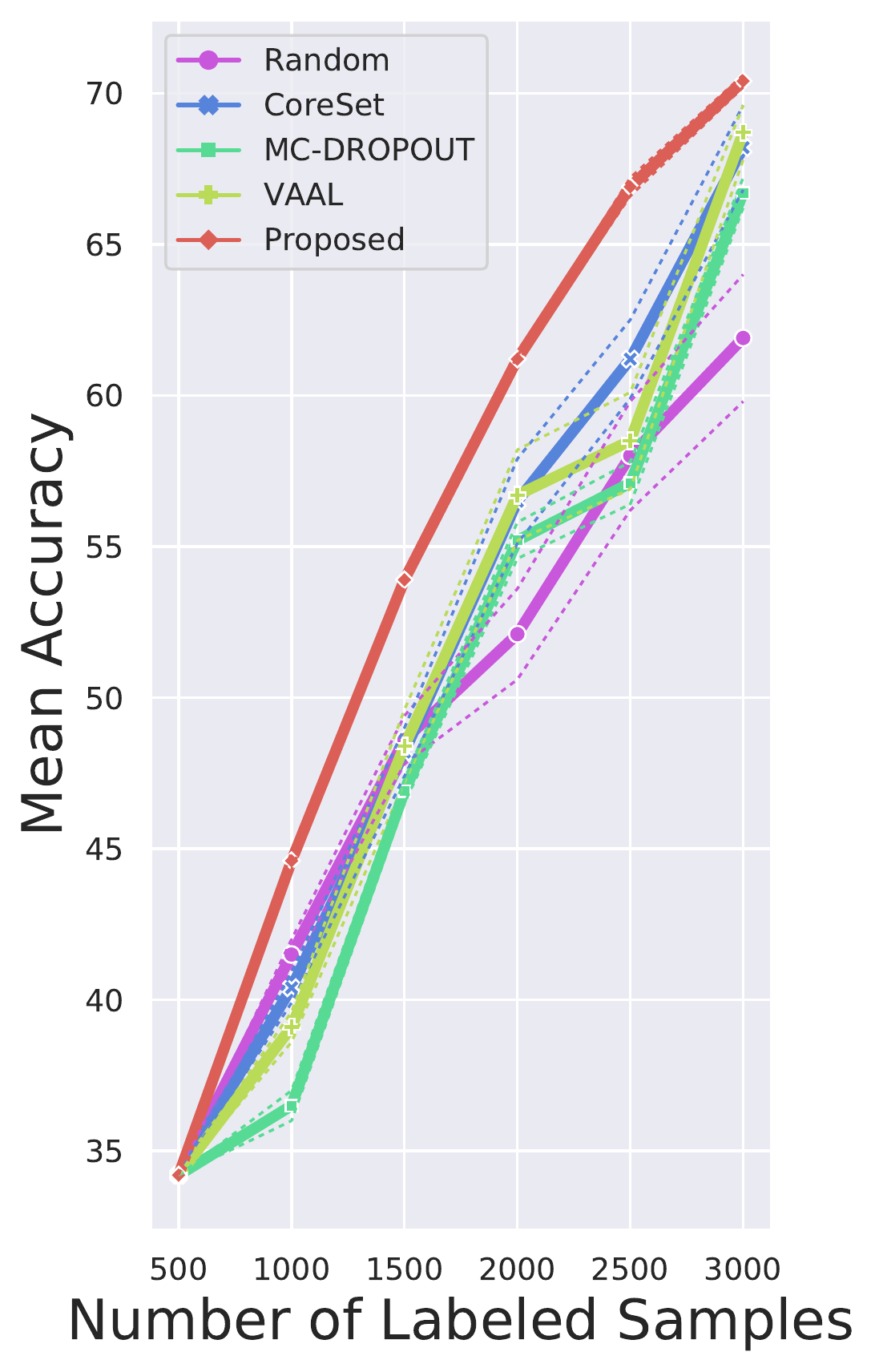}}
\caption{Results for the dominant variants of CIFAR-10}
\label{fig:dominantCIFAR10_class}
\end{figure*}

\begin{figure*}[!b]
\centering
\subfigure[Results for CIFAR-10$^{-[1]}$]{\includegraphics[width=0.32\linewidth]{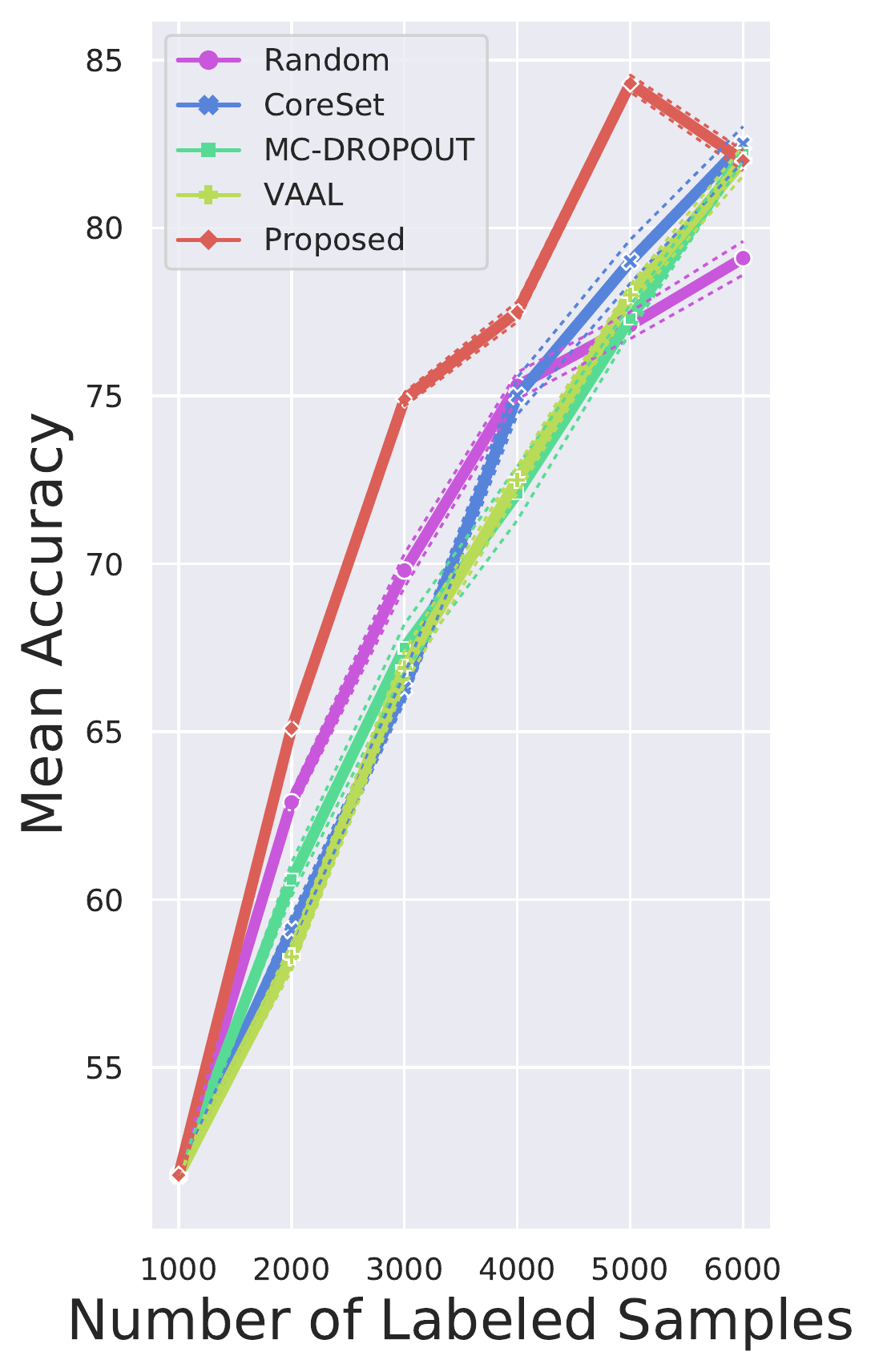}}
\subfigure[Results for CIFAR-10$^{-[5]}$]{\includegraphics[width=0.32\linewidth]{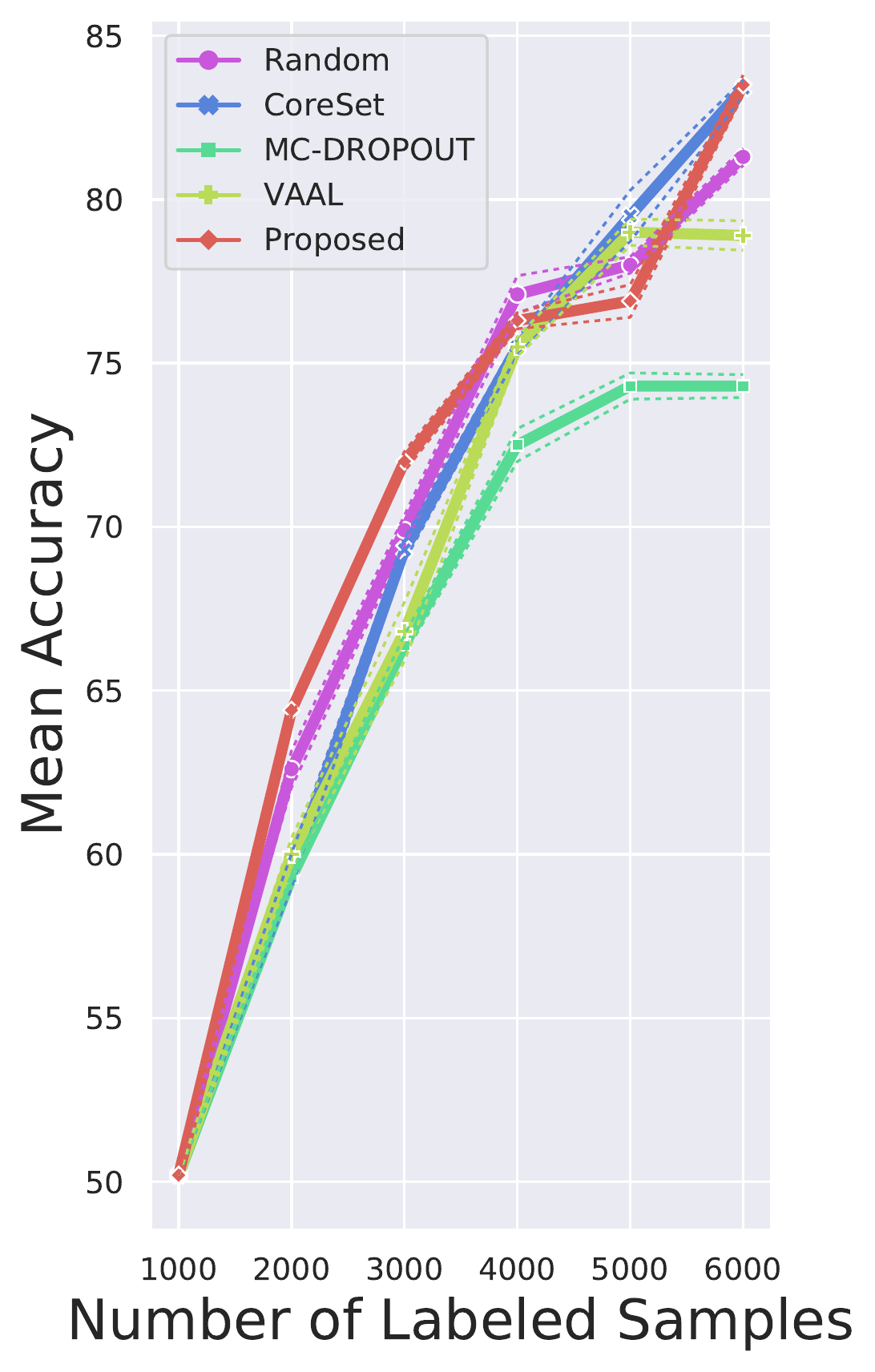}}
\subfigure[Results for CIFAR-10$^{-[10]}$]{\includegraphics[width=0.32\linewidth]{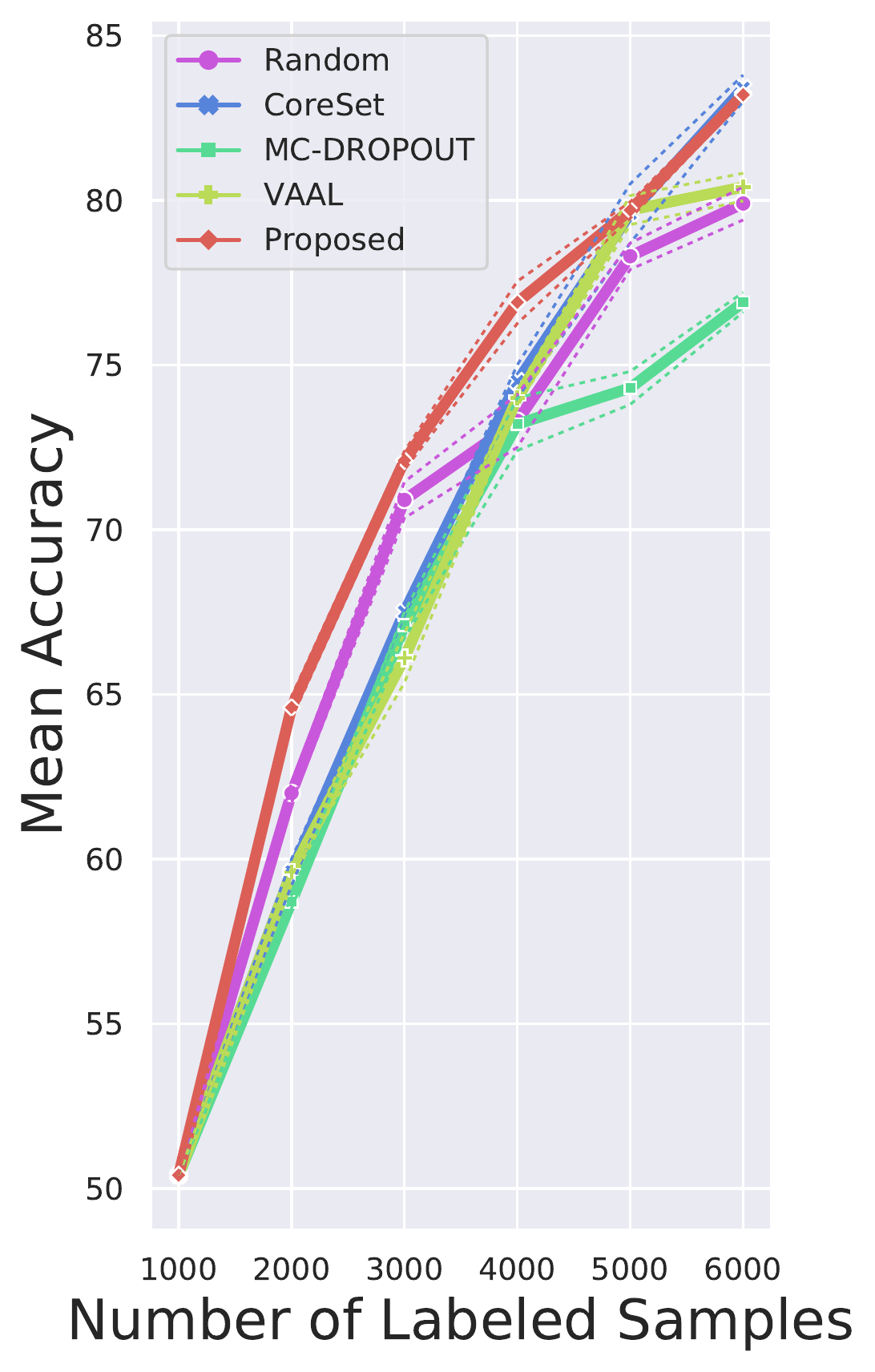}}
\caption{Results for the rare variants of CIFAR-10}
\label{fig:rareCIFAR10_class}
\end{figure*}

As mentioned in the main script, we report the individual result to demonstrate that our results are not by chance.
In Fig.~\ref{fig:dominantCIFAR10_class} we show the averages of the three trials for each dataset of CIFAR-10$^{+[1]}$, CIFAR-10$^{+[5]}$, and CIFAR-10$^{+[10]}$.
Likewise, in Fig.~\ref{fig:rareCIFAR10_class} we present the plots averaging the three trials for CIFAR-10$^{-[1]}$, CIFAR-10$^{-[5]}$, and CIFAR-10$^{-[10]}$.

\begin{figure*}[t]
\centering
\subfigure[Trial $1$]{\includegraphics[width=0.32\linewidth]{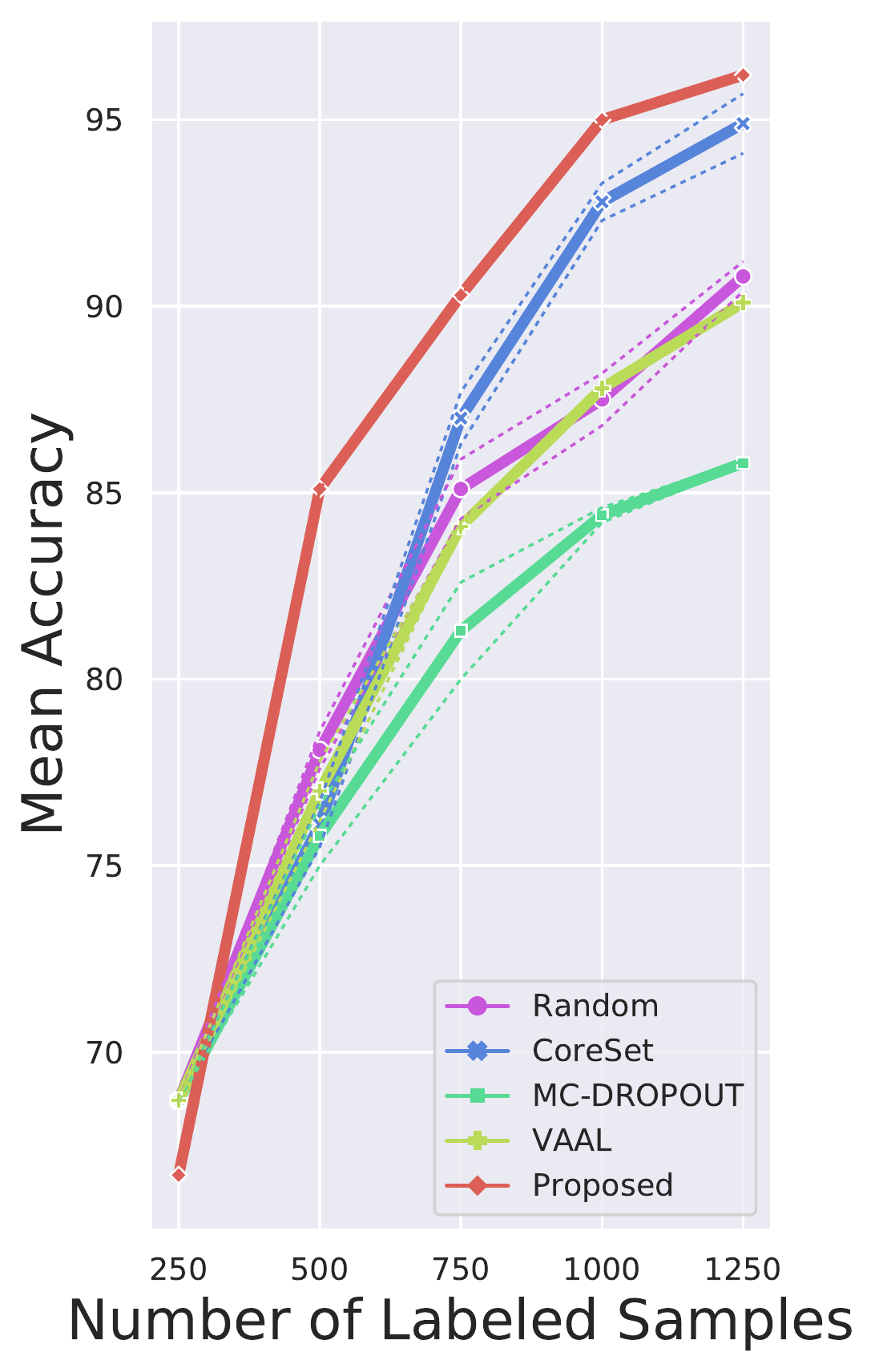}}
\subfigure[Trial $2$]{\includegraphics[width=0.32\linewidth]{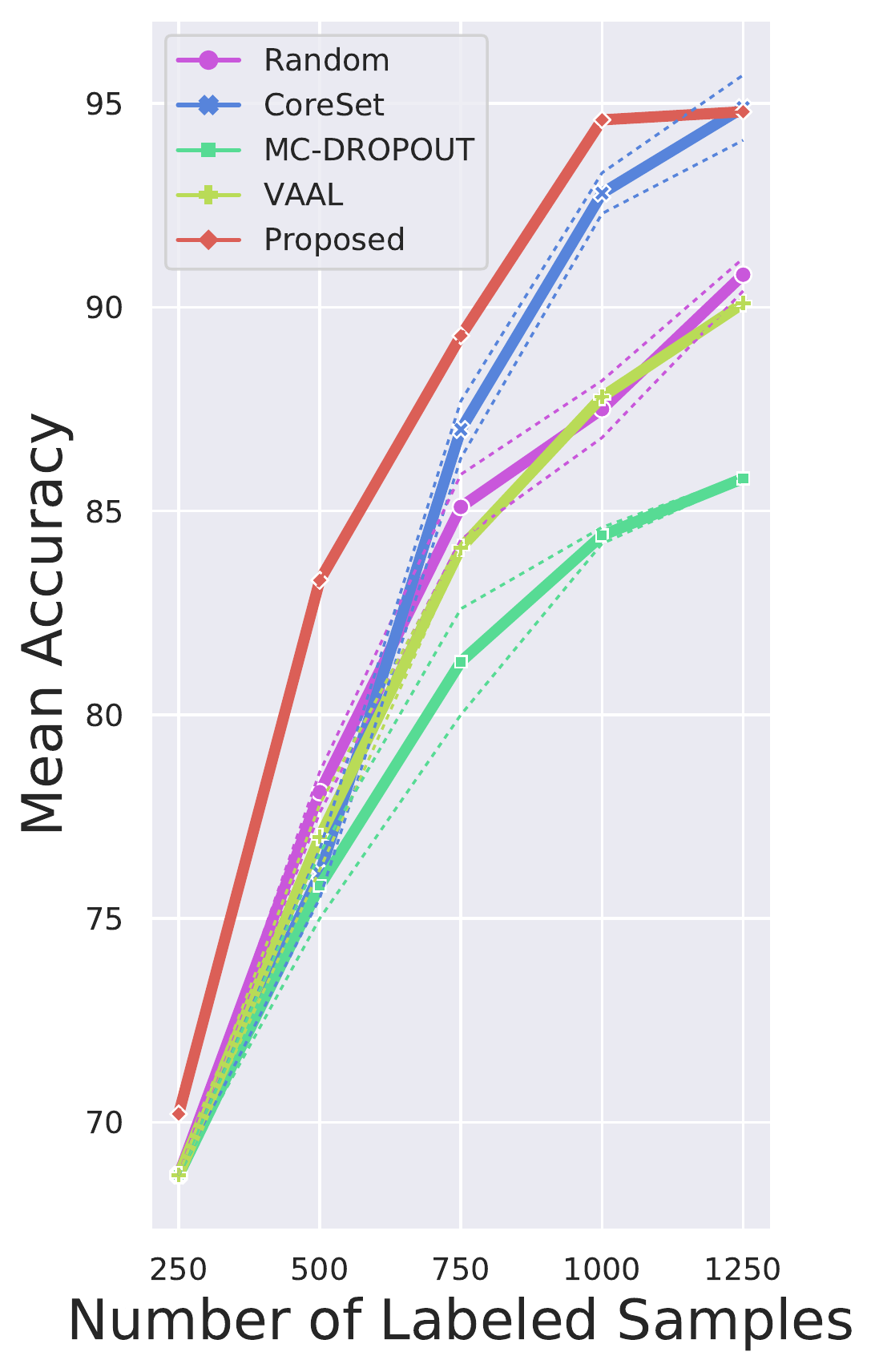}}
\subfigure[Trial $3$]{\includegraphics[width=0.32\linewidth]{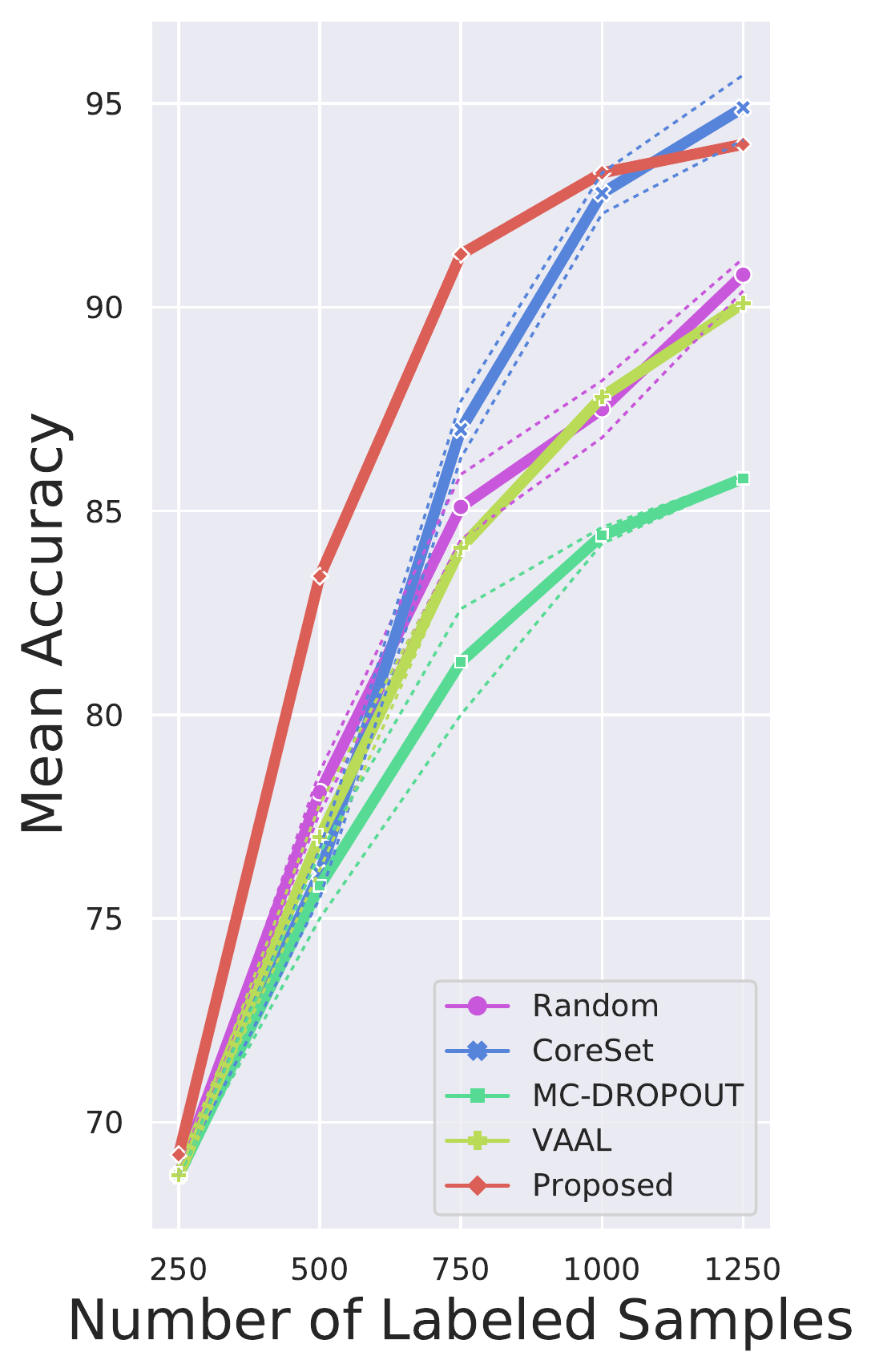}}
\caption{Individual results for the NEU dataset}
\label{fig:fullNEU_single_0}
\end{figure*}

\begin{figure*}[t]
\centering
\subfigure[Trial $1$]{\includegraphics[width=0.32\linewidth]{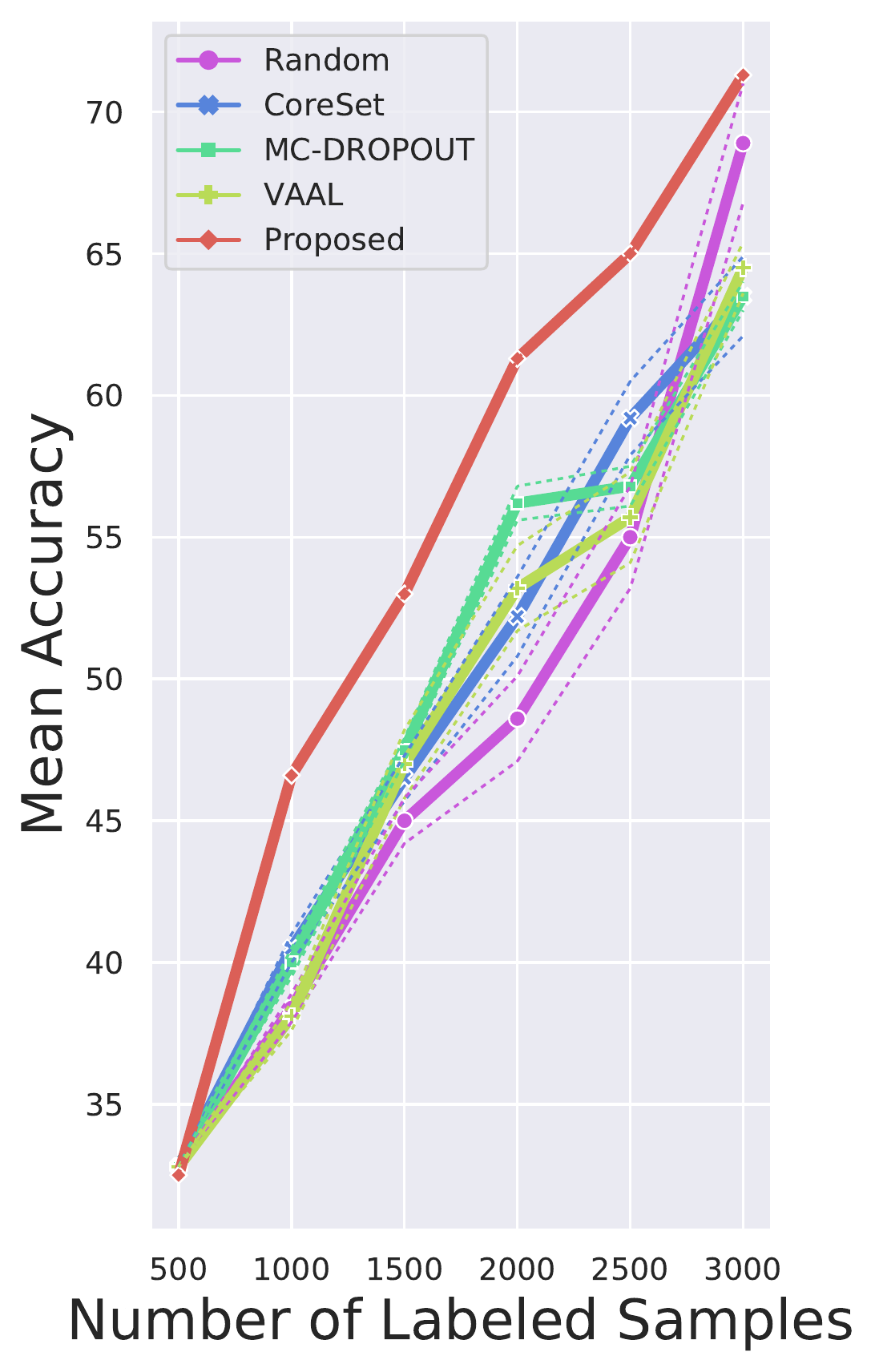}}
\subfigure[Trial $2$]{\includegraphics[width=0.32\linewidth]{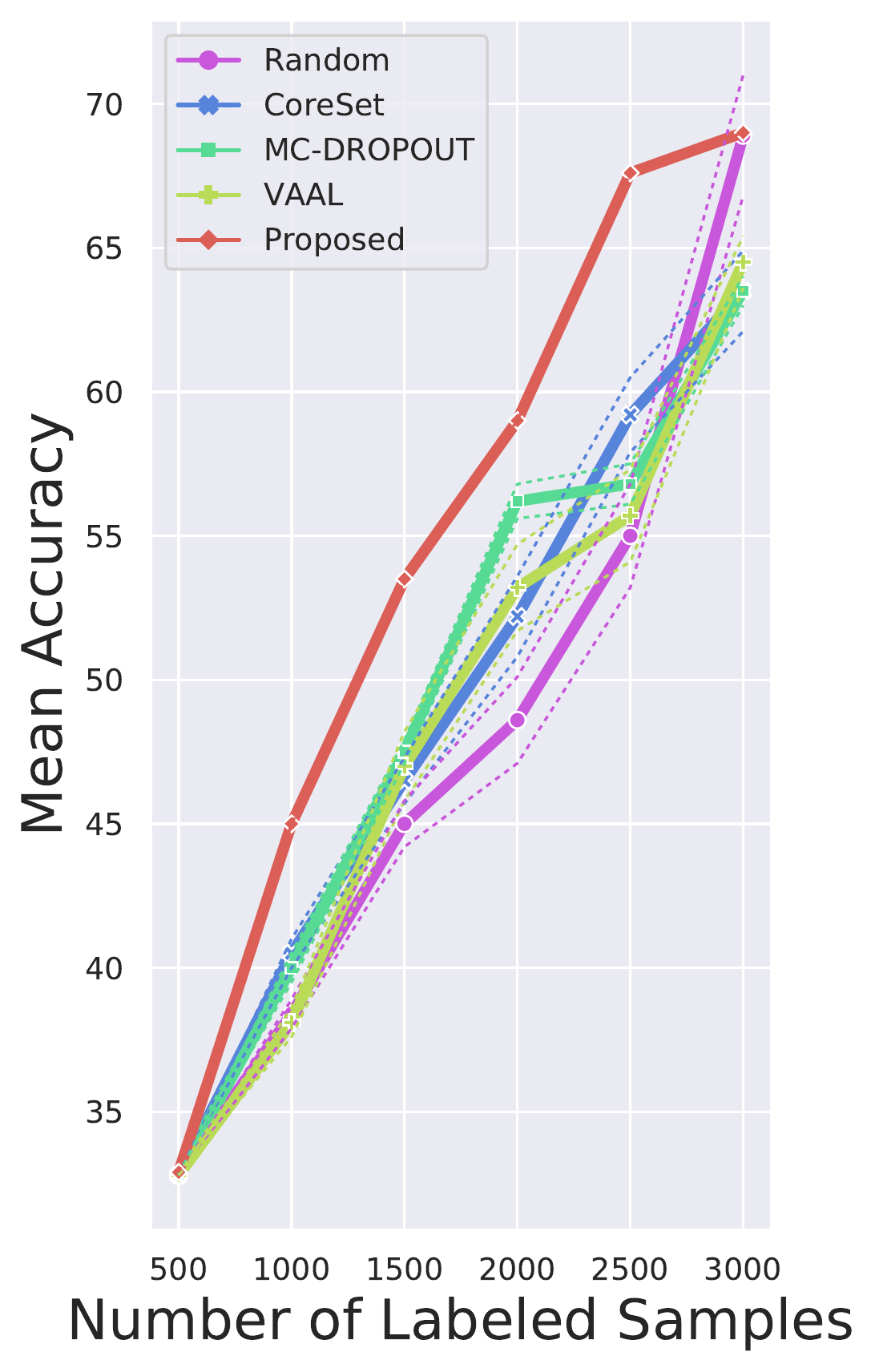}}
\subfigure[Trial $3$]{\includegraphics[width=0.32\linewidth]{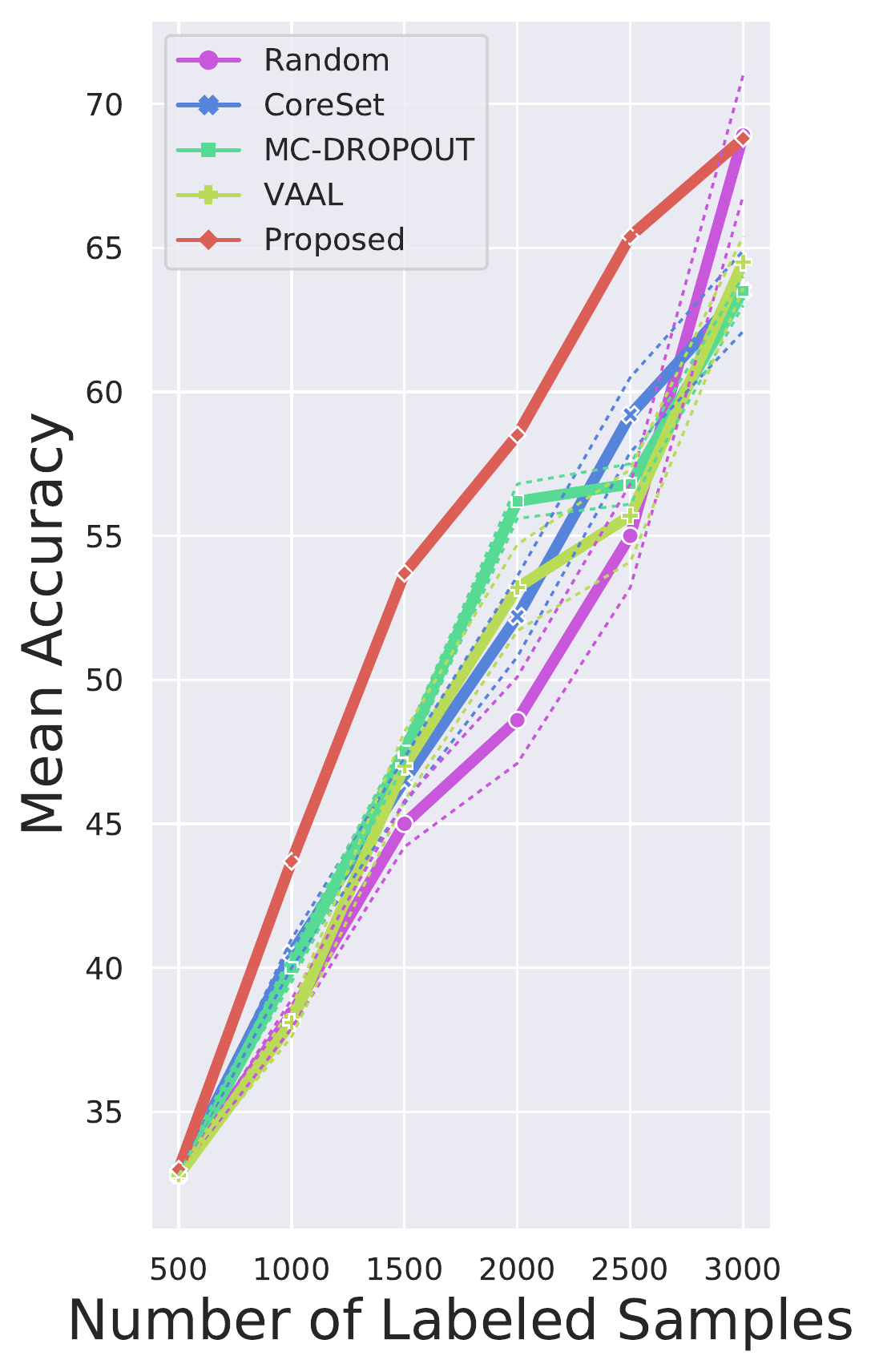}}
\caption{Individual results for CIFAR-10$^{+[1]}$}
\label{fig:dominantCIFAR10_single_0}
\end{figure*}

\begin{figure*}[t]
\centering
\subfigure[Trial $1$]{\includegraphics[width=0.32\linewidth]{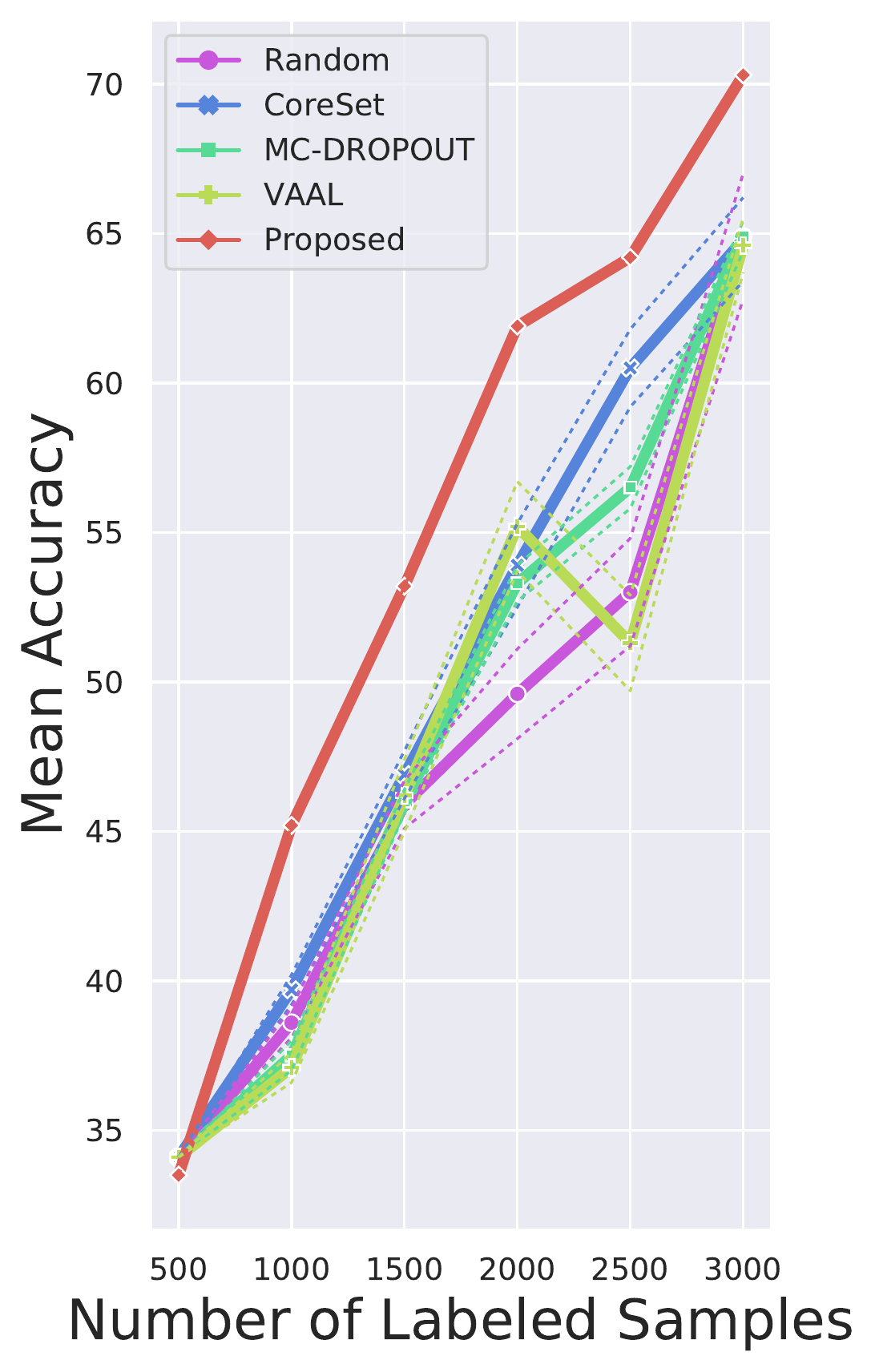}}
\subfigure[Trial $2$]{\includegraphics[width=0.32\linewidth]{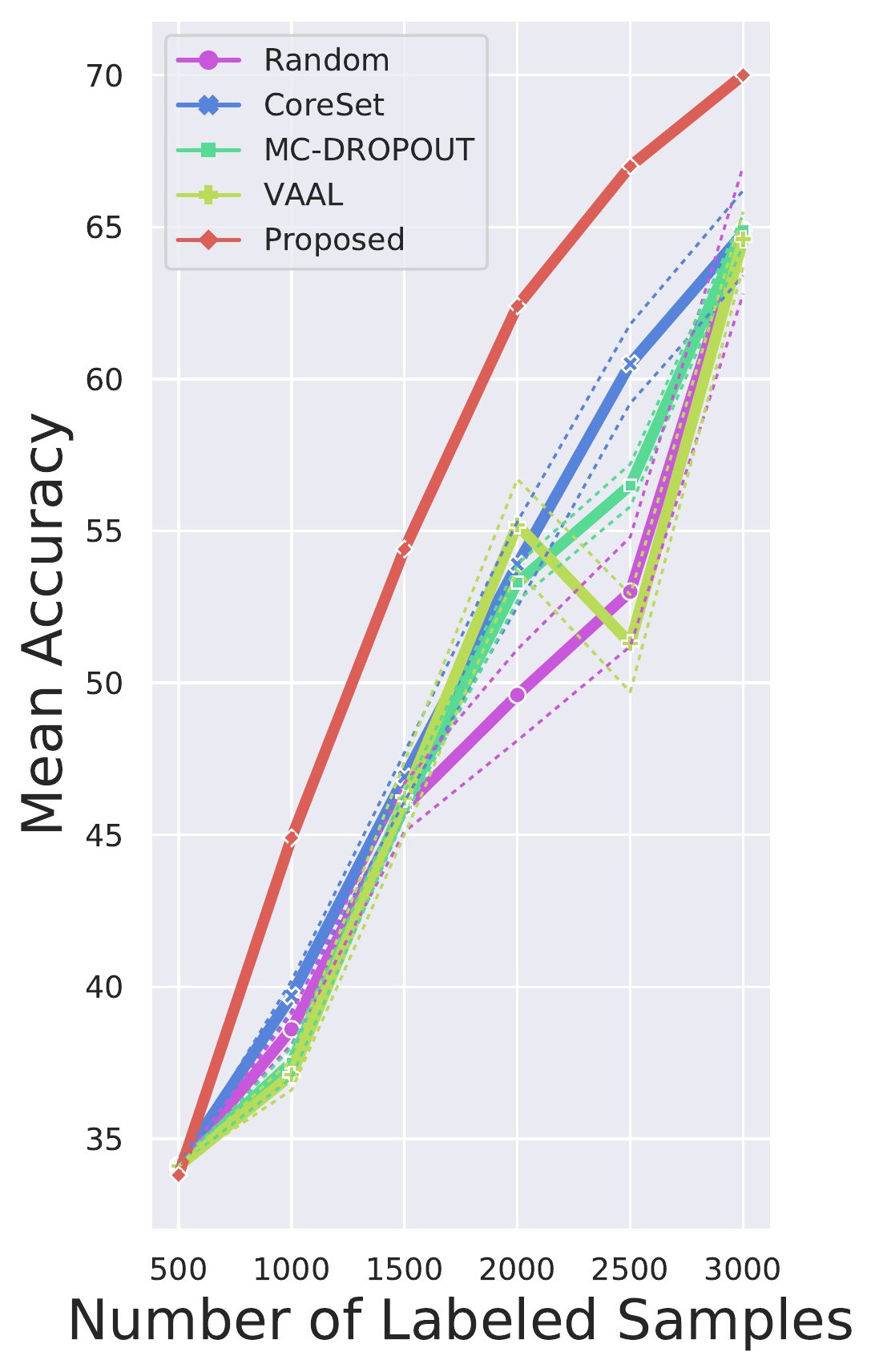}}
\subfigure[Trial $3$]{\includegraphics[width=0.32\linewidth]{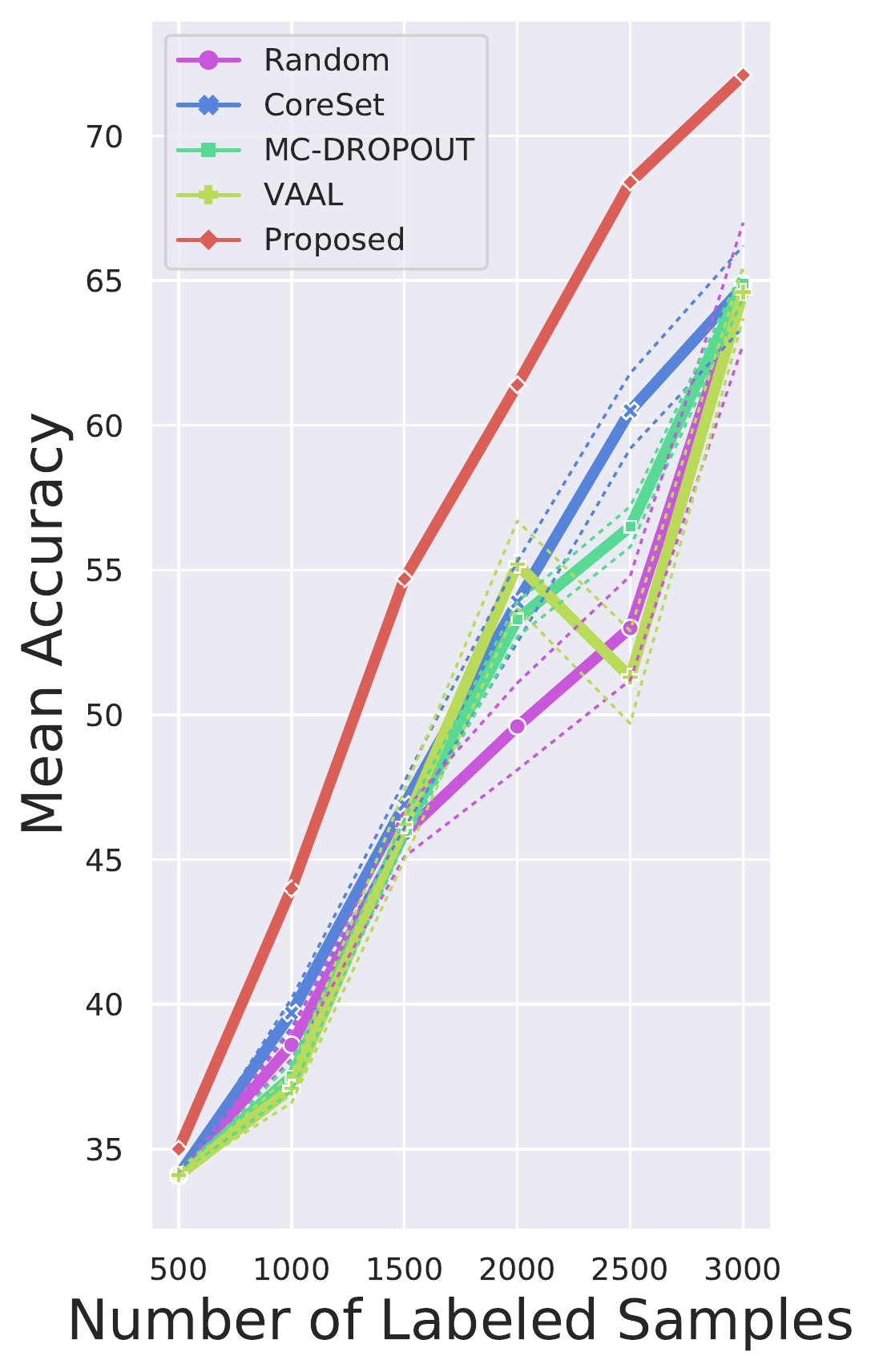}}
\caption{Individual results for CIFAR-10$^{+[5]}$}
\label{fig:dominantCIFAR10_single_3}
\end{figure*}

\begin{figure*}[t]
\centering
\subfigure[Trial $1$]{\includegraphics[width=0.32\linewidth]{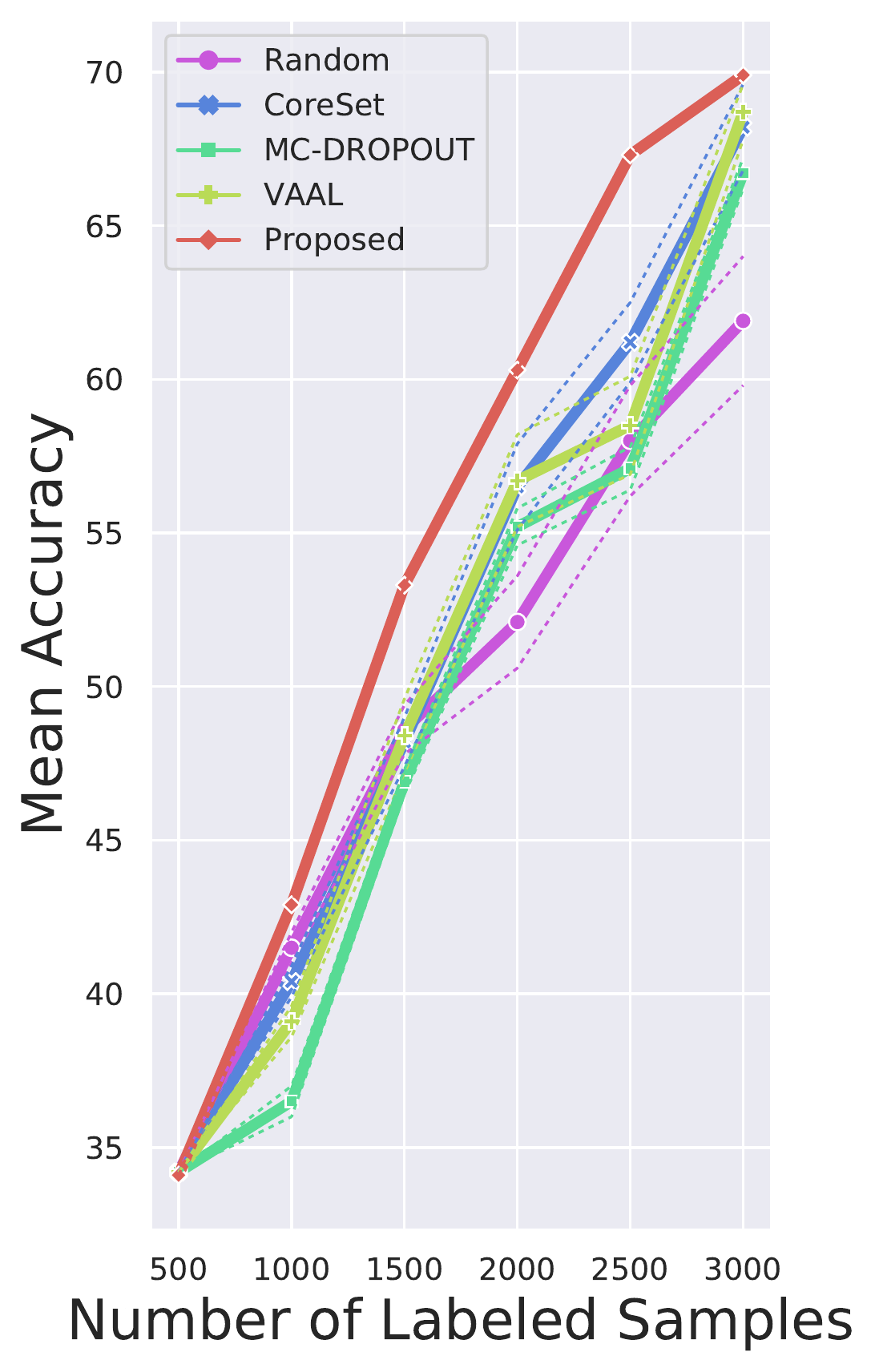}}
\subfigure[Trial $2$]{\includegraphics[width=0.32\linewidth]{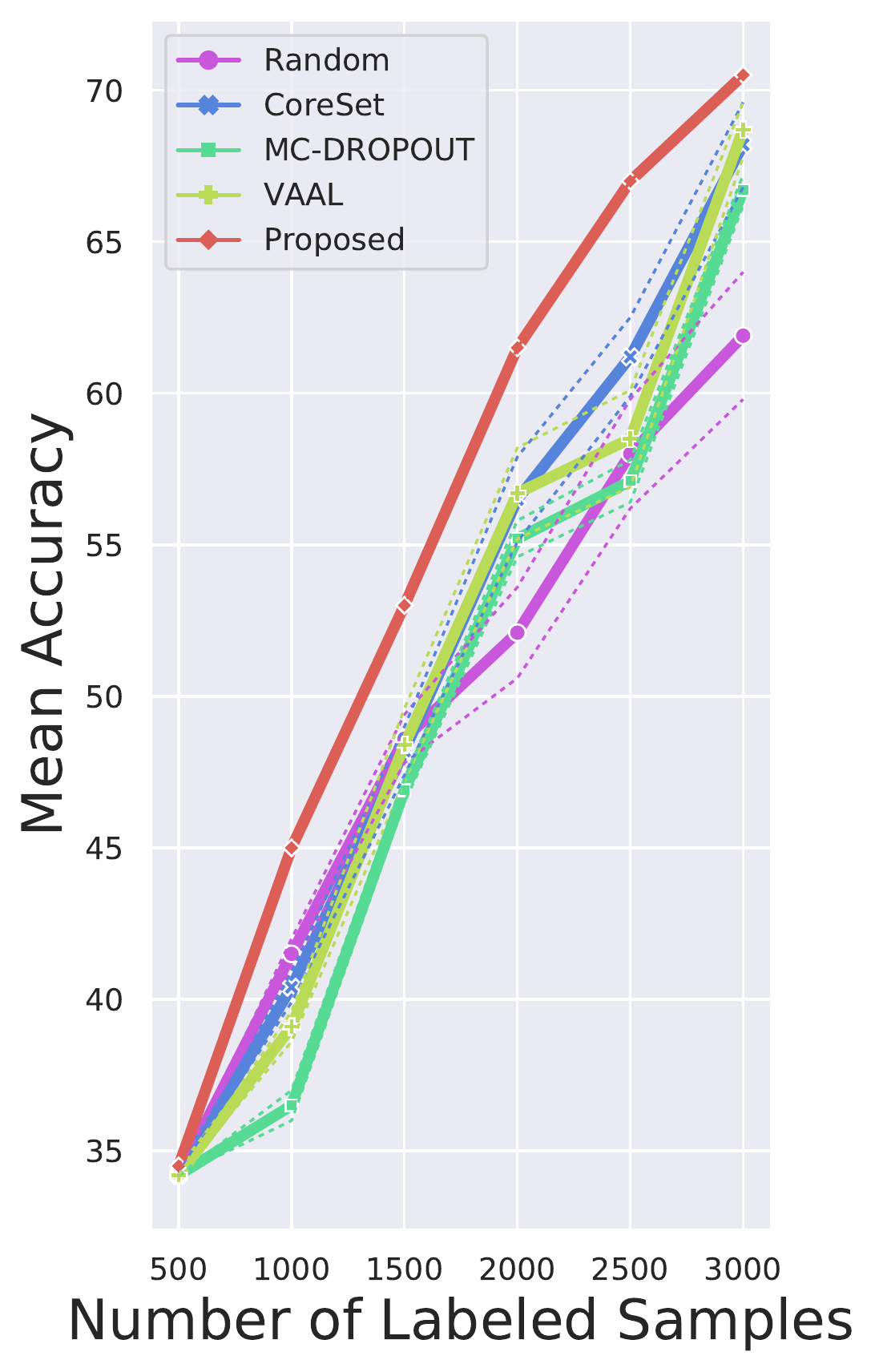}}
\subfigure[Trial $3$]{\includegraphics[width=0.32\linewidth]{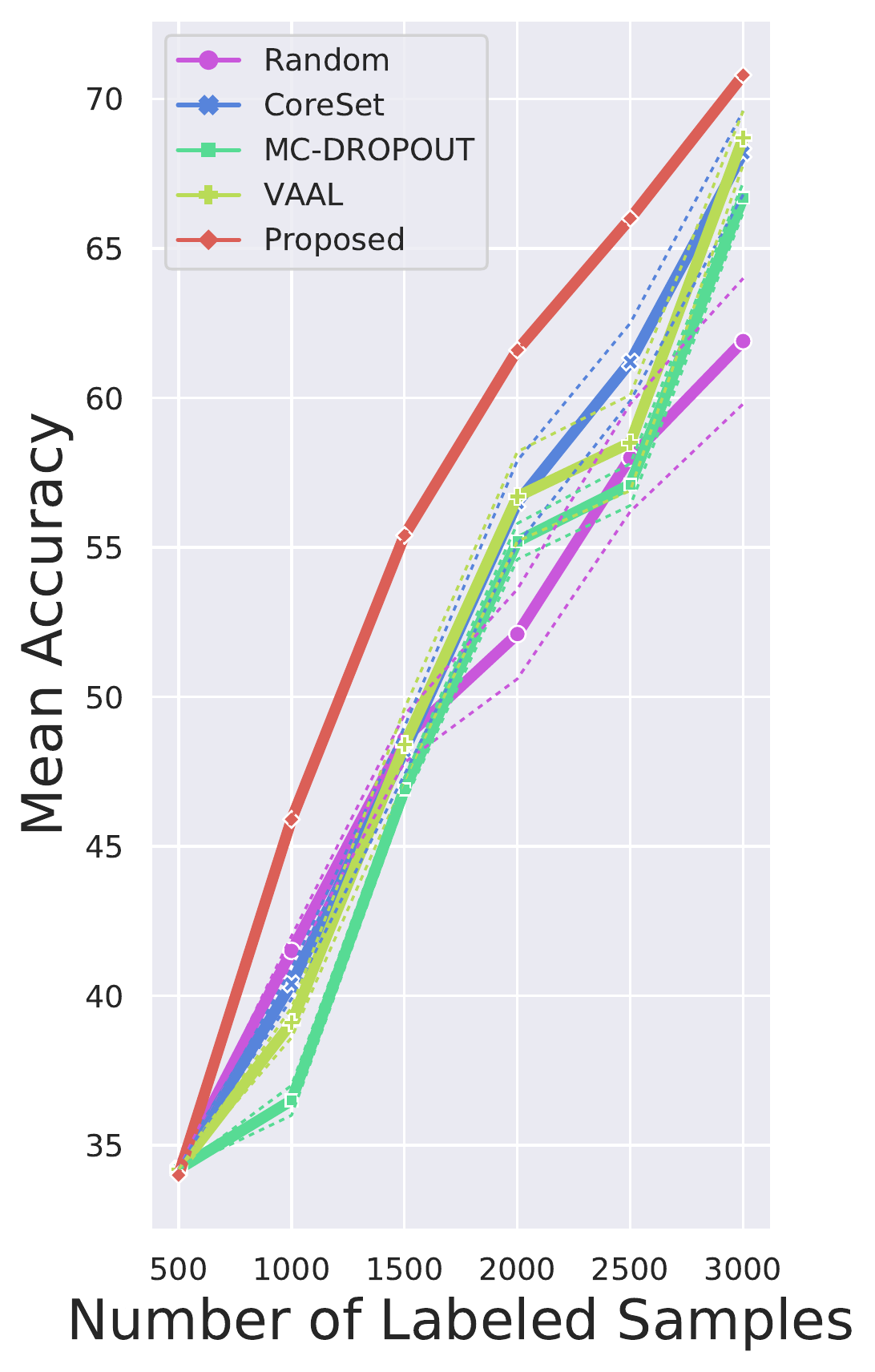}}
\caption{Individual results for CIFAR-10$^{+[10]}$}
\label{fig:dominantCIFAR10_single_6}
\end{figure*}

\begin{figure*}[t]
\centering
\subfigure[Trial $1$]{\includegraphics[width=0.32\linewidth]{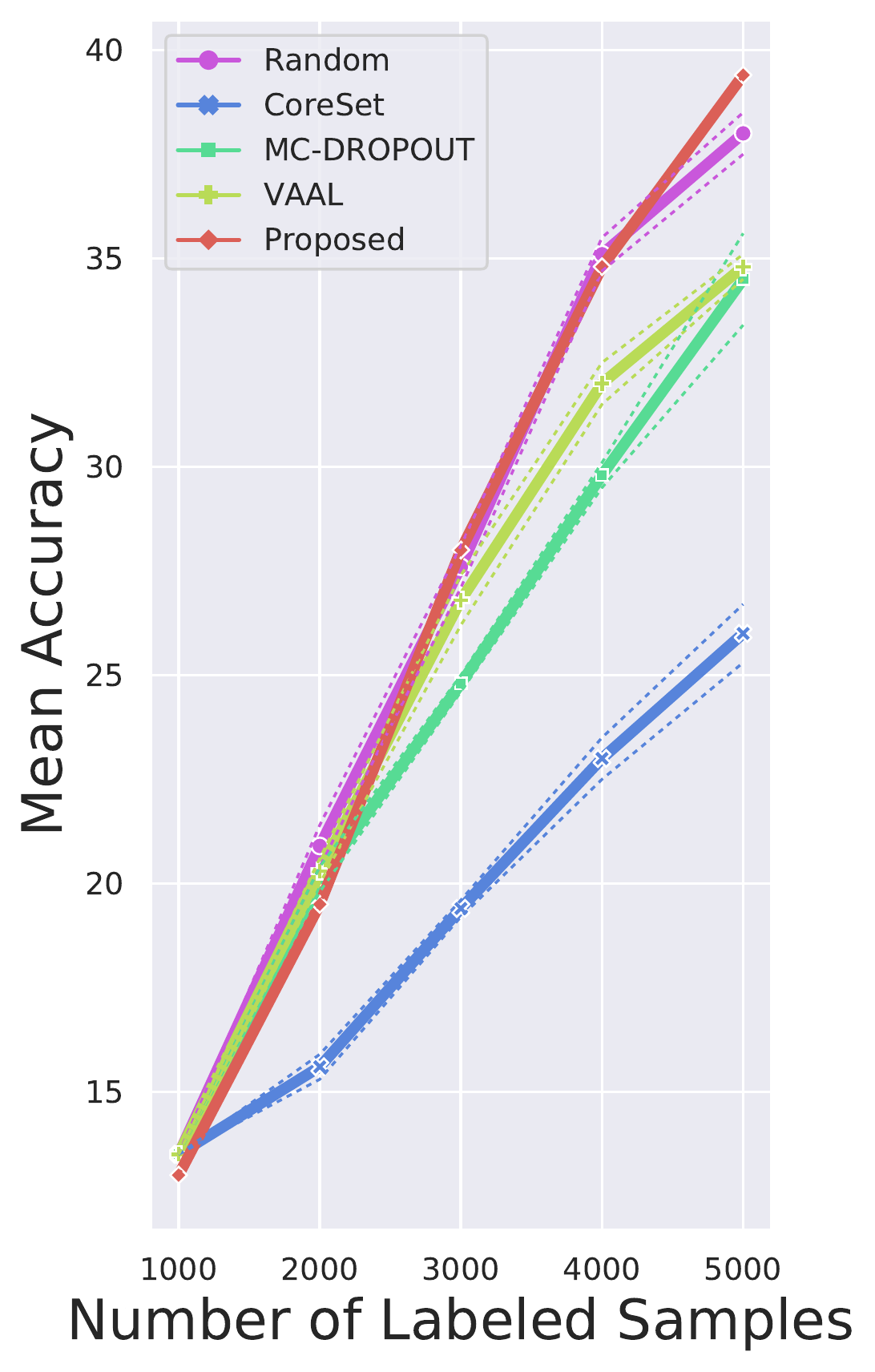}}
\subfigure[Trial $2$]{\includegraphics[width=0.32\linewidth]{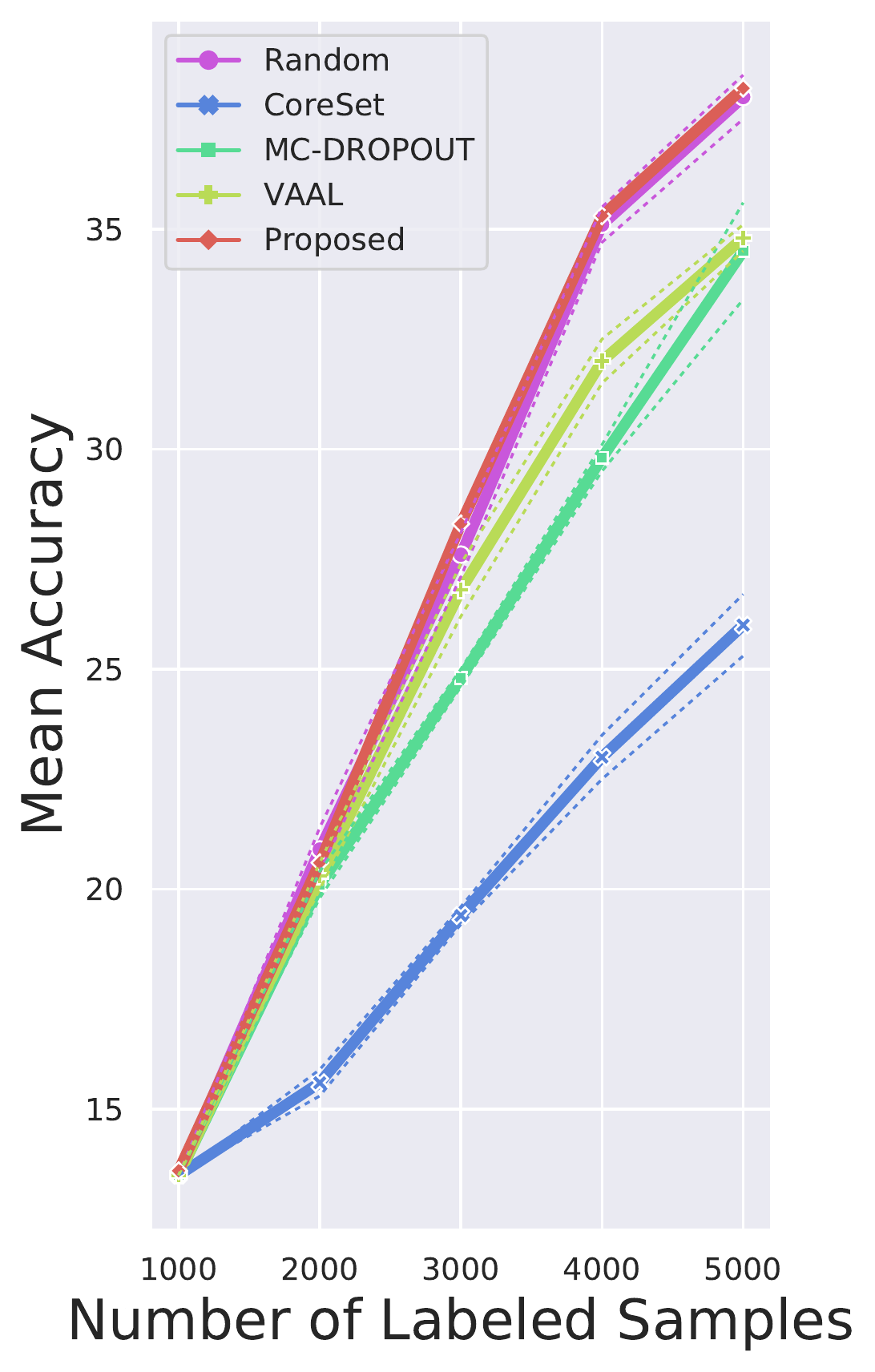}}
\subfigure[Trial $3$]{\includegraphics[width=0.32\linewidth]{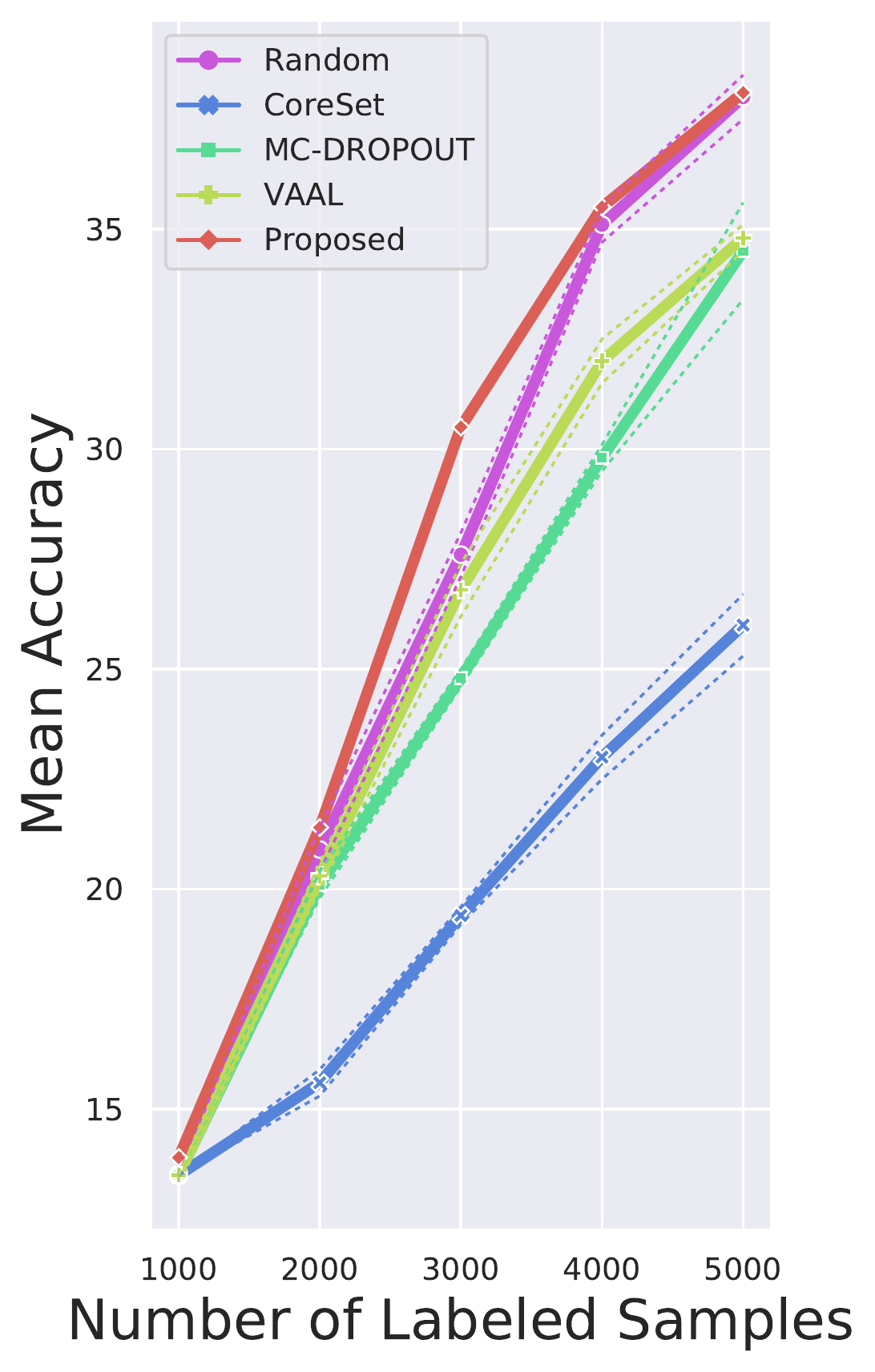}}
\caption{Individual results for CIFAR-100$^{+[45:55]}$}
\label{fig:dominantCIFAR100_single_0}
\end{figure*}

\begin{figure*}[t]
\centering
\subfigure[Trial $1$]{\includegraphics[width=0.32\linewidth]{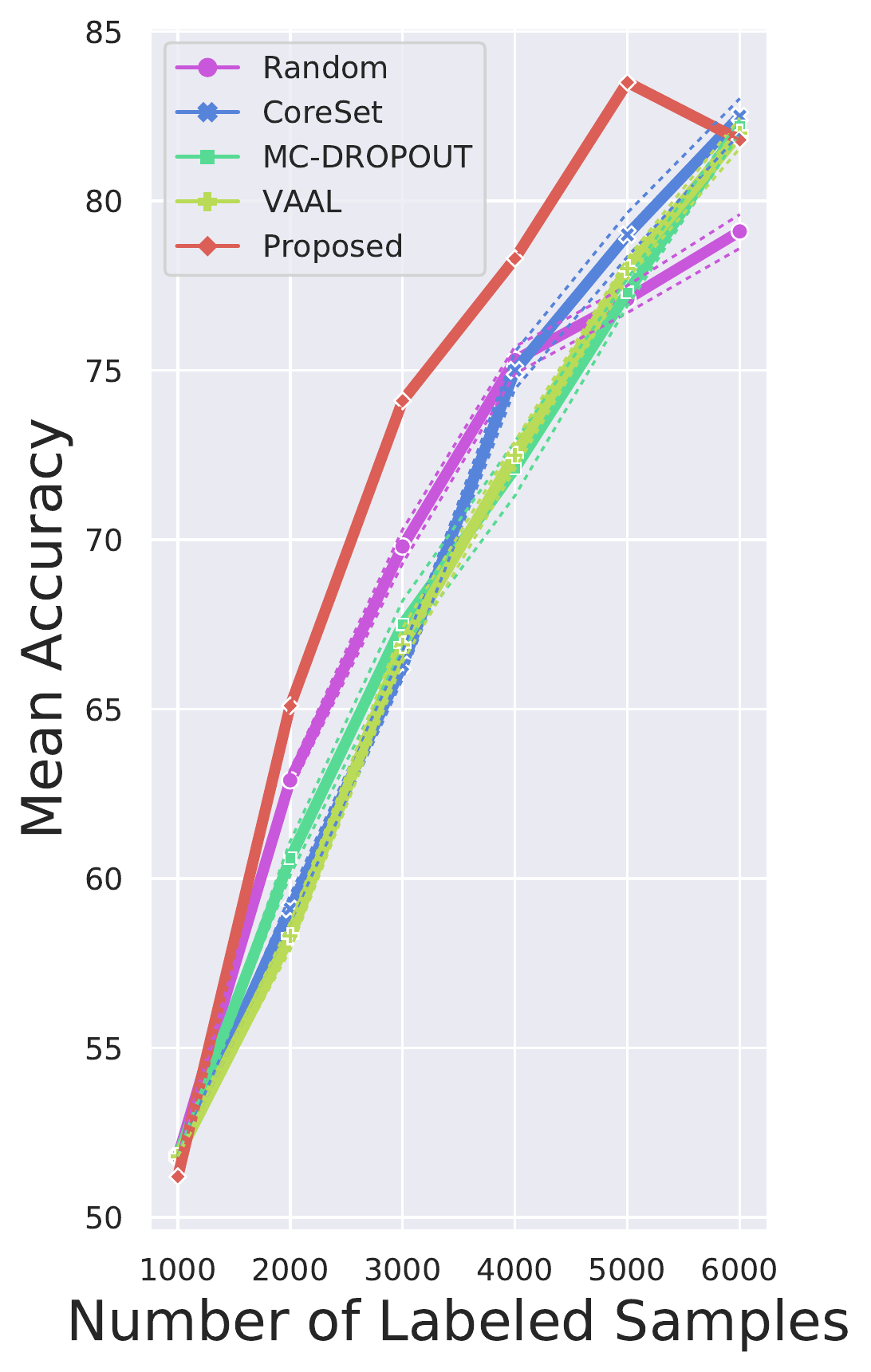}}
\subfigure[Trial $2$]{\includegraphics[width=0.32\linewidth]{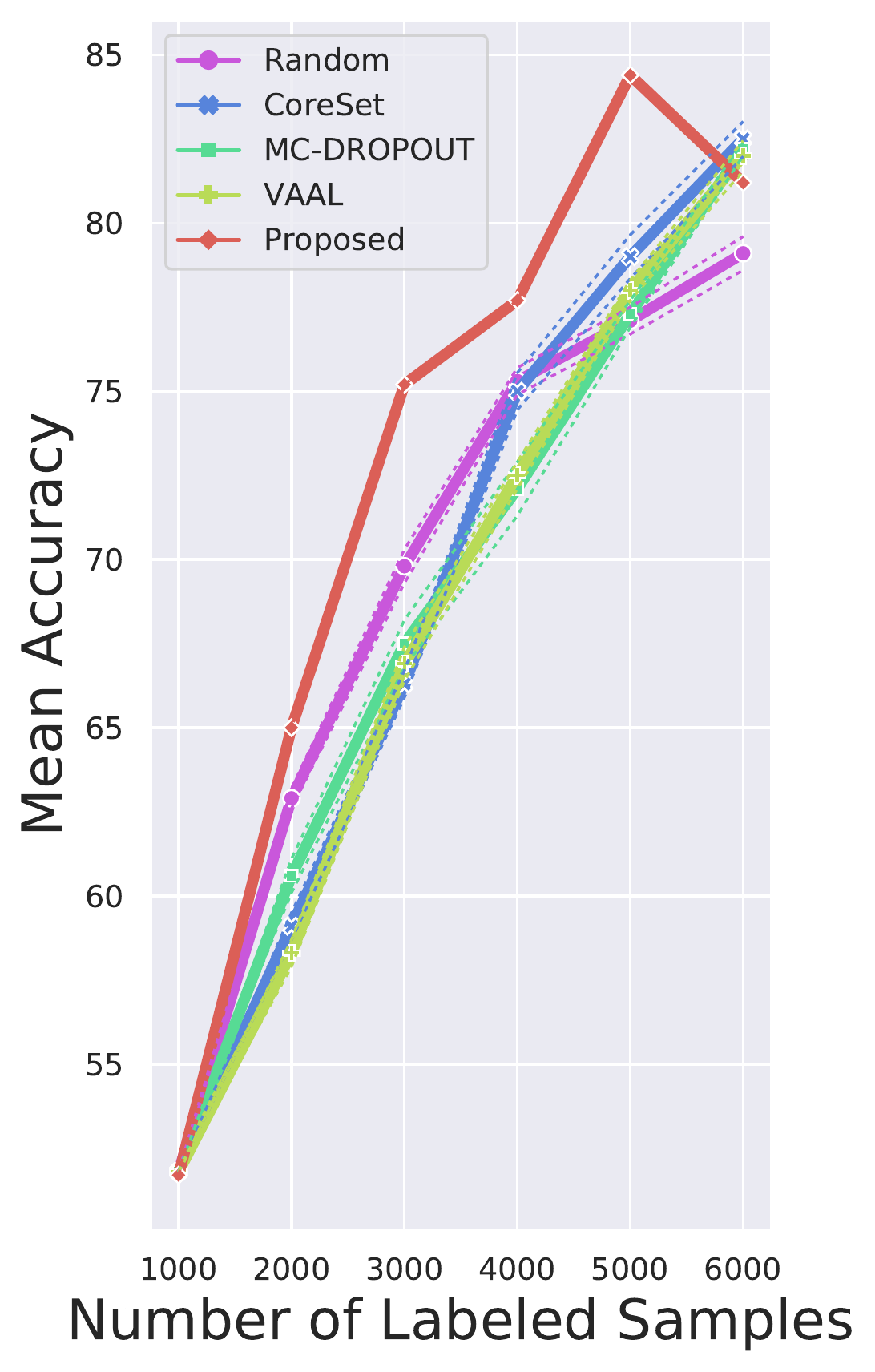}}
\subfigure[Trial $3$]{\includegraphics[width=0.32\linewidth]{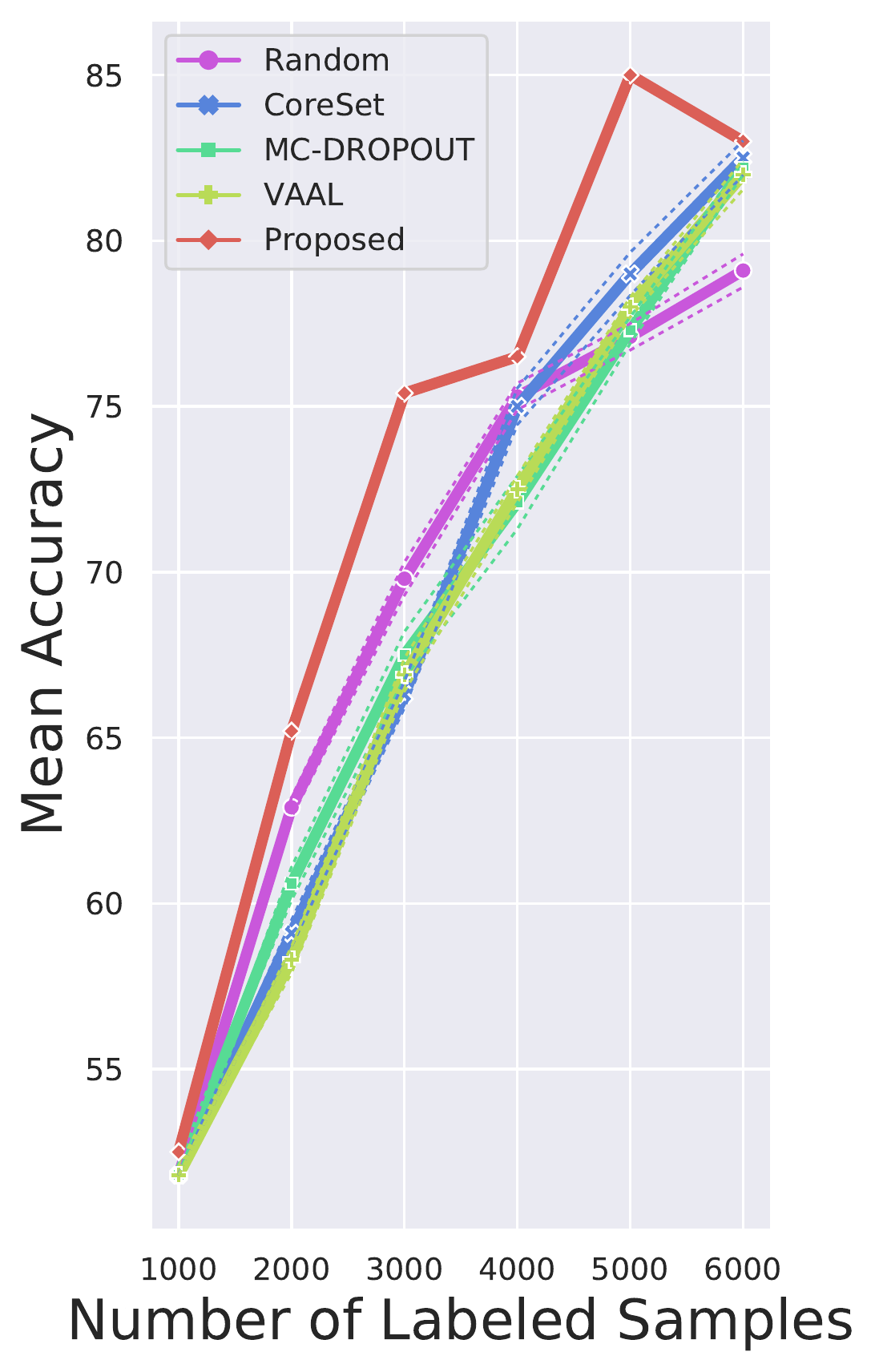}}
\caption{Individual results for CIFAR-10$^{-[1]}$}
\label{fig:rareCIFAR10_single_0}
\end{figure*}

\begin{figure*}[t]
\centering
\subfigure[Trial $1$]{\includegraphics[width=0.32\linewidth]{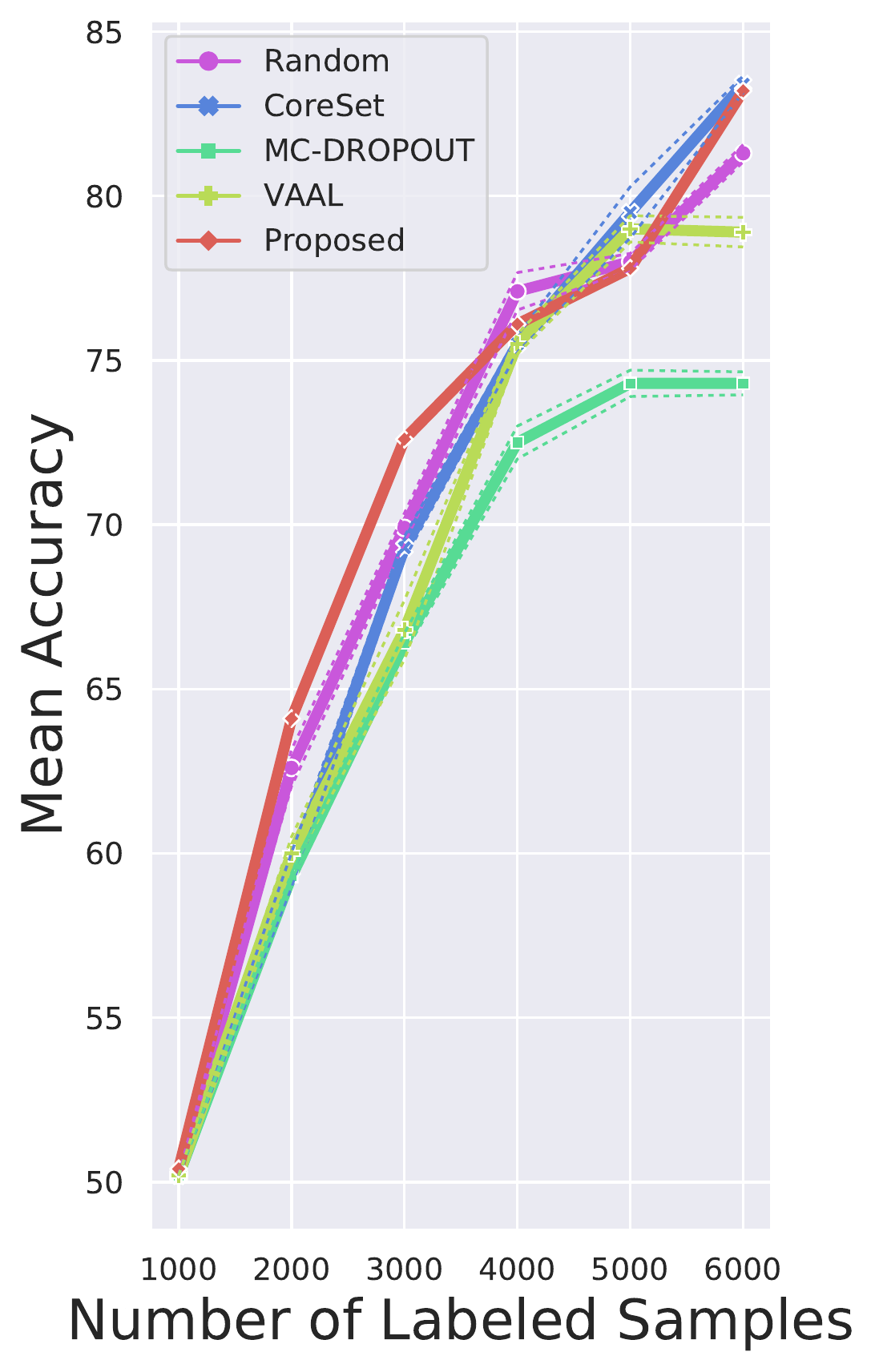}}
\subfigure[Trial $2$]{\includegraphics[width=0.32\linewidth]{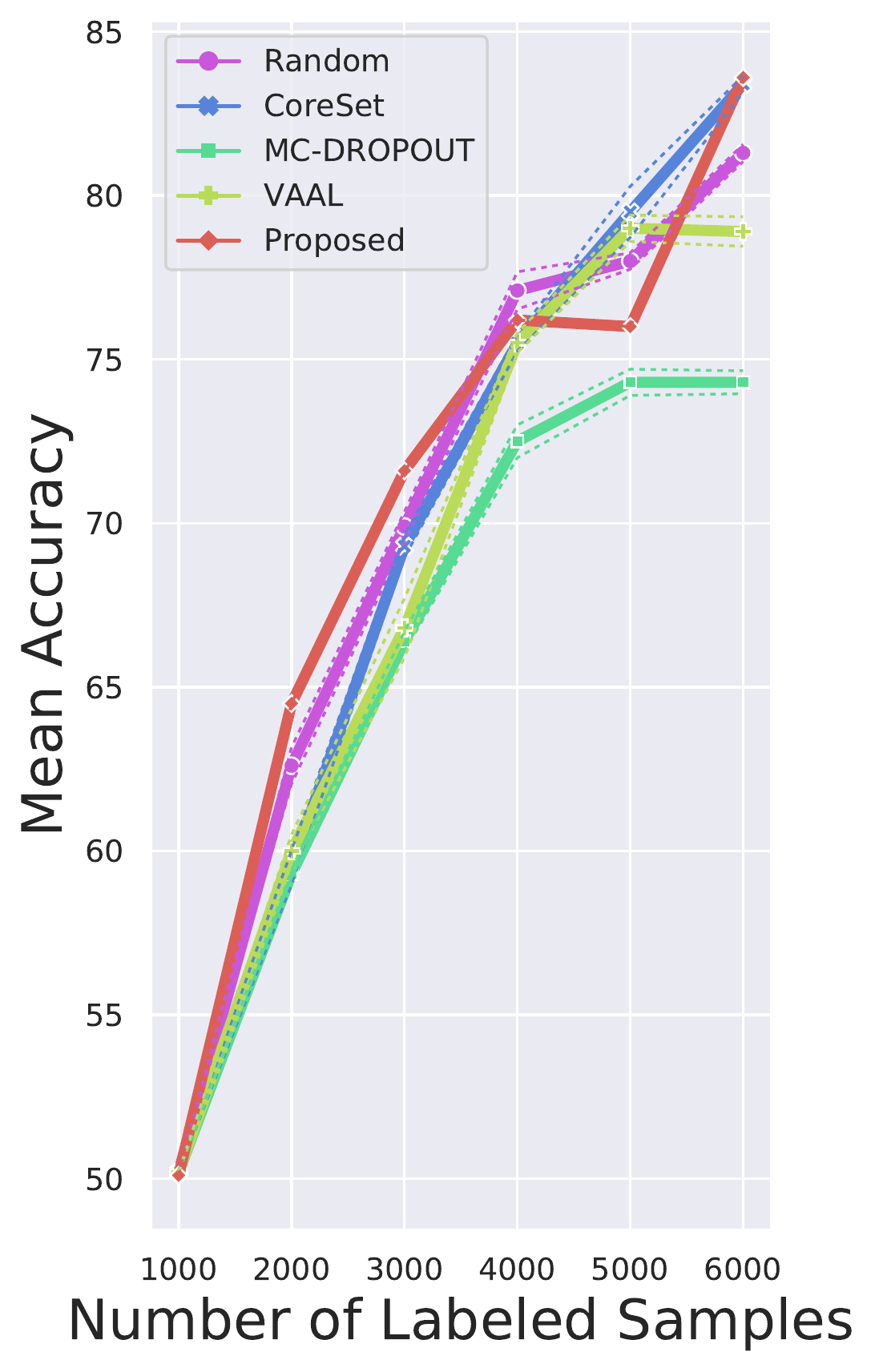}}
\subfigure[Trial $3$]{\includegraphics[width=0.32\linewidth]{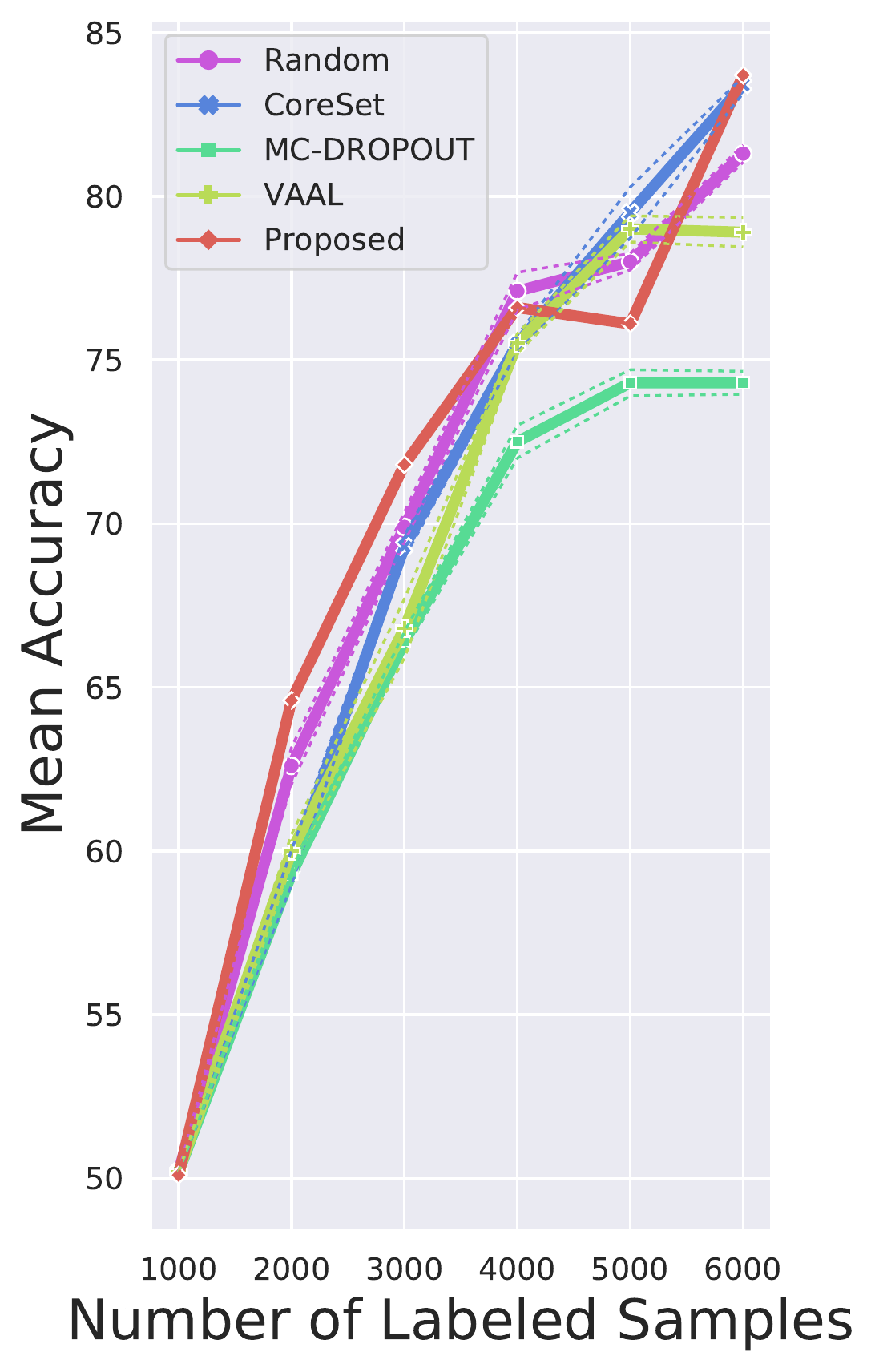}}
\caption{Individual results for CIFAR-10$^{-[5]}$}
\label{fig:rareCIFAR10_single_3}
\end{figure*}

\begin{figure*}[t]
\centering
\subfigure[Trial $1$]{\includegraphics[width=0.32\linewidth]{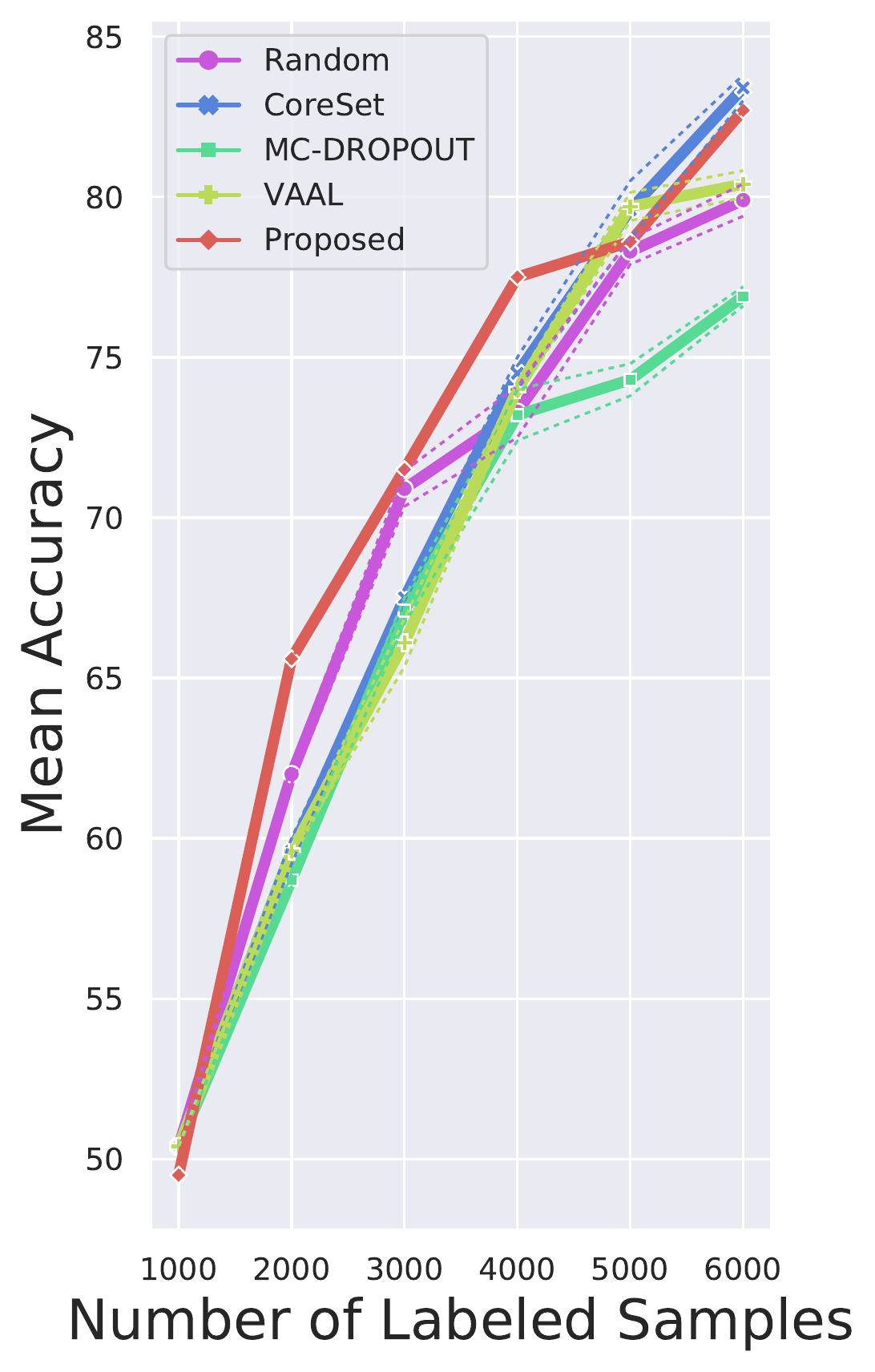}}
\subfigure[Trial $2$]{\includegraphics[width=0.32\linewidth]{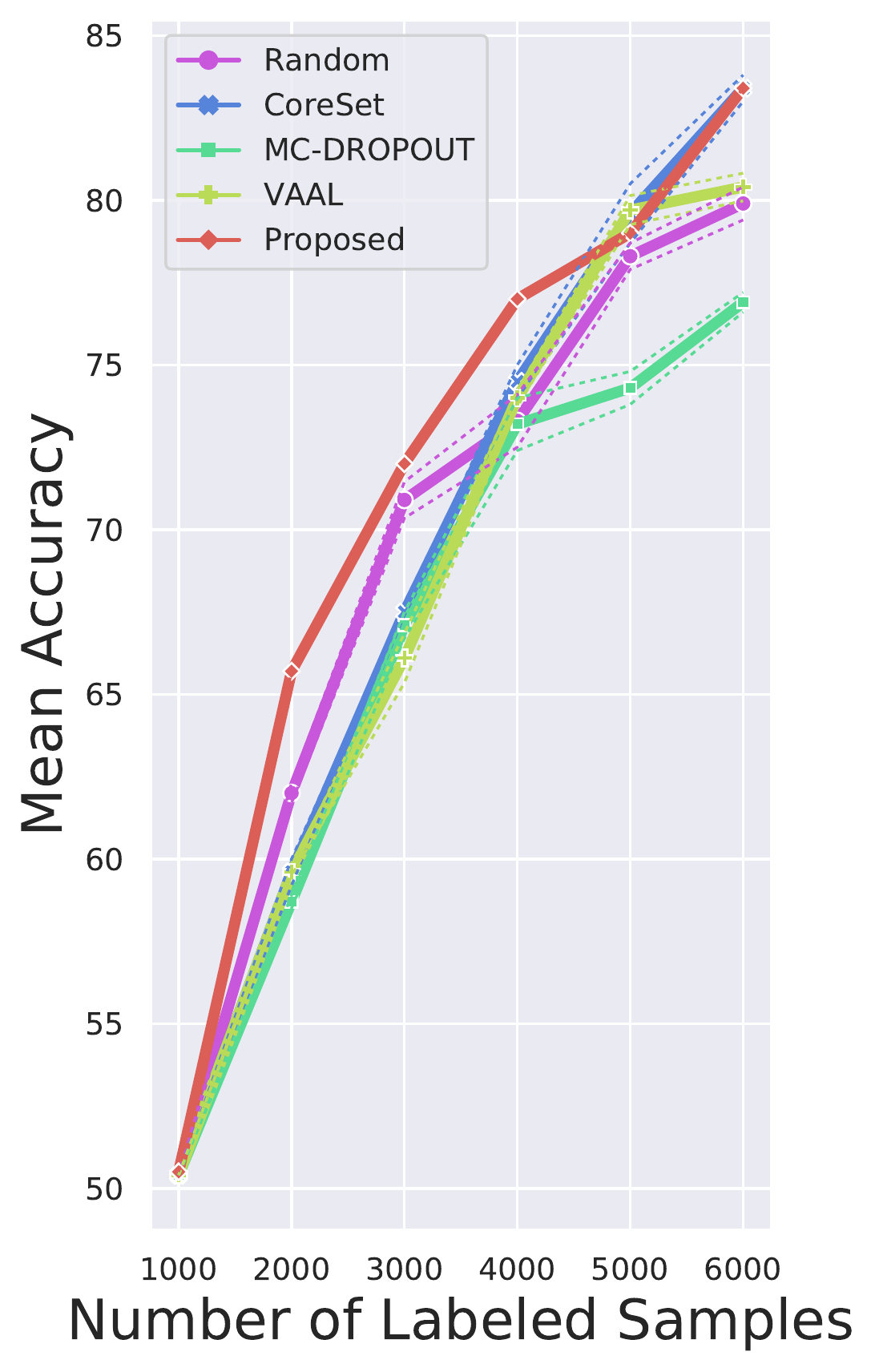}}
\subfigure[Trial $3$]{\includegraphics[width=0.32\linewidth]{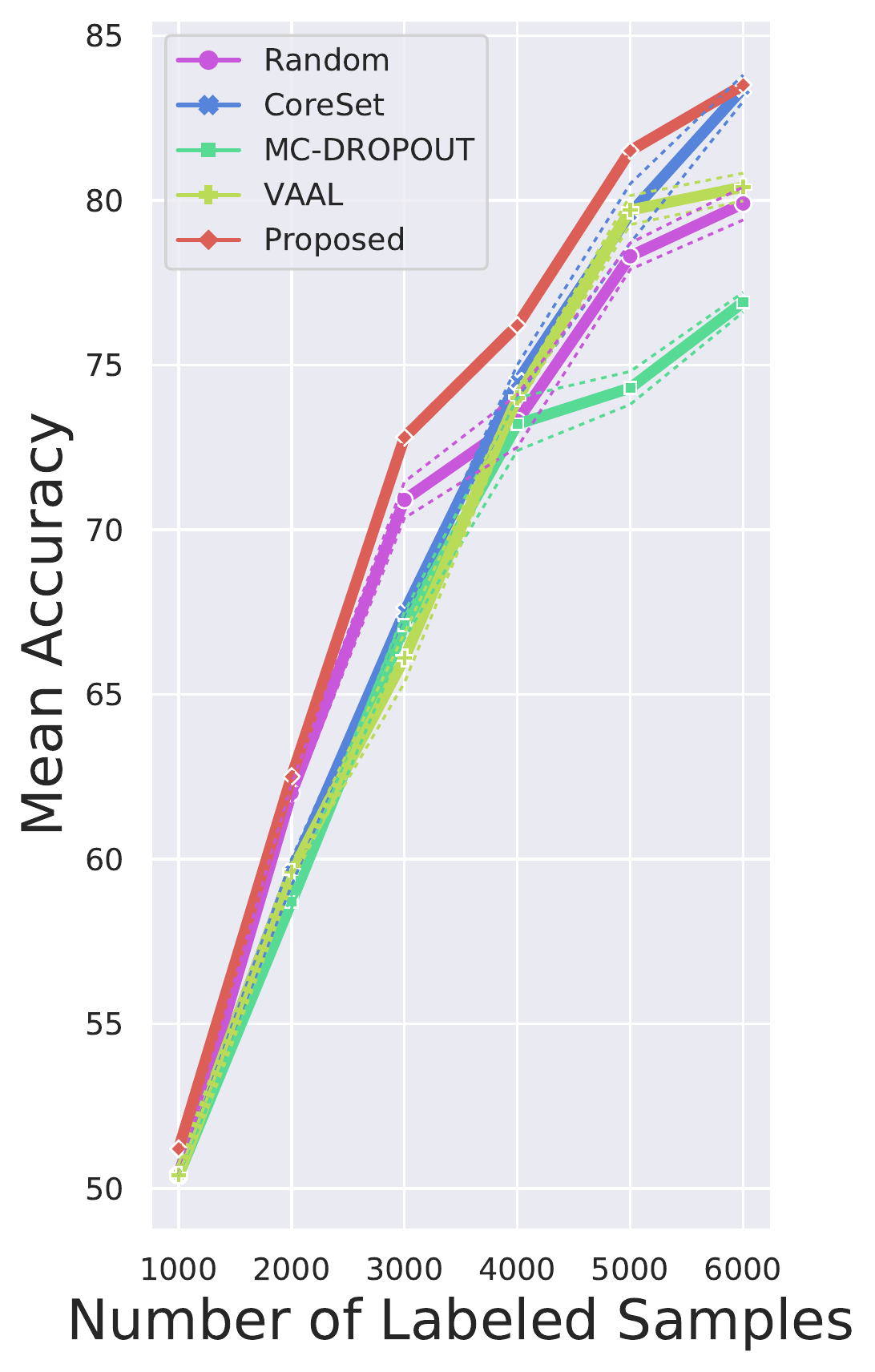}}
\caption{Individual results for CIFAR-10$^{-[10]}$}
\label{fig:rareCIFAR10_single_6}
\end{figure*}

\begin{figure*}[t]
\centering
\subfigure[Trial $1$]{\includegraphics[width=0.32\linewidth]{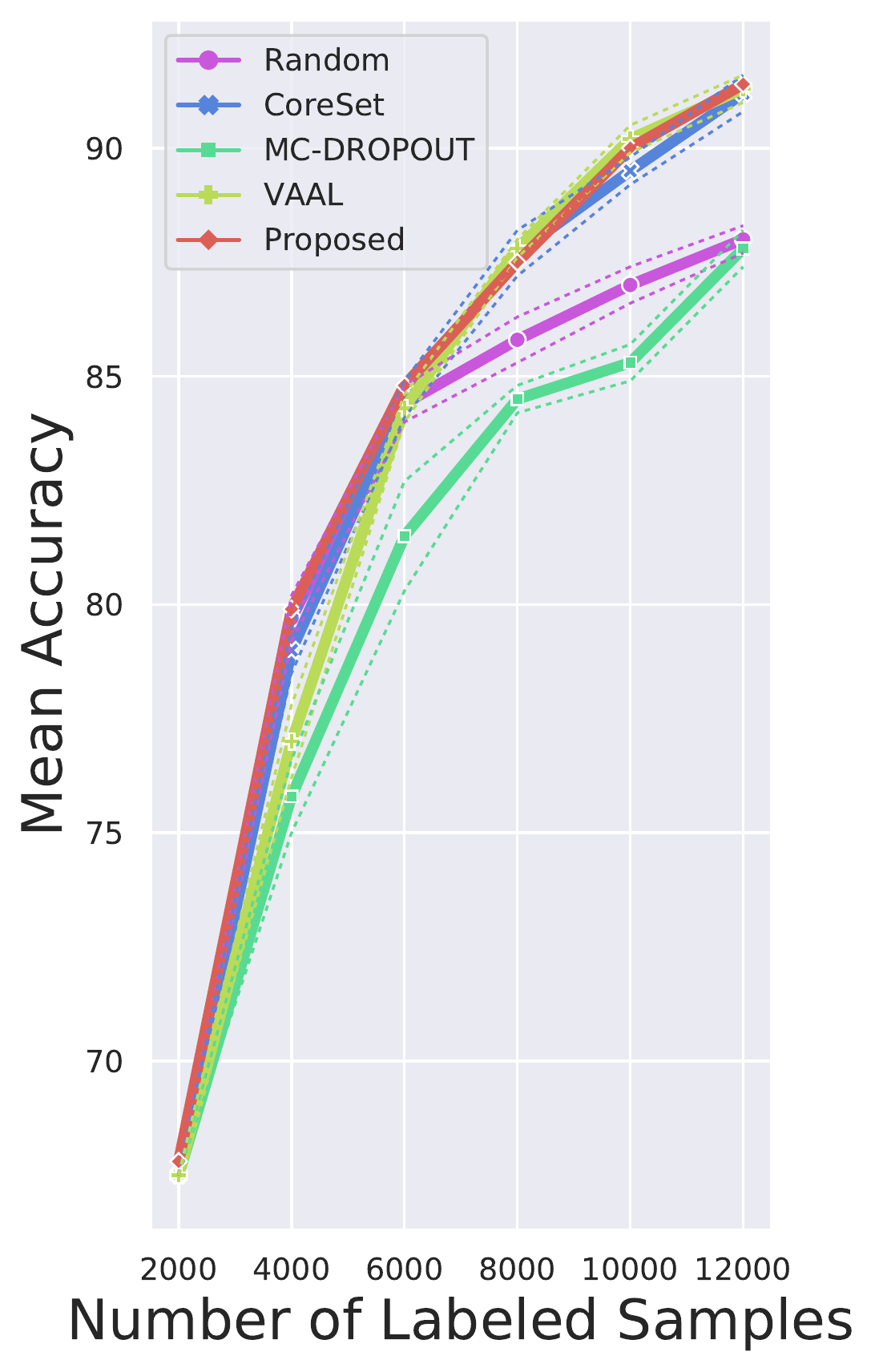}}
\subfigure[Trial $2$]{\includegraphics[width=0.32\linewidth]{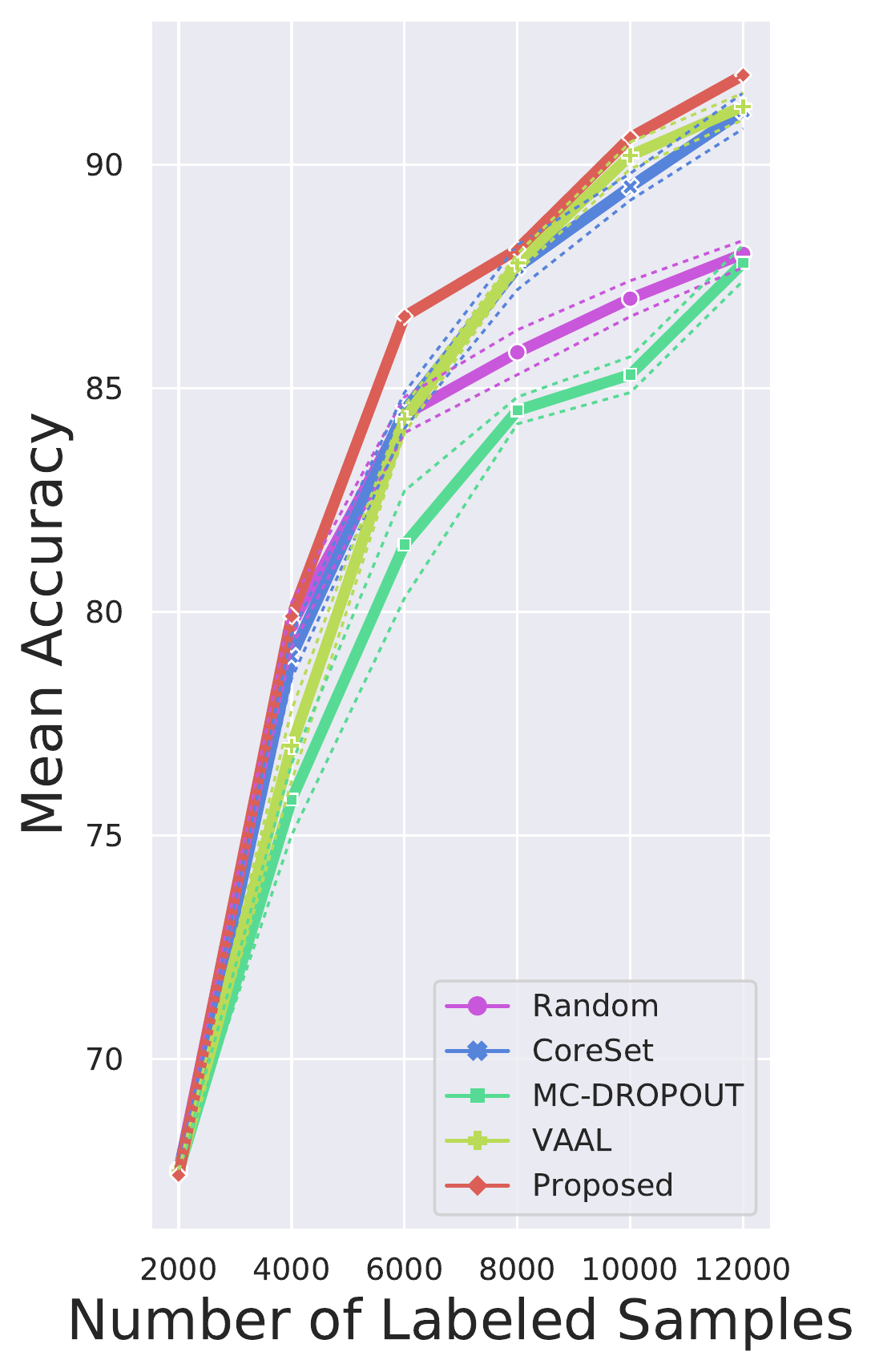}}
\subfigure[Trial $3$]{\includegraphics[width=0.32\linewidth]{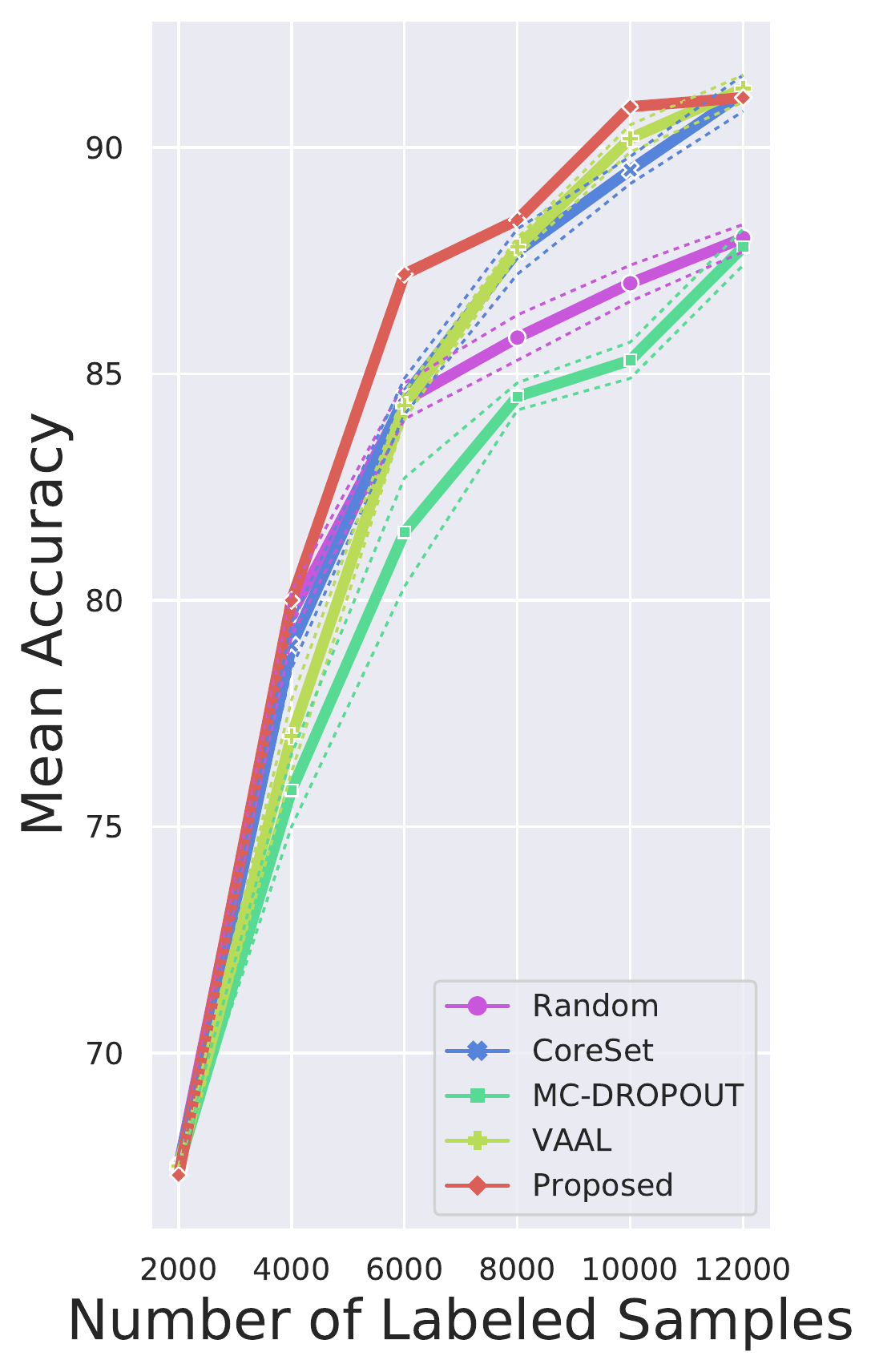}}
\caption{Individual results for the full CIFAR-10 dataset}
\label{fig:fullCIFAR10_single_0}
\end{figure*}

Finally, in Figs.~\ref{fig:fullNEU_single_0}$-$~\ref{fig:fullCIFAR10_single_0} we provide all the individual results of our framework before being averaged.
For competing methods, we plot the aggregated results for all three trials, so that one can easily compare with the maximum possible result of the competitors.
As shown, our results always outperform the competition.




\newpage




\begin{figure*}[t]
\centering
\subfigure[Trial $1$]{\includegraphics[width=0.32\linewidth]{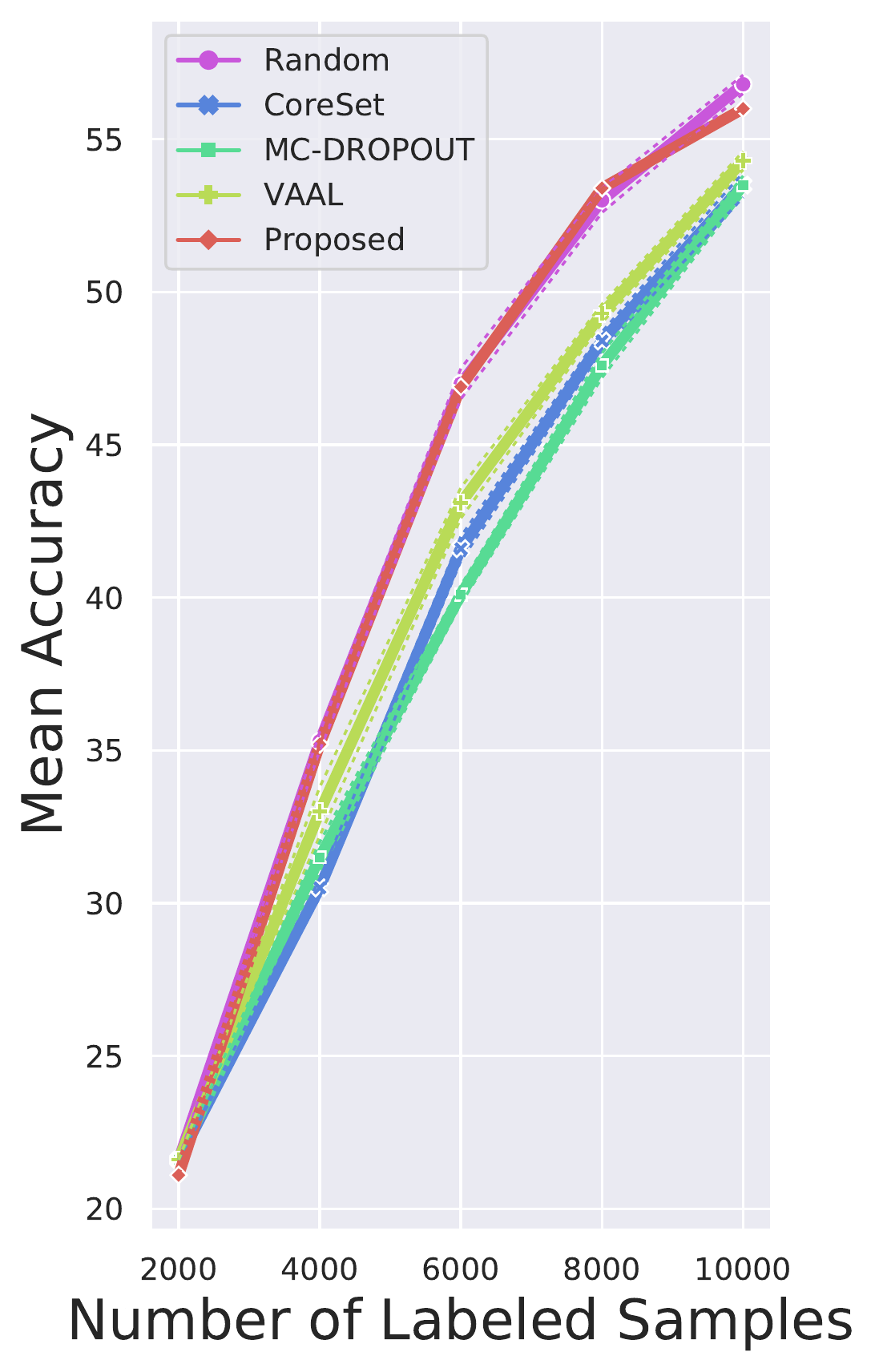}}
\subfigure[Trial $2$]{\includegraphics[width=0.32\linewidth]{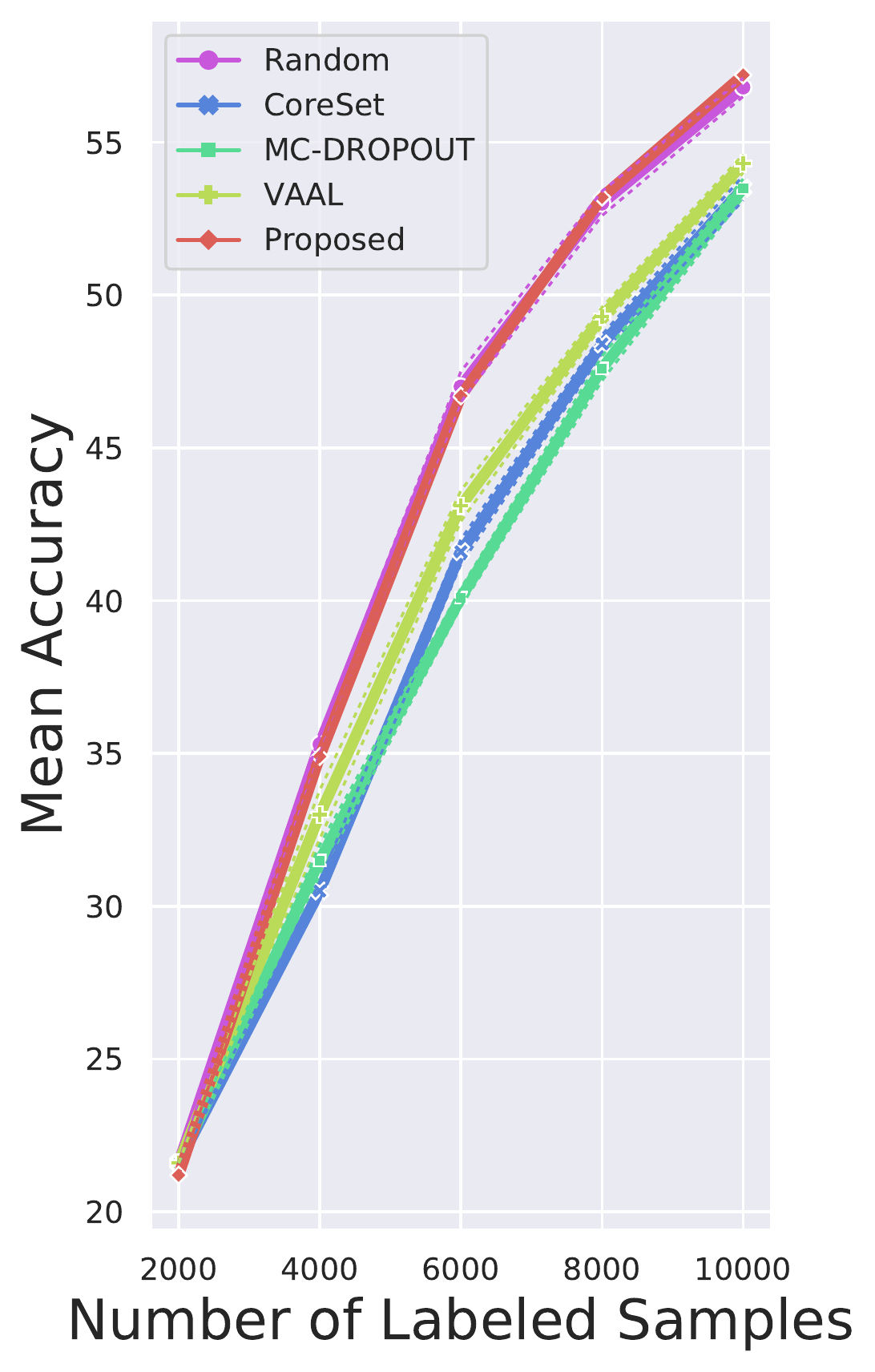}}
\subfigure[Trial $3$]{\includegraphics[width=0.32\linewidth]{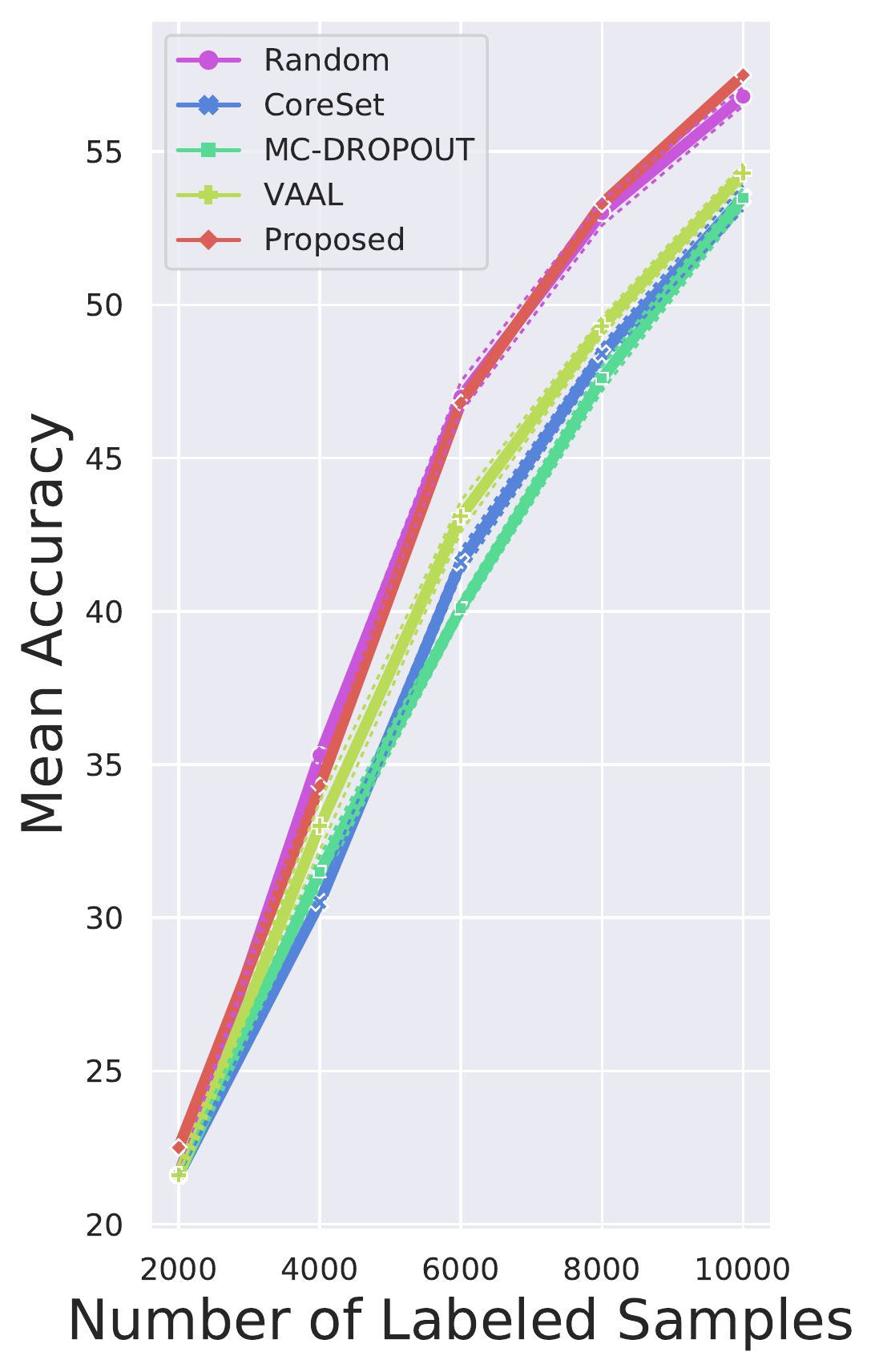}}
\caption{Individual results for the full CIFAR-100 dataset}
\label{fig:fullCIFAR100_single_0}
\end{figure*}





\end{document}